%% file: paper.tex
\documentclass[]{mosi}

\usepackage{helvet}

\usepackage{amsmath} 
\usepackage{natbib}
\usepackage{graphicx}
\usepackage{subcaption} 

\usepackage[toc,page,header]{appendix}
\usepackage[utf8]{inputenc} 
\usepackage[T1]{fontenc}    
\usepackage{hyperref}
\usepackage{url}            
\usepackage{booktabs}       

\usepackage{amsfonts}       
\usepackage{nicefrac}       
\usepackage{microtype}      
\usepackage{wrapfig}
\usepackage{amssymb}        

\usepackage{titletoc}
\usepackage{minitoc}

\usepackage{array}
\usepackage{etoolbox}

\definecolor{lightblue}{RGB}{200, 230, 255}  
\definecolor{headerblue}{RGB}{150, 200, 255} 

\usepackage{pgfplots}
\usepackage{xcolor}
\usepackage{float}
\usepackage{comment}
\usepackage{multirow}
\usepackage{makecell}
\usepackage{siunitx}
\usepackage{tikz}
\usepackage{pgf-pie}
\usepackage[export]{adjustbox}

\usepackage{ragged2e}
\usepackage{tabularx}
\usepackage{caption}
\usepackage{enumitem}
\usepackage{pifont}
\usepackage[hang,flushmargin]{footmisc}
\usepackage{times}
\usepackage{latexsym}
\usepackage{booktabs}
\usepackage{amsmath}
\usepackage{amssymb}
\usepackage{listings}
\usepackage{xcolor}
\usepackage[most]{tcolorbox}
\tcbuselibrary{listings,skins}
\usepackage{subcaption}
\usepackage{multirow}
\usepackage{algorithm}
\usepackage{algorithmic}
\usepackage{newfloat}
\usepackage{listings}

\usepackage{tcolorbox}
\tcbuselibrary{breakable}
\tcbuselibrary{skins}

\definecolor{PromptBack}{HTML}{F7FAFF}
\definecolor{PromptFrame}{HTML}{4C78A8}
\definecolor{PromptTitle}{HTML}{E8F1FF}

\newtcblisting{promptlisting}[2][]{
  enhanced,
  listing only,
  listing engine=listings,
  width=\textwidth,
  colback=PromptBack,
  colframe=PromptFrame,
  colbacktitle=PromptTitle,
  coltitle=black,
  fonttitle=\bfseries,
  title={#2},
  boxrule=0.45pt,
  arc=1.2mm,
  left=1mm,
  right=1mm,
  top=1mm,
  bottom=1mm,
  listing options={
    basicstyle=\ttfamily\footnotesize,
    numberstyle=\scriptsize,
    xleftmargin=1.5em,
    numbersep=5pt,
    breaklines=true,
    breakatwhitespace=false,
    columns=fullflexible,
    keepspaces=true,
    showstringspaces=false,
    tabsize=2
  },
  #1
}

\usepackage{threeparttable}
\usepackage{setspace}

\usepackage[scheme=plain,fontset=none]{ctex}

\usepackage[table]{xcolor}

\definecolor{oursgray}{gray}{0.95}

\definecolor{MossCyan}{HTML}{82D9FF} 
\definecolor{MossBlue}{HTML}{82B1FF}

\definecolor{tickG}{HTML}{00C853}
\definecolor{crossR}{HTML}{FF1744}

\floatstyle{ruled}
\newfloat{listing}{tb}{lst}{}
\floatname{listing}{Listing}

\newcommand{\faGithub}{\raisebox{-0.2ex}{\includegraphics[height=2.0ex]{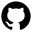}}}

\newtcolorbox{promptbox}[2][]{
    colback=white,
    coltext=black,
    arc=3mm,
    boxrule=0.5pt,
    colframe=black!60!white,
    title={#2},
    colbacktitle=black,
    coltitle=white,
    fonttitle=\bfseries,
    top=8pt,
    bottom=8pt,
    left=10pt,
    right=10pt,
    breakable,
    before upper={%
        \linespread{1}\selectfont
        \setlength{\parskip}{1ex plus 0.2ex minus 0.2ex}%
        \setlength{\parindent}{0pt}%
    },
    #1
}
\title{AgentHPOBench: A Benchmark For Evaluating LLM Agents as Sequential Hyperparameter Optimizers}

\author{
Tianyu Huai$^{2,3,4}$,
Tingshuo Fan$^{1,4}$,
Xinchi Chen$^{1,4,\dagger}$,
Yining Zheng$^{1,2,4}$,
Yuxin Wang$^{1,2,4}$,\\
Shuang Chen$^{1,2,4}$,
Jie Zhou$^{3}$,
Xuanjing Huang$^{1,2}$ 
\\[2mm] 
{\normalfont \normalsize $^{1}$Fudan University}
{\normalfont \normalsize $^{2}$Shanghai Innovation Institute}
{\normalfont \normalsize $^{3}$East China Normal University}
{\normalfont \normalsize $^{4}$OpenMOSS}\\
}

\abstract{
As LLMs evolve from code completion systems into autonomous scientific agents, evaluating their ability to conduct experiments has become increasingly important. Existing benchmarks typically focus on static code generation, paper replication, or final answer correctness, but do not directly assess whether agents can interpret experimental evidence and use it to guide subsequent hyperparameter decisions. To address this gap, we introduce AgentHPOBench, a sequential benchmark comprising 30 executable machine learning tasks across seven research categories. Each task begins with a validated baseline run, after which an agent performs several sequential interventions. At each step, the agent observes the accumulated configurations, metrics, and logs before proposing the next valid configuration. We evaluate 12 widely used agents and conventional HPO baselines under a unified protocol. The results show that current agents exhibit measurable experimental optimization ability across domains, but still face clear limitations in sustained iterative refinement, complex log diagnosis, and consistent progress toward reported reference performance.
}

\checkdata[\raisebox{-0.1ex}{\faGithub}\hspace{0.4em}GitHub]{\url{https://github.com/OpenMOSS/AgentHPOBench}}

\begin{document}
\maketitle
\begingroup
\renewcommand{\thefootnote}{\fnsymbol{footnote}}
\setcounter{footnote}{0}
% \footnotetext[1]{These authors contributed equally.}
\footnotetext[2]{Corresponding authors.}
\endgroup

% ===== Original meta lines (content unchanged) =====
% \begin{spacing}{1.0}
% {\small \noindent \textbf{Date:} December 31, 2025 \par}
% {\small \noindent \textbf{Demo:} \url{https://openmoss.github.io/MOSS-Transcribe-Diarize-demo/} \par}
% {\small \noindent \textbf{Huggingface Space:} \url{https://huggingface.co/spaces/OpenMOSS-Team/MOSS-transcribe-diarize} \par}
% \end{spacing}

% ===== Original content starts here (sections/tables/figures unchanged) =====

\input{sections/introduction}

\input{sections/related_work}

\input{sections/method}

\input{sections/evaluation}

\input{sections/conclusion}

% ===== Bibliography =====
\clearpage
\bibliographystyle{unsrtnat}
\bibliography{main}

\clearpage

% % ===== Appendix (content unchanged) =====
% \clearpage
% \beginappendix
\input{sections/appendix}

% \startcontents[app]
% \begingroup
%   \renewcommand{\contentsname}{Appendix Contents}
%   \section*{\contentsname}
%   \printcontents[app]{}{1}{}
% \endgroup
% \newpage

% \input{appendix/data_statistics}
% \input{appendix/training_configurations}

\end{document}

%% file: sections/introduction.tex
\section{Introduction}
\label{intro}

As LLMs improve in reasoning, use of long contexts, and tool interaction, agents are becoming capable of completing increasingly complex workflows that form part of empirical research. Recent benchmarks have therefore moved beyond static question answering and isolated code generation toward executable research environments. MLGym~\cite{nathani2025mlgym} and MLE-Dojo~\cite{qiang2025mledojo} already provide interactive environments in which agents execute experiments and refine solutions through feedback. PaperBench~\cite{starace2025paperbench}, AutoExperiment~\cite{kim2025autoexperiment}, RE-Bench~\cite{wijk2025re}, MLR-Bench~\cite{huang2025mlrbench}, and AIRS-Bench~\cite{lupidi2026airsbench} further extend evaluation toward paper replication, research engineering, and broader stages of the ML research lifecycle. These benchmarks demonstrate the value of iterative evaluation, but they generally assess broad research or engineering workflows in which improvements may arise from data processing, code modification, architecture design, debugging, hyperparameter tuning, or combinations of these actions.

Traditional HPO benchmarks address a different aspect of the problem. They provide controlled tabular, surrogate, or executable objectives for comparing optimization algorithms~\cite{eggensperger2021hpobench,pfisterer2022yahpo}, but typically abstract away the logs, configurations, and procedural context of research repositories. This leaves a specific capability insufficiently examined: whether an agent can interpret evidence from completed repository experiments and convert it into the next effective hyperparameter configuration. In a typical ML workflow, researchers execute a baseline, inspect target and auxiliary metrics together with execution logs, and decide which parameters to adjust in the next experiment. We therefore ask a focused question: can an autonomous agent improve real ML experiments through a sequence of hyperparameter interventions guided by empirical feedback?

\begin{figure}[t]
\centering
\includegraphics[width=0.7\columnwidth]{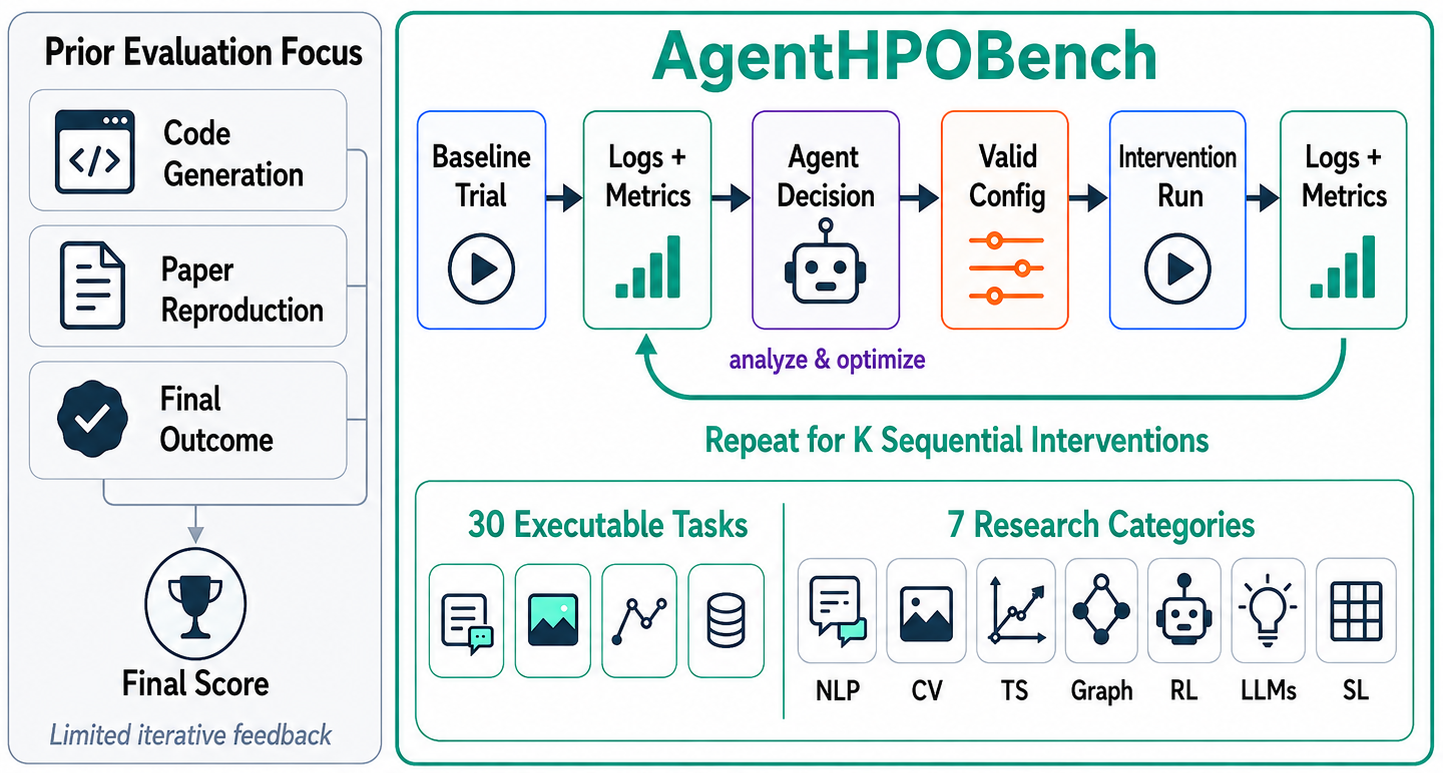}
\caption{Conceptual overview of AgentHPOBench.}
\label{fig:overview}
\end{figure}

\begin{table*}[!t]
\centering
\small
\setlength{\tabcolsep}{4pt}
\resizebox{0.98\textwidth}{!}{%
\begin{tabular}{lccccc}
\toprule
Benchmark & Interface & Task Unit & Decision Space & Observation & Primary Target \\
\midrule
HPO-B~\cite{arango2021hpo}
& Tabular & Objective & Hyperparameters & Metric & Optimizer \\
LCBench~\cite{zimmer2021lcbench}
& Tabular & Learning curve & Hyperparameters & Metric curve & Optimizer \\
HPOBench~\cite{eggensperger2021hpobench}
& Mixed & Objective & Hyperparameters & Metric & Optimizer \\
YAHPO Gym~\cite{pfisterer2022yahpo}
& Surrogate & Objective & Hyperparameters & Metric & Optimizer \\
JAHS-Bench-201~\cite{bansal2022jahsbench}
& Surrogate & Search space & Architecture + hyperparameters & Metrics & Optimizer \\
DACBench~\cite{eimer2021dacbench}
& Executable & Environment & Dynamic configuration & State + reward & Policy \\
\midrule
\textbf{AgentHPOBench}
& Executable & Research repository & Hyperparameters & Metrics + logs & HPO agent \\
\bottomrule
\end{tabular}%
}
\caption{Comparison with representative HPO and algorithm configuration benchmarks.}
\label{tab:hpo_benchmark_comparison}
\end{table*}

To address this question, we introduce \textbf{AgentHPOBench}, a dedicated and controlled benchmark for evaluating sequential HPO by agents in executable research repositories, as shown in Figure~\ref{fig:overview}. AgentHPOBench contains 30 executable tasks constructed from recent ML research repositories across seven categories: natural language processing, computer vision, time series forecasting, graph learning, reinforcement learning, large language modeling, and structured learning. By focusing on recent research repositories, the task suite captures contemporary training pipelines, configuration interfaces, and evaluation practices that are often abstracted away by conventional HPO benchmarks. Table~\ref{tab:hpo_benchmark_comparison} summarizes the differences between AgentHPOBench and representative HPO benchmarks.

AgentHPOBench comprises three components. First, the task construction protocol maps each repository to an executable optimization task with a reference baseline, a target metric, a constrained intervention space, and a paper or repository anchor. Second, the unified execution harness validates proposed configurations, executes experiments, and records the complete sequence of configurations, metrics, logs, and decisions. Third, the scoring and auditing layer verifies task traces and converts heterogeneous task results into mean bounded normalized score, baseline win rate, and mean anchor attainment. These metrics distinguish relative improvement over the reference baseline from absolute attainment of reported performance. Together, these components implement a common sequential evaluation protocol. Each task begins with a validated reference baseline that is shared by all agents and methods evaluated under the same budget setting. After the baseline run, the agent receives a fixed number of intervention opportunities. At each intervention, it observes the accumulated configurations, target and auxiliary metrics, and execution logs, and proposes a new configuration within the intervention space derived from the official training scripts, configuration files, or repository documentation. The dataset, data split, target metric, evaluation code, and benchmark metadata remain fixed. The audited result after the final intervention is used for scoring. AgentHPOBench therefore isolates the ability to convert experimental feedback into the next valid configuration.

We evaluate models with open weights and API agents, together with conventional HPO methods under the same baseline and number of intervention opportunities. The results show that current agents can improve reference baselines and that several agents obtain stronger aggregate results than conventional HPO methods. However, this advantage is not consistent across task categories or evaluation settings. Evaluation under the full training budget improves both mean bounded normalized score and mean anchor attainment, but does not uniformly increase baseline win rate. The feedback ablation shows that removing intermediate experimental evidence reduces optimization performance, while trajectory analysis reveals that later interventions may plateau or discard earlier gains. These findings show that discovering a useful configuration and reliably refining it through feedback are distinct capabilities.

In a nutshell, our contributions are as follows: 

\begin{itemize}
\item We introduce AgentHPOBench for evaluating whether agents can convert experimental feedback into effective hyperparameter decisions in executable research repositories. The benchmark contains 30 tasks from recent ML repositories across seven research categories.
\item We develop a unified evaluation framework with intervention spaces and an execution harness that validates, executes, and records agent interventions under shared reference baselines and fixed task definitions. The framework measures both improvement over the baseline and attainment of reported reference performance.
\item We evaluate several agents together with conventional HPO methods on AgentHPOBench. The results provide a systematic view of the capabilities and limitations of current agents in repository based HPO.
\end{itemize}

%% file: sections/related_work.tex
\section{Related Work}
\label{related}

\begin{figure*}[t]
\centering
\includegraphics[width=0.98\textwidth]{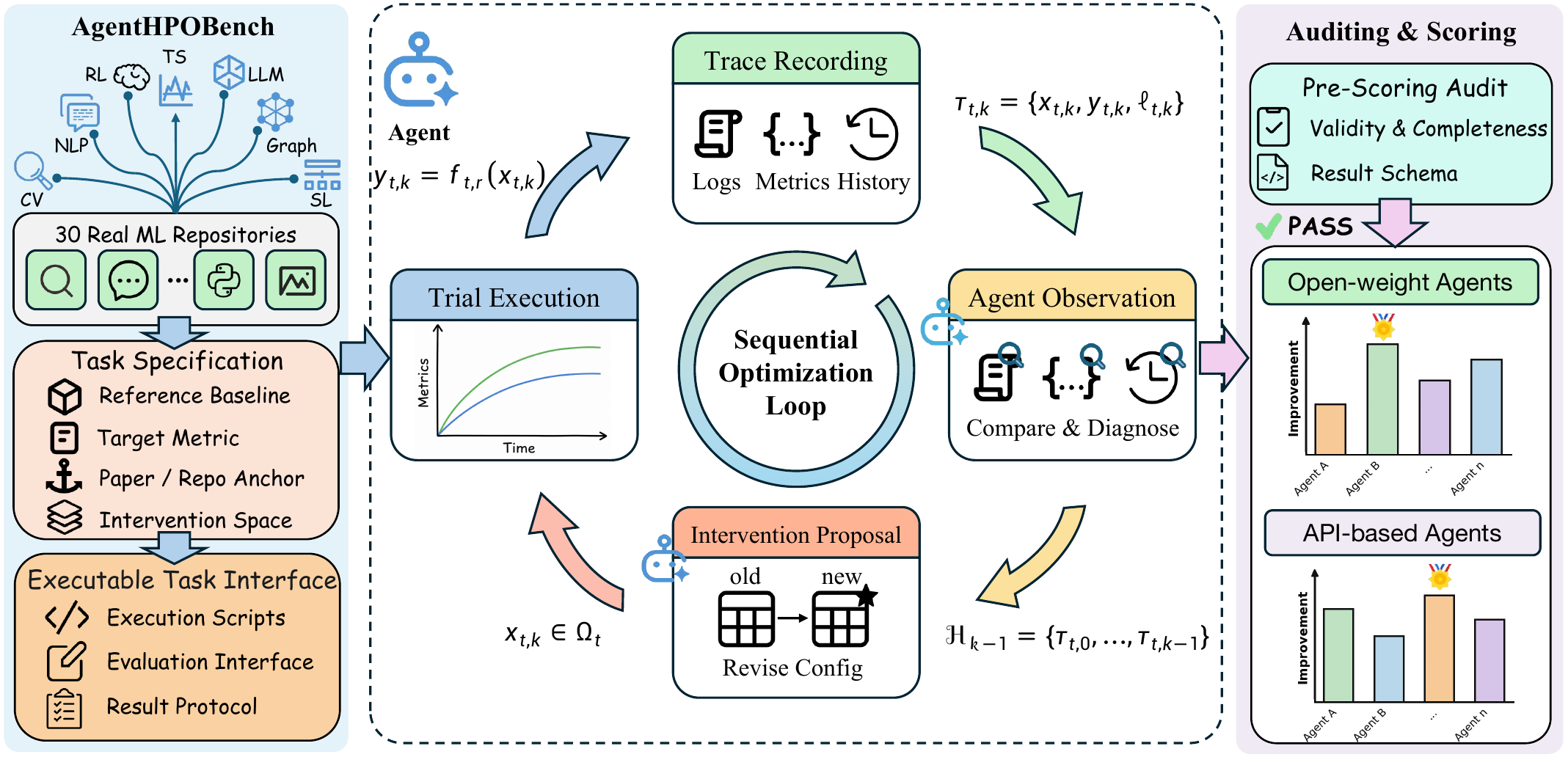}
\caption{Overview of AgentHPOBench. Each task is constructed from an executable ML repository with a reference baseline, target metric, intervention space, anchor, and standardized task interface. At each step, the agent observes the accumulated metrics and logs and proposes a valid configuration for the next run. The harness validates and executes the proposal, records the resulting trace, and audits the completed trajectory before scoring.}
\label{main}
\end{figure*}

\paragraph{Benchmarks for language agents.}
A growing body of work evaluates agents in ML research workflows. MLAgentBench~\cite{huang2024mlagentbench}, ML-Bench~\cite{yan2023mlbench}, and MLE-bench~\cite{chan2024mlebench} examine codebase use, model training, and ML engineering, while CORE-Bench~\cite{siebert2024corebench}, RE-Bench~\cite{wijk2025re}, and PaperBench~\cite{starace2025paperbench} focus on reproducibility, research engineering, and paper replication. MLGym~\cite{nathani2025mlgym}, MLE-Dojo~\cite{qiang2025mledojo}, MLR-Bench~\cite{huang2025mlrbench}, and AIRS-Bench~\cite{lupidi2026airsbench} broaden evaluation to more complete ML research workflows. These benchmarks assess general research execution or reproduction, whereas AgentHPOBench isolates whether agents can convert repository metrics and logs into the next valid hyperparameter configuration.

\paragraph{HPO methods and benchmarks.}
OpenML benchmark suites~\cite{bischl2017openml} and HPOBench~\cite{eggensperger2021hpobench} provide standardized tasks for comparing optimization algorithms, while NAS-Bench-101~\cite{ying2019nasbench101} provides reusable objectives for architecture search. Common HPO approaches include random search~\cite{bergstra2012random}, methods that use surrogate models~\cite{hutter2011smac,bergstra2011algorithms,snoek2012practical}, resource allocation methods such as Hyperband~\cite{li2017hyperband} and BOHB~\cite{falkner2018bohb}, and Population Based Training~\cite{jaderberg2017population}. These methods and benchmarks generally assume a predefined search space, a structured objective interface, and relatively clean numerical feedback.

\paragraph{LLMs and agents for optimization.}
Recent work also uses language models as optimizers or HPO assistants. OptFormer learns an optimizer from tuning traces~\cite{chen2022towards}, LLAMBO incorporates LLMs into Bayesian optimization~\cite{liu2024llambo}, and other studies investigate HPO decisions guided by LLMs and optimization frameworks based on agents~\cite{zhang2023llmhpo,liu2024agenthpo,mahammadli2024sllmbo}. Rather than proposing another optimization method, AgentHPOBench evaluates how reliably agents convert accumulated feedback from executable research repositories into valid configurations and empirical improvement.

%% file: sections/method.tex
\section{Method}
\label{sec:method}

We introduce AgentHPOBench, a benchmark and evaluation harness for assessing whether agents can improve executable ML experiments through sequential hyperparameter interventions, as shown in Figure~\ref{main}. Unlike agent benchmarks that evaluate general execution across multiple steps, code modification, or paper reproduction, AgentHPOBench isolates a specific experimental capability: converting the logs and metrics of an executed repository experiment into a valid configuration for the next run. The agent selects values only from the predefined intervention space, while the harness validates and executes the proposed configuration and returns the resulting observations. This process is repeated for a predefined number of intervention steps. Table~\ref{tab:hpo_benchmark_comparison} compares AgentHPOBench with representative HPO benchmarks. Most prior HPO benchmarks provide controlled tabular, surrogate, or wrapped objectives for comparing optimization algorithms, but abstract away repository execution details, textual logs, and sequential interaction. AgentHPOBench instead exposes real research repositories as executable task units and evaluates sequential optimization using task metrics tied to results reported by the corresponding papers or repositories.

\subsection{AgentHPOBench}
\label{sec:agenthpobench}

AgentHPOBench consists of 30 tasks constructed from 30 executable ML repositories on GitHub. Task construction follows three principles. First, each task must require a substantive experimental decision within an executable training or evaluation pipeline. Second, each task must provide measurable feedback after every intervention, allowing the benchmark to evaluate how the agent uses previous outcomes to inform subsequent decisions. Third, each task must remain close to real research practice by preserving the scripts, dependencies, logs, and failure modes of the original repository whenever possible.

Each task is based on an executable experiment from the original repository or its official reproduction environment. An executable task interface specifies how the harness launches the experiment, extracts the target metric, and records the outputs under a standardized result protocol. For each task, we construct the intervention space from the hyperparameters and valid values exposed by the official training scripts, configuration files, or repository documentation. We retain only fields that affect the execution or outcome of the experiment. Agents may modify only these predefined fields, while the dataset, data split, target metric, evaluation code, and benchmark metadata remain fixed.

\subsection{Problem Formulation}
\label{sec:formulation}
% baseline定义
We formulate autonomous hyperparameter optimization over executable research repositories as a sequential decision problem. Each task $t$ is defined by a repository specific experimental objective, a constrained intervention space $\Omega_t$, an evaluation protocol, and a scalar performance metric. For task $t$ under budget setting $r$, let $x_{t,0} \in \Omega_t$ denote the reference baseline configuration. This configuration is selected and validated during task construction and is shared by all agents and conventional optimizers. Executing $x_{t,0}$ under $r$ produces the reference baseline performance
\begin{equation}
y_{t,0}=f_{t,r}(x_{t,0}).
\end{equation}
The same $y_{t,0}$ serves as the reference baseline for all agents and conventional optimizers evaluated under the same budget setting. The configuration $x_{t,0}$ remains fixed across budget settings, but executing it under different budget may produce different baseline performance.

The agent first receives the reference baseline configuration $x_{t,0}$ and observes its performance $y_{t,0}$. The harness records each completed run, including the baseline, as a trace entry
\begin{equation}
\tau_{t,k}=\{x_{t,k},y_{t,k},\ell_{t,k}\},
\end{equation}
where $\ell_{t,k}$ contains the execution logs and auxiliary metrics made available to the agent. At intervention step $k \in \{1,\ldots,K\}$, the agent observes the accumulated trace history
\begin{equation}
\mathcal{H}_{t,k-1}
=
\{\tau_{t,0},\tau_{t,1},\ldots,\tau_{t,k-1}\},
\label{eq:history}
\end{equation}
uses this history to form an implicit experimental state, and proposes a new valid configuration $x_{t,k} \in \Omega_t$. After validating the proposal, the harness executes $x_{t,k}$ under budget setting $r$ and obtains
\begin{equation}
y_{t,k}=f_{t,r}(x_{t,k}),
\end{equation}
where the direction of improvement is specified by the task metric. The resulting trace $\tau_{t,k}$ is then added to the history, and the corresponding feedback is returned to the agent before the next intervention.

After $K$ sequential interventions, this process produces the trajectory
\begin{equation}
(x_{t,0},y_{t,0}),
(x_{t,1},y_{t,1}),
\ldots,
(x_{t,K},y_{t,K}).
\end{equation}

This formulation differs from conventional HPO benchmarks in two respects. First, the objective is embedded in an executable research repository rather than exposed through a clean objective interface. Second, the agent must interpret experimental evidence, including logs, metrics, task constraints, and prior execution outcomes, and convert this evidence into a valid configuration for the next experiment.

\subsection{Evaluation Harness}
\label{sec:harness}

To make the benchmark executable and comparable across agents, we implement a unified evaluation harness. The harness standardizes the interaction between agents and heterogeneous research repositories while preserving the original execution logic of each task. It launches experiments, provides task context to the agent, validates proposed configurations, and records the resulting outputs under a standardized result schema.

For each task, the harness maintains the agent visible trace, together with an internal execution record containing the proposed configurations, extracted metrics, execution logs, runtime metadata, and completion status. These records preserve the complete sequence of interventions and support consistent auditing across tasks and agents.

The harness also enforces comparability between models with open weights and API agents. All agents are evaluated using the same task definitions, budget setting, number of interventions, allowed intervention space, scoring rules, and result schema. The different backends receive the same task information through interfaces specific to each backend. Their outputs are parsed into a common configuration schema and validated against $\Omega_t$ before execution.

Before scoring, the harness audits all task records. The audit verifies the presence of the baseline and the required intervention records, compliance with the result schema, correct metric extraction, and consistency with the reference baseline and anchor. Only complete records that pass these checks are included in aggregate scoring.

\begin{table*}[!t]
\centering
\small
\setlength{\tabcolsep}{2.8pt}
\resizebox{0.98\textwidth}{!}{%
\begin{tabular}{lcccccccccc}
\toprule
\multirow{2}{*}{\textbf{Agent / Method}}
& \multicolumn{8}{c}{\textbf{Mean bounded normalized score}}
& \multicolumn{2}{c}{\textbf{Overall metrics}} \\
\cmidrule(lr){2-9}
\cmidrule(lr){10-11}
& NLP (3) & CV (5) & TS (7) & Graph (2) & RL (3) & LLM (4) & SL (6)
& Overall & BWR (\%) & MAA (\%) \\
\midrule
\multicolumn{11}{l}{\textit{Conventional HPO baselines}} \\
\midrule
Random search
& 0.034 & -0.135 & 0.299 & -0.463 & 0.124 & -0.292 & -0.136
& -0.034 & 48.9 & 62.6 \\
TPE
& 0.020 & -0.110 & 0.005 & -0.406 & 0.192 & -0.219 & -0.302
& -0.113 & 40.0 & 62.4 \\
BOHB variant
& -0.062 & -0.006 & 0.235 & -0.824 & 0.260 & -0.236 & 0.153
& 0.018 & 45.6 & 65.3 \\
\midrule
\multicolumn{11}{l}{\textit{Open-weight agents}} \\
\midrule
Gemma2-2B
& 0.014 & -0.103 & 0.233 & -0.459 & 0.005 & -0.171 & 0.066
& -0.001 & 55.6 & 64.2 \\
DeepSeek-R1-Qwen-14B
& 0.008 & -0.223 & 0.316 & -0.333 & 0.074 & -0.271 & 0.157
& 0.018 & 54.4 & 64.1 \\
Qwen3-8B
& -0.001 & -0.140 & 0.233 & -0.271 & 0.158 & -0.219 & 0.124
& 0.024 & 53.3 & 65.1 \\
Llama-3.1-8B
& 0.019 & -0.142 & 0.253 & -0.525 & 0.125 & -0.144 & 0.172
& 0.030 & 44.4 & 66.3 \\
Phi-4-14B
& 0.047 & -0.077 & 0.278 & -0.448 & 0.306 & -0.098 & 0.430
& 0.130 & 63.3 & 66.9 \\
Qwen3-32B
& \textbf{0.060} & -0.120 & 0.285 & -0.302 & 0.349 & -0.120 & 0.485
& 0.148 & 60.0 & 69.1 \\
\midrule
\multicolumn{11}{l}{\textit{API agents}} \\
\midrule
GLM-5.1
& -0.088 & -0.108 & 0.267 & -1.000 & 0.032 & -0.258 & \textbf{0.713}
& 0.080 & 56.7 & 67.0 \\
Kimi-2.6
& -0.053 & -0.106 & 0.277 & -1.000 & 0.466 & -0.173 & 0.556
& 0.110 & 56.7 & 70.4 \\
GLM-4.7
& -0.088 & 0.119 & 0.266 & -1.000 & 0.061 & -0.285 & 0.689
& 0.112 & 60.0 & 67.8 \\
DeepSeek-V4-Pro
& -0.029 & -0.184 & 0.346 & 0.094 & 0.408 & 0.036 & 0.378
& 0.175 & 63.3 & 70.1 \\
GPT-5.5
& -0.028 & -0.033 & 0.375 & 0.281 & \textbf{0.821} & 0.092 & 0.565
& 0.305 & 66.7 & 76.7 \\
Claude Sonnet 4.6
& -0.025 & \textbf{0.191} & \textbf{0.403} & \textbf{0.877} & 0.710
& \textbf{0.120} & 0.691 & \textbf{0.407} & \textbf{76.7} & \textbf{79.5} \\
\bottomrule
\end{tabular}%
}
\caption{AgentHPOBench results under the limited budget protocol for conventional HPO methods, open-weight agents, and API agents. Category columns report mean bounded normalized score. BWR and MAA are computed over all 30 tasks. \textbf{Bold} indicates the best performance across all evaluated methods.}
\label{tab:main-category-results}
\vspace{-3mm}
\end{table*}

\subsection{Scoring}
\label{sec:scoring}

Because the benchmark tasks use heterogeneous metrics, we evaluate each task using the audited result after the final intervention. For task $t$ under budget setting $r$, let $b_{t,r}=y_{t,0}$ denote the reference baseline, $s_{t,r}=y_{t,K}$ denote the final result, and $a_t$ denote the repository anchor. Before scoring, these values are expressed on the same numerical scale and oriented so that larger values indicate better performance. We denote the oriented values by $\tilde{b}_{t,r}$, $\tilde{s}_{t,r}$, and $\tilde{a}_t$.

We use mean bounded normalized score and baseline win rate as the primary metrics. The normalized score is
\begin{equation}
\mathrm{NS}_{t,r}
=
\frac{\tilde{s}_{t,r}-\tilde{b}_{t,r}}
{\tilde{a}_t-\tilde{b}_{t,r}},
\end{equation}
where the audit ensures that the denominator is positive and nonzero. To reduce the influence of tasks with a small baseline to anchor gap, we bound each task score to $[-1,1]$:
\begin{equation}
\mathrm{BNS}_{t,r}
=
\min\{1,\max\{-1,\mathrm{NS}_{t,r}\}\}.
\end{equation}
The mean bounded normalized score (MBNS) is
\begin{equation}
\mathrm{MBNS}_{r}
=
\frac{1}{T}\sum_{t=1}^{T}\mathrm{BNS}_{t,r},
\end{equation}
where $T$ is the number of evaluated tasks. A bounded score of $0$ matches the baseline, $1$ reaches or exceeds the anchor, and a negative value indicates degradation.

The baseline win rate (BWR) is defined as
\begin{equation}
\mathrm{BWR}_{r}
=
\frac{1}{T}\sum_{t=1}^{T}
\mathbf{1}\left[\tilde{s}_{t,r}>\tilde{b}_{t,r}\right].
\end{equation}
It measures the proportion of tasks for which the final result improves over the reference baseline.

As a secondary metric, we report anchor attainment in the original metric direction:
\begin{equation}
\mathrm{AA}_{t,r}
=
\begin{cases}
s_{t,r}/a_t, & \text{for higher is better metrics},\\
a_t/s_{t,r}, & \text{for lower is better metrics}.
\end{cases}
\end{equation}
Mean anchor attainment (MAA) is defined as
\begin{equation}
\mathrm{MAA}_{r}
=
\frac{1}{T}\sum_{t=1}^{T}\mathrm{AA}_{t,r}.
\end{equation}
An anchor attainment of $100\%$ indicates that the final result matches the repository anchor, while values above indicate that it exceeds the anchor.

%% file: sections/evaluation.tex
\section{Experiments}
\label{sec:experiments}

\paragraph{Experimental Setup and Metrics.}
We evaluate agents on AgentHPOBench across seven categories: NLP, CV, TS, Graph, RL, LLM, and SL. Unless otherwise stated, we use the limited budget protocol. The harness first executes the reference baseline and then requests five sequential interventions from the agent. Under this protocol, both the baseline and each intervention use approximately 10\% of the training budget of the corresponding original experiment. This setting retains executable training and empirical feedback for each task while making broad evaluation across repositories and agents computationally feasible. Before each intervention, the agent receives the accumulated trace, including previous configurations, target and auxiliary metrics, and execution logs. All task scores use the result after the final intervention rather than the best intermediate result. We report mean bounded normalized score (MBNS), baseline win rate (BWR), and mean anchor attainment (MAA) as the metrics. Category scores average the tasks within each category, whereas overall scores are computed directly over all 30 tasks, giving each task equal weight.

\paragraph{Agents and Implementation Details.}
We evaluate both models with open weights and API agents. The models with open weights include Qwen3-8B and Qwen3-32B~\citep{yang2025qwen3}, Gemma2-2B~\citep{team2024gemma}, DeepSeek-R1-Qwen-14B~\citep{guo2025deepseek}, Phi-4-14B~\citep{abdin2024phi4}, and Llama-3.1-8B-Instruct~\citep{grattafiori2024llama}. The API agents include DeepSeek-V4-Pro, GPT-5.5, GLM-4.7, GLM-5.1, Kimi-2.6, and Claude Sonnet 4.6. We additionally compare these agents with random search~\citep{bergstra2012random}, TPE~\citep{bergstra2011algorithms}, and a BOHB-style method~\citep{falkner2018bohb} as conventional HPO baselines. All agents and HPO baselines use the same task definitions, reference baselines, intervention spaces, number of configuration evaluations, budget settings, and scoring rules. The conventional HPO baselines operate on previously evaluated configurations and target metric values, whereas agents additionally receive task context, auxiliary metrics, and execution logs. We repeat all locally executable open-weight agents and conventional HPO baselines using three experiment seeds, \(\{0,1,42\}\). For these methods, the implementation, model checkpoint or optimization algorithm, runtime environment, and random state can be explicitly controlled, allowing the repetitions to measure sensitivity to benchmark stochasticity. API agents are evaluated once on the canonical benchmark instance. Further details are provided in Appendix.

\begin{table*}[!t]
\centering
\small
\setlength{\tabcolsep}{2.8pt}
\resizebox{0.98\textwidth}{!}{%
\begin{tabular}{lcccccccccc}
\toprule
\multirow{2}{*}{\textbf{Agents}}
& \multicolumn{8}{c}{\textbf{Mean bounded normalized score}}
& \multicolumn{2}{c}{\textbf{Overall metrics}} \\
\cmidrule(lr){2-9}
\cmidrule(lr){10-11}
 & NLP (3) & CV (5) & TS (7) & Graph (2) & RL (3) & LLM (4) & SL (6)
& Overall & BWR (\%) & MAA (\%) \\
\midrule
DeepSeek-R1-Qwen-14B
& 0.029 & -0.125 & 0.369 & -0.094 & 0.185 & -0.145 & 0.448
& 0.151 & 50.0 & 78.6 \\
Qwen3-32B
& 0.029 & 0.145 & 0.293 & -0.594 & 0.488 & -0.015 & 0.442
& 0.191 & 56.7 & 81.8 \\
Claude Sonnet 4.6
& \textbf{0.722} & \textbf{0.266} & \textbf{0.455} & \textbf{0.502}
& \textbf{0.754} & \textbf{0.184} & \textbf{0.582}
& \textbf{0.472} & \textbf{76.7} & \textbf{89.1} \\
\bottomrule
\end{tabular}%
}
\caption{Full-budget final-step AgentHPOBench results.}
\label{tab:full-budget-category-results}
\end{table*}

\begin{table*}[!t]
\centering
\small
\setlength{\tabcolsep}{2.8pt}
\resizebox{0.98\textwidth}{!}{%
\begin{tabular}{lcccccccccc}
\toprule
\multirow{2}{*}{\textbf{Harness}}
& \multicolumn{8}{c}{\textbf{Mean bounded normalized score}}
& \multicolumn{2}{c}{\textbf{Overall metrics}} \\
\cmidrule(lr){2-9}
\cmidrule(lr){10-11}
& NLP (3) & CV (5) & TS (7) & Graph (2) & RL (3) & LLM (4) & SL (6)
& Overall & BWR (\%) & MAA (\%) \\
\midrule
\multicolumn{11}{l}{\textit{Claude Sonnet 4.6}} \\
\midrule
Claude Code CLI
& \textbf{0.090} & 0.120 & 0.295 & 0.518 & \textbf{0.800}
& -0.170 & \textbf{0.712} & 0.332 & \textbf{76.7} & \textbf{82.2} \\
AgentHPOBench
& -0.025 & \textbf{0.191} & \textbf{0.403} & \textbf{0.877} & 0.710
& \textbf{0.120} & 0.691 & \textbf{0.407} & \textbf{76.7} & 79.5 \\
\midrule
\multicolumn{11}{l}{\textit{GPT-5.5}} \\
\midrule
Codex CLI
& \textbf{0.050} & \textbf{0.108} & 0.318 & \textbf{0.565} & 0.666
& -0.145 & 0.418 & 0.266 & \textbf{70.0} & \textbf{78.7} \\
AgentHPOBench
& -0.028 & -0.033 & \textbf{0.375} & 0.281 & \textbf{0.821}
& \textbf{0.092} & \textbf{0.565} & \textbf{0.305} & 66.7 & 76.7 \\
\bottomrule
\end{tabular}%
}
\caption{Harness ablation under the limited-budget protocol.}
\label{tab:harness-ablation}
\end{table*}

\paragraph{Main Results.}
Table~\ref{tab:main-category-results} summarizes the results under the limited-budget protocol. Claude Sonnet 4.6 achieves the highest overall MBNS of 0.407, the highest BWR of 76.7\%, and the highest MAA of 79.5\%. Among open-weight agents, Qwen3-32B obtains the highest overall MBNS (0.148) and MAA (69.1\%), whereas Phi-4-14B achieves the highest BWR (63.3\%). The conventional HPO methods generally trail the stronger agents in overall performance under the same five intervention opportunities. Among these baselines, the BOHB variant obtains the highest overall MBNS (0.018) and MAA (65.3\%), while random search achieves the highest BWR (48.9\%).

Performance varies substantially across task categories, and no method performs best in every category. Qwen3-32B achieves the highest MBNS on NLP. Claude Sonnet 4.6 leads on CV, TS, Graph, and LLM, GPT-5.5 leads on RL, and GLM-5.1 leads on SL. These results demonstrate that the relative effectiveness of agents and conventional HPO methods depends strongly on the experimental domain. The category-level HPO results also show that conventional optimization remains competitive in specific domains despite its lower overall performance.

Aggregate leadership does not imply consistent improvement across individual tasks. Although Claude Sonnet 4.6 obtains the strongest aggregate results, its BWR of 76.7\% means that its final configurations exceed the reference baseline on 23 of the 30 tasks. Qwen3-32B obtains a mean BWR of 60.0\% across the three controlled seeds. The remaining failures, together with the negative and near-zero category scores, show that even strong agents can produce final configurations that do not improve the baseline. These findings indicate that the main challenge is not only to identify useful interventions, but also to refine and preserve their benefits throughout the sequential optimization process.

\begin{figure*}[t]
\centering
\begin{subfigure}[b]{0.46\textwidth}
    \centering
    \includegraphics[width=\textwidth]{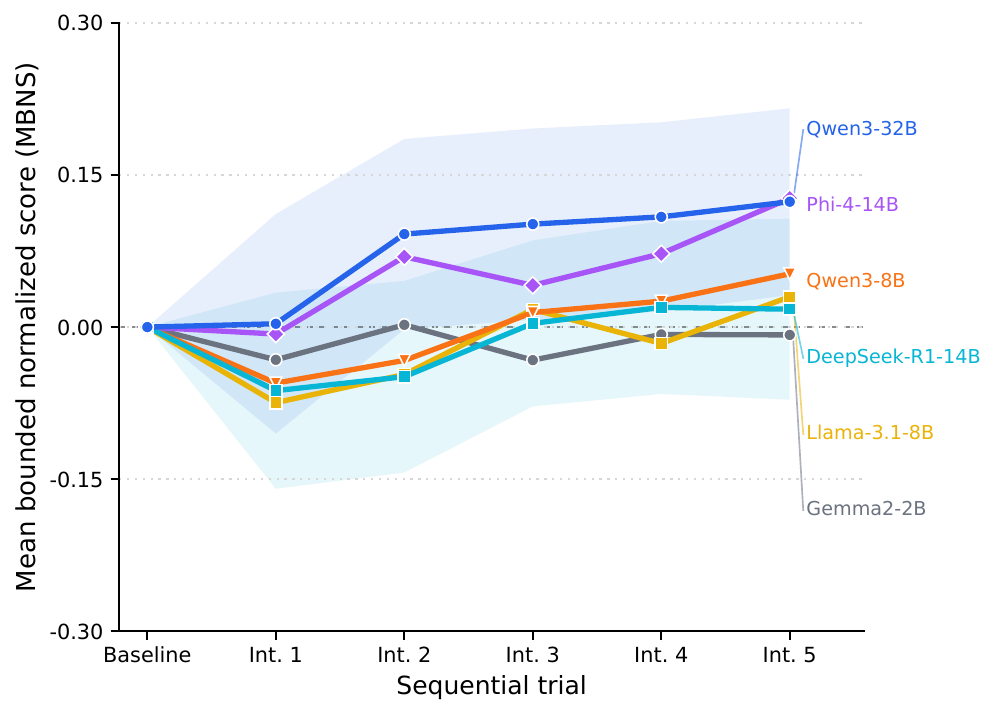}
    \vspace{-5mm}
    \caption{Open-weight agents}
    \vspace{-1mm}
    \label{exp1:open}
\end{subfigure}
\hfill
\begin{subfigure}[b]{0.46\textwidth}
    \centering
    \includegraphics[width=\textwidth]{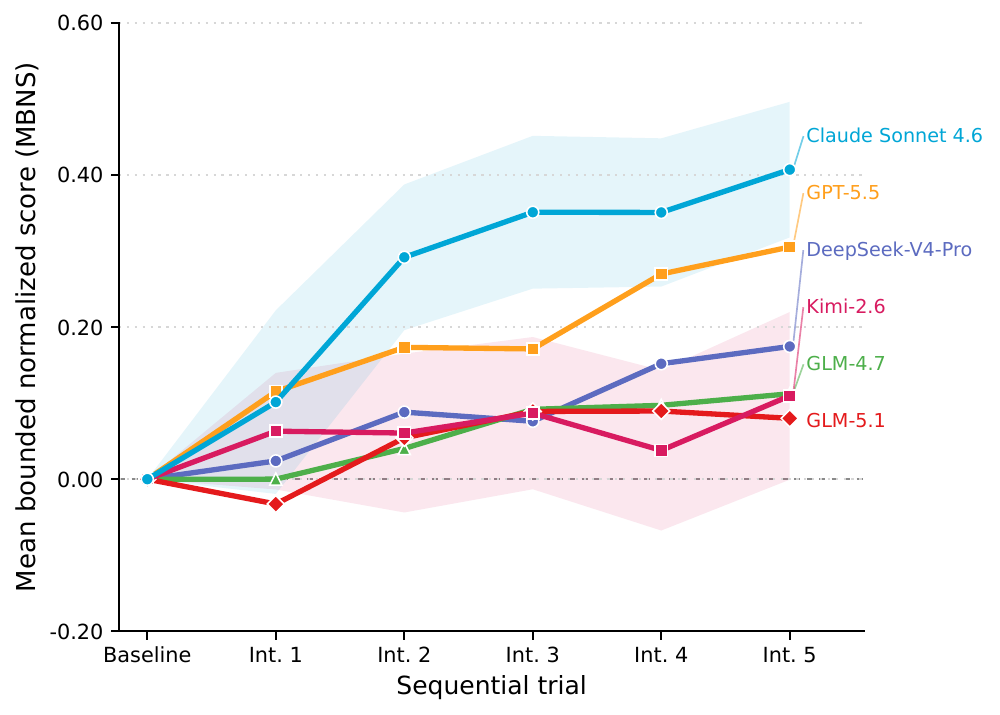}
    \vspace{-5mm}
    \caption{API-based agents}
    \vspace{-1mm}
    \label{exp1:api}
\end{subfigure}
\caption{Sequential optimization trajectories under the limited budget protocol. Lines show MBNS from the reference baseline through five interventions, and shaded regions indicate cross-task standard error.}
\label{exp1}
\end{figure*}

\begin{table*}[!t]
\centering
\small
\setlength{\tabcolsep}{2.8pt}
\resizebox{0.98\textwidth}{!}{%
\begin{tabular}{lcccccccccc}
\toprule
\multirow{2}{*}{\textbf{Feedback setting}}
& \multicolumn{8}{c}{\textbf{Mean bounded normalized score}}
& \multicolumn{2}{c}{\textbf{Overall metrics}} \\
\cmidrule(lr){2-9}
\cmidrule(lr){10-11}
& NLP (3) & CV (5) & TS (7) & Graph (2) & RL (3) & LLM (4) & SL (6)
& Overall & BWR (\%) & MAA (\%) \\
\midrule
Standard feedback
& \textbf{0.060} & \textbf{-0.120} & \textbf{0.285}
& \textbf{-0.302} & \textbf{0.349} & \textbf{-0.120}
& \textbf{0.485} & \textbf{0.148} & \textbf{60.0} & \textbf{69.1} \\
No intermediate feedback
& 0.021 & -0.227 & 0.211
& -0.333 & 0.346 & -0.234
& 0.286 & 0.052 & 50.0 & 65.2 \\
\bottomrule
\end{tabular}%
}
\caption{Ablation of intermediate experimental feedback for Qwen3-32B under the limited budget protocol.}
\label{tab:no-feedback-ablation}
\end{table*}

% \paragraph{Full Budget Results.}
% To examine how the training budget affects sequential optimization, we evaluate three selected agents using the full training budget of the original repositories, as shown in Table~\ref{tab:full-budget-category-results}. The full budget increases MAA for all three agents, with Claude Sonnet 4.6 achieving the highest MAA of 89.1\%. However, neither MBNS nor BWR improves uniformly. The MBNS of Qwen3-32B decreases slightly from 0.229 to 0.218, whereas the MBNS values of DeepSeek-R1-Qwen-14B and Claude Sonnet 4.6 increase from 0.100 to 0.178 and from 0.407 to 0.472, respectively. The BWR of Claude Sonnet 4.6 remains at 76.7\%, while those of Qwen3-32B and DeepSeek-R1-Qwen-14B decrease to 56.7\% and 50.0\%. These changes are not contradictory because the three metrics capture different aspects of performance. MAA measures absolute attainment relative to the anchor, MBNS measures the magnitude of improvement over the reference baseline under the same budget, and BWR measures the proportion of tasks on which the final result exceeds that baseline. Full training can change both the baseline performance and the final performance obtained by an agent. Consequently, an agent may finish closer to the anchor while achieving a smaller relative improvement or exceeding the corresponding full budget baseline on fewer tasks. The limited and full budget settings therefore provide complementary evidence about optimization under different computational constraints.

\paragraph{Full Budget Results.}
To examine how the training budget affects sequential optimization, we evaluate three selected agents using the full training budget of the original repositories, as shown in Table~\ref{tab:full-budget-category-results}. The full budget increases both MBNS and MAA for all three agents. The MBNS values of Qwen3-32B, DeepSeek-R1-Qwen-14B, and Claude Sonnet 4.6 increase from 0.148, 0.018, and 0.407 under the limited-budget protocol to 0.191, 0.151, and 0.472, respectively. Claude Sonnet 4.6 also achieves the highest full-budget MAA of 89.1\%. However, BWR does not improve uniformly: Claude Sonnet 4.6 remains at 76.7\%, while Qwen3-32B and DeepSeek-R1-Qwen-14B decrease from 60.0\% and 54.4\% to 56.7\% and 50.0\%, respectively. These changes are not contradictory because the three metrics capture different aspects of performance. Full training can change both the baseline performance and the final performance obtained by an agent. Consequently, an agent may finish closer to the anchor while achieving a larger average normalized improvement, yet exceed the corresponding full-budget baseline on fewer tasks. The limited- and full-budget settings therefore provide complementary evidence about optimization under different computational constraints.

\paragraph{Harness Ablation.}
We examine how the execution harness affects performance by comparing the native AgentHPOBench harness with corresponding CLI harnesses under the same tasks, intervention spaces, budgets, and scoring rules. The results are shown in Table~\ref{tab:harness-ablation}. For Claude Sonnet 4.6, the native harness achieves a higher MBNS than Claude Code CLI (0.407 vs. 0.332), while both harnesses obtain the same BWR of 76.7\%. Claude Code CLI achieves a higher MAA (82.2\% vs. 79.5\%). Similarly, for GPT-5.5, the native harness achieves a higher MBNS than Codex CLI (0.305 vs. 0.266), whereas Codex CLI obtains a higher BWR (70.0\% vs. 66.7\%) and MAA (78.7\% vs. 76.7\%). These results show that the native harness produces greater relative improvement over the reference baseline for both agents, while the CLI harnesses achieve higher absolute anchor attainment. However, no harness consistently dominates across all metrics or task categories. Harness choice can therefore affect the configurations produced during sequential optimization and the resulting performance. This sensitivity motivates the use of a common harness for the main comparison and the separate reporting of harness effects.

\paragraph{Sequential Optimization Trajectories.}
Figure~\ref{exp1} reports the MBNS obtained after the reference baseline and each subsequent intervention. Claude Sonnet 4.6 and Qwen3-32B achieve substantial gains during the first two interventions, while GPT-5.5 improves more gradually across the trajectory. Other agents exhibit less stable behavior. DeepSeek-R1-Qwen-14B initially falls below the baseline before recovering, whereas Gemma2-2B and Llama-3.1-8B fluctuate around the baseline for most of the trajectory. Several agents also show nonmonotonic refinement. For example, Phi-4-14B loses part of its early improvement before recovering at the final intervention, while Kimi-2.6 declines at the fourth intervention and subsequently rebounds. In contrast, Claude Sonnet 4.6 and GPT-5.5 achieve their highest MBNS at the final intervention, showing that later feedback can still produce useful refinements. Overall, current agents can identify beneficial configurations, but their ability to preserve and improve earlier gains remains inconsistent across models.

% \paragraph{Intermediate Feedback Ablation.}
% We isolate the contribution of intermediate experimental feedback by evaluating Qwen3-32B under an otherwise identical protocol. The baseline observation and all other task inputs remain available, but the metrics and execution logs produced by each intervention are withheld from subsequent decisions. As shown in Table~\ref{tab:no-feedback-ablation}, removing intermediate feedback reduces MBNS from 0.229 to 0.082 and BWR from 66.7\% to 53.3\%, while MAA decreases from 69.5\% to 66.8\%. Performance decreases in six of the seven task categories, with CV as the only exception. These results indicate that intermediate feedback improves sequential optimization for Qwen3-32B, although the benefit is not uniform across task categories.

% \paragraph{Intermediate Feedback Ablation.}
% We isolate the contribution of intermediate experimental feedback by evaluating Qwen3-32B under an otherwise identical protocol. The baseline observation and all other task inputs remain available, but the metrics and execution logs produced by each intervention are withheld from subsequent decisions. As shown in Table~\ref{tab:no-feedback-ablation}, removing intermediate feedback reduces overall MBNS from 0.124 to 0.012 and BWR from 61.1\% to 46.7\%, and MAA decreases from 68.6\% to 64.4\%. Standard feedback yields higher MBNS in six task categories, with LLM as the only exception. These results provide evidence that access to intermediate outcomes improves sequential optimization for Qwen3-32B under the limited-budget protocol.

\paragraph{Intermediate Feedback Ablation.}
We isolate the contribution of intermediate experimental feedback by evaluating Qwen3-32B under an otherwise identical protocol. The baseline observation and all other task inputs remain available, but the metrics and execution logs produced by each intervention are withheld from subsequent decisions. As shown in Table~\ref{tab:no-feedback-ablation}, removing intermediate feedback reduces overall MBNS from 0.148 to 0.052 and BWR from 60.0\% to 50.0\%, while MAA decreases from 69.1\% to 65.2\%. Standard feedback yields higher MBNS in all seven task categories. These results provide evidence that access to intermediate outcomes improves sequential optimization for Qwen3-32B under the limited-budget protocol.

%% file: sections/conclusion.tex
\section{Conclusion}

In this work, we present AgentHPOBench, a benchmark for evaluating whether agents can improve executable ML experiments through sequential hyperparameter interventions. AgentHPOBench moves beyond conventional HPO benchmarks with clean black box objective interfaces by requiring agents to interpret metrics and logs from research repositories and convert this evidence into valid configurations. Experiments across 30 tasks show that current agents can improve reference baselines, but performance remains uneven across task domains, resource settings, and execution harnesses, and often remains below the reported reference performance. These findings establish HPO in research repositories as a challenging setting and provide a foundation for developing agents with stronger capabilities for experimental diagnosis and sequential decision making.

%% file: sections/appendix.tex
\section{Appendix}
This supplementary material provides additional details on the experimental protocol, task suite, reference baselines, and implementation settings. It also presents complete results, supplementary analyses, and qualitative examples that complement the findings in the main paper.

\section{AgentHPOBench Task Suite}
\label{app:tasks}

AgentHPOBench contains 30 tasks drawn from 30 distinct executable ML repositories associated with recent papers when available~\citep{drokin2024kolmogorov,luo2025agentlightning,woo2024moirai,das2024timesfm,warner2025smarter,jordan2024airbench,wang2024noisygl,knauer2024pmlbmini,kinakh2024binarydiffusion,beck2024xlstm,harrison2024vbll,gorishniy2025tabm,wang2024timexer,huang2024ablkit,luo2024classicgnn,sheng2024hybridflow,jolicoeurmartineau2023forestdiffusion,tian2024var,ji2024alignanything,ansari2024chronoslearninglanguagetime,bdeir2024hyperboliccv,liu2023itransformer,wang2023timemixer,huang2024rankup,lin2024sparsetsf}. Table~\ref{tab:appendix-task-suite} lists the full task suite. The anchor is taken from a paper, an official repository report, or a documented full setting reproduction.

\begin{table*}[t]
\centering
\scriptsize
\setlength{\tabcolsep}{2.2pt}
\resizebox{\textwidth}{!}{%
\begin{tabular}{cllllllc}
\toprule
\# & Repository & Task & Cat. & Dataset / Task & Metric & Direction & Anchor \\
\midrule
1 & \texttt{llm.c} & FineWeb Pretraining & NLP & FineWeb sample + HellaSwag & validation loss & lower & 3.425 \\
2 & \texttt{torch-conv-kan} & ConvKAN CIFAR-10 & CV & CIFAR-10 & accuracy & higher & 84.170 \\
3 & \texttt{open-r1} & Open-R1 MATH-500 & LLM & MATH-500 & exact match & higher & 83.100 \\
4 & \texttt{agent-lightning} & Room Selector Tuning & RL & APO room selector & validation accuracy & higher & 0.721 \\
5 & \texttt{uni2ts} & Uni2TS ETTh1 Forecasting & TS & ETTh1 LSF & forecast error & lower & 0.375 \\
6 & \texttt{timesfm} & TimesFM Long Horizon & TS & ETTh1 long horizon & WAPE & lower & 0.509 \\
7 & \texttt{ModernBERT} & ModernBERT MNLI & NLP & GLUE MNLI & accuracy & higher & 90.400 \\
8 & \texttt{cifar10-airbench} & AirBench CIFAR-10 & CV & CIFAR-10 & accuracy & higher & 94.010 \\
9 & \texttt{NoisyGL} & NoisyGL Cora GCN & Graph & Cora + 30\% label noise & accuracy & higher & 71.060 \\
10 & \texttt{TabMini} & TabMini Promoters & SL & molecular biology promoters & AUC & higher & 0.930 \\
11 & \texttt{binary-diffusion-tabular} & Adult Tabular Diffusion & SL & Adult & test accuracy & higher & 85.740 \\
12 & \texttt{xlstm} & xLSTM Parity & NLP & formal language Parity & scaled accuracy & higher & 1.000 \\
13 & \texttt{vbll} & VBLL Yacht Regression & SL & UCI Yacht & RMSE & lower & 0.860 \\
14 & \texttt{tabm} & TabM California Housing & SL & California Housing & RMSE & lower & 0.441 \\
15 & \texttt{TimeXer} & TimeXer PJM Forecasting & TS & EPF / PJM & MSE & lower & 0.093 \\
16 & \texttt{ABLkit} & ABLkit HWF Reasoning & SL & HWF & reasoning accuracy & higher & 99.200 \\
17 & \texttt{tunedGNN} & tunedGNN Cora GCN & Graph & Cora & accuracy & higher & 85.100 \\
18 & \texttt{verl} & verl GRPO GSM8K & LLM & GSM8K & accuracy & higher & 86.100 \\
19 & \texttt{ForestDiffusion} & ForestDiffusion Iris & SL & Iris & F1\_fake & higher & 0.970 \\
20 & \texttt{VAR} & VAR ImageNet 256 & CV & ImageNet-1K 256 & FID & lower & 3.550 \\
21 & \texttt{align-anything} & RAGEN Bandit Alignment & RL & RAGEN Bandit & success rate & higher & 1.000 \\
22 & \texttt{chronos-forecasting} & Chronos Weather Forecasting & TS & monash weather & WQL & lower & 0.148 \\
23 & \texttt{HyperbolicCV} & HyperbolicCV CIFAR-100 & CV & CIFAR-100 & accuracy & higher & 78.070 \\
24 & \texttt{iTransformer} & iTransformer ETTm2 Forecasting & TS & ETTm2 horizon 96 & MSE & lower & 0.180 \\
25 & \texttt{TimeMixer} & TimeMixer ETTm2 Forecasting & TS & ETTm2 horizon 96 & MSE & lower & 0.175 \\
26 & \texttt{ART} & ART 2048 & RL & 2048 game rollouts & win rate & higher & 0.600 \\
27 & \texttt{open-r1-multimodal} & Multimodal Open-R1 MathVista & LLM & MathVista-mini & accuracy & higher & 51.600 \\
28 & \texttt{semi-supervised-regression} & RankUp UTKFace Regression & CV & UTKFace & MAE & lower & 4.851 \\
29 & \texttt{SparseTSF} & SparseTSF ETTm1 Forecasting & TS & ETTm1 horizon 96 & MSE & lower & 0.314 \\
30 & \texttt{simpleRL-reason} & SimpleRL MATH-500 & LLM & MATH-500 & accuracy & higher & 34.400 \\
\bottomrule
\end{tabular}%
}
\caption{AgentHPOBench task suite. Each task is built from one executable repository and one scoreable target metric.}
\label{tab:appendix-task-suite}
\end{table*}

Table~\ref{tab:appendix-reference-baselines} reports the reference baseline performance used for scoring under the limited- and full-budget settings. The baseline configuration of each task is fixed across settings. The full-budget protocol expands the task-specific training budget where applicable. For pretrained inference tasks and tasks without a distinct scalable training stage, the evaluation protocol remains unchanged, so identical baseline values are expected. For example, ART 2048 uses the same fixed evaluation of 25 games with at most 70 moves per game in both settings, and its baseline heuristic wins none of these games, yielding a win rate of zero. For the full-budget study, each selected agent executes the fixed baseline configuration, and the median of the three observations is used as the common full-budget reference for that task. Values are reported in the metric units defined in Table~\ref{tab:appendix-task-suite} and rounded to three decimal places. All scores are computed from the corresponding full-precision values.

\begin{table*}[t]
\centering
\small
\setlength{\tabcolsep}{3pt}
\begin{tabular}{p{3.7cm}cc@{\hspace{10pt}}p{3.7cm}cc}
\toprule
Task & Limited & Full & Task & Limited & Full \\
\midrule
FineWeb Pretraining & 4.344 & 3.698
& ABLkit HWF Reasoning & 97.450 & 97.600 \\
ConvKAN CIFAR-10 & 28.970 & 35.500
& tunedGNN Cora GCN & 83.500 & 83.500 \\
Open-R1 MATH-500 & 1.563 & 2.200
& verl GRPO GSM8K & 84.685 & 84.230 \\
Room Selector Tuning & 0.414 & 0.379
& ForestDiffusion Iris & 0.901 & 0.901 \\
Uni2TS ETTh1 Forecasting & 1.155 & 1.144
& VAR ImageNet 256 & 70.713 & 70.713 \\
TimesFM Long Horizon & 0.551 & 0.551
& RAGEN Bandit Alignment & 0.168 & 0.178 \\
ModernBERT MNLI & 38.961 & 63.495
& Chronos Weather Forecasting & 0.372 & 0.372 \\
AirBench CIFAR-10 & 89.090 & 93.680
& HyperbolicCV CIFAR-100 & 16.390 & 56.630 \\
NoisyGL Cora GCN & 70.000 & 70.000
& iTransformer ETTm2 Forecasting & 0.183 & 0.183 \\
TabMini Promoters & 0.928 & 0.928
& TimeMixer ETTm2 Forecasting & 0.178 & 0.178 \\
Adult Tabular Diffusion & 74.106 & 85.519
& ART 2048 & 0.000 & 0.000 \\
xLSTM Parity & 0.011 & 0.067
& Multimodal Open-R1 MathVista & 48.800 & 48.800 \\
VBLL Yacht Regression & 1.487 & 0.467
& RankUp UTKFace Regression & 34.542 & 6.427 \\
TabM California Housing & 0.536 & 0.533
& SparseTSF ETTm1 Forecasting & 0.341 & 0.340 \\
TimeXer PJM Forecasting & 0.133 & 0.114
& SimpleRL MATH-500 & 15.800 & 5.200 \\
\bottomrule
\end{tabular}
\caption{Reference baseline performance $y_{t,0}^{(r)}$ under the limited and full budget settings. Each value is reported in the original metric unit of the corresponding task.}
\label{tab:appendix-reference-baselines}
\end{table*}

\section{Agent Prompt Template}
\label{app:prompt-template}

AgentHPOBench standardizes the agent interface through a structured decision schema. At each intervention step, the task script constructs a task-specific context block that contains the task description, target metric, metric direction, current configuration, allowed search space, paper or repository anchor, baseline result, and previous trial history when available. This context is then passed to the decision backend, and the returned intervention is parsed into a structured configuration before execution. Figure~\ref{fig:prompt-local-open-weight} shows the shared prompt template used for local open-weight agents, while Figure~\ref{fig:prompt-api} shows the compact decision prompt used for API-based agents. Figure~\ref{fig:prompt-task-specific} provides an example task-specific context, illustrating how concrete trial evidence and the allowed search space are supplied before requesting the next intervention. We use ``search space'' to denote the task-specific allowed intervention space, namely the configurable fields exposed to the agent and the valid values for each field, rather than the decoding hyperparameters of the agent.

\begin{figure*}[t]
\centering
\begin{promptlisting}{Local open-weight agent prompt}
You are an expert machine learning researcher helping optimize hyperparameters for model training.

## Task
Optimize hyperparameters to {optimization_goal} the target metric within the remaining budget.
- Metric Direction: {metric_direction}

## Current Training Progress
- Current Epoch: {epoch}
- Budget Remaining: {budget_remaining} epochs out of {total_budget} total
- Decision Number: {decision_number}

## Recent Training Logs
{training_log}

## Current Hyperparameters
{current_config}

## Hyperparameter Search Space
{search_space}

## Instructions
- Be conservative unless the run is clearly plateauing or unstable.
- Only return values inside the search space.
- If training is improving normally, keeping the current config is acceptable.
- Output plain text only.
- Do not use markdown code fences, XML tags, or extra sections.
- Follow the exact headers below.

## Response Format
REASONING: <short explanation>
NEW_CONFIG: {"learning_rate": 0.05, "batch_size": 128, "optimizer": "sgd"}
CONFIDENCE: 0.8
\end{promptlisting}
\caption{Prompt template used for local open-weight agents. The optimization goal is set to \texttt{maximize} for higher-is-better metrics and \texttt{minimize} for lower-is-better metrics.}
\label{fig:prompt-local-open-weight}
\end{figure*}

\begin{figure*}[!t]
\centering
\begin{promptlisting}{API-based agent prompt}
Task: choose the next hyperparameter config for {task_name}.
Return exactly one JSON object containing only proposed hyperparameter keys and values.

Current config:
{current_config}

Search space:
{search_space}

Budget remaining: {budget_remaining} of {total_budget}.

Training history:
{training_log}
\end{promptlisting}
\caption{Compact decision prompt used for API-based agents.}
\label{fig:prompt-api}
\end{figure*}

\begin{figure*}[!t]
\centering
\begin{promptlisting}{Example task-specific context}
You are choosing one hyperparameter configuration for a budget-limited reproduction benchmark.

Task: SparseTSF ETTm1 multivariate long-term forecasting, pred_len=96.
Metric: test_mse_at_best_val over the fixed epoch budget, lower is better.
Paper anchor: ICML 2024 SparseTSF official long-term forecasting table reports ETTm1 horizon-96 MSE=0.314.

Budget baseline:
test_mse_at_best_val=0.340816, config={...}

Trial history:
Trial 0 baseline:
config={...}
test_mse=0.340816
val_mse=0.440882

Trial 1 intervention_1:
config={...}
test_mse=0.340816
val_mse=0.440882

Trial 2 intervention_2:
config={...}
test_mse=0.318475
val_mse=0.407163

Search space:
{...}

Return exactly one line:
NEW_CONFIG: {"lr": 0.02, "weight_decay": 0.0, "batch_size": 256, ...}

Choose the config most likely to reduce test_mse_at_best_val under the fixed budget. Do not change epochs, data split, dataset, seq_len, pred_len, model family, or metric.
\end{promptlisting}
\caption{Example task-specific context used to request the next intervention.}
\label{fig:prompt-task-specific}
\end{figure*}

The local and API templates preserve the interfaces used by their respective agent implementations rather than enforcing identical surface wording. Prompt construction, response parsing, configuration validation, retries, and other harness behavior are therefore part of the evaluated agent system. The resulting comparisons should be interpreted as comparisons between complete agent pipelines, not as isolated rankings of the underlying language models.

\section{Implementation and Reproducibility Details}
\label{app:implementation-reproducibility}

\paragraph{Common execution protocol.}
Unless stated otherwise, all results use the limited-budget protocol. Each task first executes a fixed reference baseline and then permits five sequential interventions. The baseline and every intervention use the same task-specific data split, metric, intervention space, and execution budget. The limited setting uses approximately \(10\%\) of the corresponding original training or evaluation budget. The full-budget study changes only this execution budget and retains the task definition, prompt, intervention space, five-decision protocol, result schema, and scoring pipeline. Each decision must return one complete configuration in the structured \texttt{NEW\_CONFIG} format. Omitted fields retain their current values, and configurations are validated and clamped to the task-specific discrete intervention space before execution. The reported task result is the metric after intervention five, rather than the best intermediate metric.

The benchmark optimizes the objective exposed by each upstream repository. Some repositories report a test-set metric, or a test metric at the checkpoint selected by validation performance, and this repository-defined objective is visible during sequential decision making. AgentHPOBench therefore evaluates optimization of an observable experimental objective. It does not provide a separate hidden test set and should not be interpreted as estimating generalization after adaptive model selection.

\paragraph{Open-weight decision models.}
Table~\ref{tab:appendix-open-weight-settings} lists the exact checkpoints and realized generated-token usage for the six open-weight agents under the limited-budget protocol. We load all checkpoints with Hugging Face Transformers using the checkpoint's chat template and \texttt{bfloat16} weights. Agent decoding is deterministic: temperature is \(0\), sampling is disabled, and optional thinking output is disabled so that generation begins with the requested structured decision. Token counts are computed from the generated decision text using the tokenizer associated with each checkpoint. They exclude input-prompt tokens. Each checkpoint contributes 450 decisions from 30 tasks, five interventions, and three experiment seeds.

\begin{table*}[t]
\centering
\small
\setlength{\tabcolsep}{5pt}
\begin{tabular}{lccccc}
\toprule
Hugging Face checkpoint & Decisions & Total tokens & Mean & Median & P95 \\
\midrule
\texttt{Qwen/Qwen3-8B} & 450 & 60{,}636 & 134.7 & 127.0 & 383.5 \\
\texttt{google/gemma-2-2b-it} & 450 & 40{,}857 & 90.8 & 85.5 & 142.5 \\
\texttt{meta-llama/Meta-Llama-3.1-8B-Instruct} & 450 & 53{,}700 & 119.3 & 123.0 & 209.0 \\
\texttt{deepseek-ai/DeepSeek-R1-Distill-Qwen-14B} & 450 & 153{,}673 & 341.5 & 130.0 & 1{,}380.2 \\
\texttt{microsoft/Phi-4} & 450 & 91{,}327 & 202.9 & 214.0 & 338.5 \\
\texttt{Qwen/Qwen3-32B} & 450 & 54{,}771 & 121.7 & 125.0 & 191.5 \\
\bottomrule
\end{tabular}
\caption{Realized generated-token usage of the open-weight decision models under the limited-budget protocol. P95 denotes the 95th percentile across individual decisions.}
\label{tab:appendix-open-weight-settings}
\end{table*}

\paragraph{API-based agents.}
We evaluate DeepSeek-V4-Pro, GPT-5.5, GLM-4.7, GLM-5.1, Kimi-2.6, and Claude Sonnet 4.6 using the same five logical decisions, task contexts, output schema, and configuration validation as the open-weight agents. The request temperature is set to \(0\) when the endpoint exposes this control. For every decision, the result trace records the provider and model identifier returned by the harness, the raw response, timestamp, token-usage fields when available, and retry or error metadata. Each API agent is evaluated with one complete, timestamped, audited 30-task run. This choice is not presented as a controlled seed replicate: hosted endpoints do not expose an immutable checkpoint build, serving replica, batching state, or a reproducible end-to-end random seed, and these service-side states may change independently of the benchmark. Repeating an API request would therefore measure a mixture of model and serving changes rather than the benchmark stochasticity isolated by the controlled local runs. Table~\ref{tab:appendix-api-usage} reports the provider-recorded usage of the API evaluation campaign.

\begin{table}[t]
\centering
\small
\setlength{\tabcolsep}{4pt}
\begin{tabular}{lccc}
\toprule
API agent & Requests & Total tokens & Tokens/request \\
\midrule
Claude Sonnet 4.6 & 1{,}695 & 2{,}498{,}302 & 1{,}474 \\
GPT-5.5 & 1{,}418 & 2{,}132{,}497 & 1{,}504 \\
DeepSeek-V4-Pro & 1{,}321 & 2{,}000{,}752 & 1{,}515 \\
GLM-5.1 & 1{,}296 & 1{,}978{,}744 & 1{,}527 \\
Kimi-2.6 & 1{,}237 & 1{,}922{,}535 & 1{,}554 \\
GLM-4.7 & 1{,}231 & 1{,}866{,}319 & 1{,}516 \\
\bottomrule
\end{tabular}
\caption{Provider-recorded API usage during the limited-budget evaluation campaign. Request counts include retries and harness-level calls in addition to the five logical decisions per task.}
\label{tab:appendix-api-usage}
\end{table}

Each API agent makes 150 accepted logical decisions across the 30 tasks, while the campaign records 1,231--1,695 requests, or approximately 8.2--11.3 requests per accepted decision. The request totals include retries, parsing or validation recovery, and other harness-level calls and therefore are not additional intervention opportunities. Provider-recorded token counts include both prompt and generated tokens and are not directly comparable with the generated-output-only counts in Table~\ref{tab:appendix-open-weight-settings}.

\paragraph{Seeds, repetitions, and aggregation.}
We run every open-weight agent and each conventional HPO baseline with experiment seeds \(s\in\{0,1,42\}\). The same three seeds are used consistently across the benchmark harness, optimizer, and stochastic task execution, while dataset splits fixed by an upstream protocol remain unchanged. Random search, TPE, and the fixed-budget BOHB-style method use the same reference baseline, discrete intervention space, and five configuration evaluations as the agents. Their proposal rules are detailed in Appendix~\ref{app:conventional-hpo}.

For each controlled method, the bounded normalized score, baseline win indicator, and anchor attainment are computed independently for every task and seed. Category and overall metrics are then computed for each seed and reported as the arithmetic mean across the three seeds. Table~\ref{tab:appendix-multiseed-results} additionally reports the corresponding sample standard deviation. This preserves equal weight for every benchmark task. The local agent checkpoints, decoding settings, runtime environment, and task random states are explicitly controlled, so these repetitions quantify sensitivity to benchmark execution stochasticity rather than provider-side variation.

\paragraph{No intermediate feedback ablation.}
The no intermediate feedback condition is paired with the standard feedback run at the task and replicate-seed level. It uses the exact baseline configuration, target metric, and baseline observation recorded by the corresponding standard-feedback run rather than independently re-evaluating the baseline. The baseline metric is therefore available to the agent. However, for all five decisions, the visible history contains only this fixed baseline observation. Metrics, auxiliary outputs, and logs produced by interventions one through four are withheld. The proposed interventions are still executed and recorded normally. Thus, the ablation isolates access to intermediate experimental evidence while holding the initial observation, evaluator seed, task budget, search space, and decision model fixed.

\paragraph{Execution environment and reproducibility resources.}
The controlled experiments run on Linux development machines equipped with NVIDIA H200 GPUs with \(143{,}771\) MiB of visible memory. The orchestration layer uses Python 3.10 or newer. Because the 30 tasks depend on heterogeneous upstream repositories, each task adapter invokes its repository-specific Conda environment rather than forcing all tasks into one dependency stack. Model weights and datasets are downloaded before execution, and the reported runs use offline Hugging Face modes.

For reproducibility, we record the upstream repository version, benchmark adaptations, environment and asset-preparation requirements, model checkpoint, task seeds, baseline configuration, metric extraction rule, intervention space, execution budget, and scoring reference for every task. Detailed setup instructions and machine-readable task specifications will be provided in the public GitHub repository.

\paragraph{Result Validation.}
We include only task runs that complete the reference baseline and all five interventions, with a valid configuration and metric recorded at every step. Interrupted or malformed runs are excluded and rerun at the task level. All reported aggregates are computed from complete task records. The public GitHub repository includes the corresponding validation and aggregation utilities.

\section{Scoring and Aggregation Details}
\label{app:scoring-details}

For a fixed budget setting $r$, let $y_{t,0}^{(r)}$ denote the reference baseline performance for task $t$ reported in Table~\ref{tab:appendix-reference-baselines}, let $y_{t,k}^{(r)}$ denote the result after intervention $k$, and let $a_t$ denote the paper or repository anchor. The main task result is $y_{t,K}^{(r)}$, which is obtained after the final intervention rather than selected as the best intermediate result. This convention measures whether an optimizer preserves or refines its improvements after observing the complete trajectory.

Before scoring, every baseline, intervention result, and anchor is converted to a common numeric unit within its task. For example, an accuracy represented as a fraction is converted to a percentage when the corresponding anchor is reported as a percentage. Let $g_t(z)=z$ for a metric in which higher values are better and $g_t(z)=-z$ for a metric in which lower values are better. We define the normalized score as
\begin{equation}
\mathrm{NS}_{t,r}=
\frac{g_t\!\left(y_{t,K}^{(r)}\right)-g_t\!\left(y_{t,0}^{(r)}\right)}
{g_t(a_t)-g_t\!\left(y_{t,0}^{(r)}\right)}.
\end{equation}
A value of $0$ matches the reference baseline, a value of $1$ matches the anchor, and a negative value indicates degradation. Because a small baseline to anchor gap can give one task disproportionate influence, we bound each task score before aggregation:
\begin{equation}
\mathrm{BNS}_{t,r}
=
\min\!\left\{1,\max\!\left\{-1,\mathrm{NS}_{t,r}\right\}\right\}.
\end{equation}
For the task set $\mathcal{T}$, the mean bounded normalized score is
\begin{equation}
\mathrm{MBNS}_{r}
=
\frac{1}{|\mathcal{T}|}
\sum_{t\in\mathcal{T}}\mathrm{BNS}_{t,r}.
\end{equation}

The baseline win rate reports the percentage of tasks for which the final result strictly improves over the reference baseline:
\begin{equation}
\mathrm{BWR}_{r}
=
\frac{100}{|\mathcal{T}|}
\sum_{t\in\mathcal{T}}
\mathbb{I}\!\left[
g_t\!\left(y_{t,K}^{(r)}\right)
>
g_t\!\left(y_{t,0}^{(r)}\right)
\right].
\end{equation}
Ties are not counted as wins. Anchor attainment measures absolute performance relative to the anchor in the original metric direction:
\begin{equation}
\mathrm{AA}_{t,r}
=
\begin{cases}
y_{t,K}^{(r)}/a_t, & \text{if higher values are better},\\
a_t/y_{t,K}^{(r)}, & \text{if lower values are better}.
\end{cases}
\end{equation}
The mean anchor attainment is
\begin{equation}
\mathrm{MAA}_{r}
=
\frac{100}{|\mathcal{T}|}
\sum_{t\in\mathcal{T}}\mathrm{AA}_{t,r}.
\end{equation}
Thus, MBNS measures bounded improvement over the common reference baseline, BWR measures the coverage of positive improvements, and MAA measures absolute attainment of reported reference performance. MAA may exceed $100\%$ when a result surpasses its anchor.

The reference baseline configuration is fixed across budget settings, but its measured performance can change with the execution budget. Within each budget setting, the same $y_{t,0}^{(r)}$ is used for all agents and conventional optimizers. For the intervention level tables below, $\mathrm{BNS}_{t,r,k}$ and $\mathrm{AA}_{t,r,k}$ are computed by replacing $y_{t,K}^{(r)}$ with $y_{t,k}^{(r)}$ in the definitions above.

\section{Additional Experimental Protocols and Results}
\label{app:additional-experiments}

This section provides implementation details and supporting results for the conventional HPO baselines and the intermediate feedback ablation reported in the main paper. All methods use the task-specific intervention spaces defined by the benchmark task specifications, begin from the same limited-budget reference baseline, receive five intervention opportunities, and are scored using the result after the fifth intervention. Table~\ref{tab:appendix-task-suite} summarizes the corresponding tasks, objectives, metric directions, and anchors.

\subsection{Conventional HPO Baselines}
\label{app:conventional-hpo}

\paragraph{Implementation.}
Random search samples every configurable field independently and uniformly from its allowed discrete values. We evaluate random search with seeds $0$, $1$, and $42$.

The TPE baseline uses the observations available before each intervention to rank previous configurations in the direction of the target metric. Proposals are sampled uniformly until at least two executed observations are available. TPE then assigns the best $\lceil 0.35n\rceil$ of the $n$ observations to the good set and samples each field according to its smoothed good to bad frequency ratio. The additive smoothing constant is $1.0$, and a proposal uses uniform exploration with probability $0.15$ for each field.

Because every intervention in the limited budget protocol receives the same training budget, standard multi fidelity resource allocation is not available. We therefore implement a fixed budget adaptation of BOHB. Proposals are sampled uniformly until at least two executed observations are available. Later proposals rank the observations, retain the best $\lceil n/3\rceil$ as the current elite set, select a parent with rank based weights, and mutate each field with probability $0.35$. A mutated field is sampled from the smoothed good to bad frequency ratio with probability $0.5$ and uniformly otherwise. Random search, TPE, and this BOHB adaptation are each evaluated with seeds $\{0,1,42\}$. Thus, the BOHB result evaluates its configuration selection policy under the common five intervention protocol, rather than the resource allocation component of standard BOHB.

The conventional optimizers operate on configurations and scalar target values, while the agents additionally process the task description and experimental feedback through their native decision interface. Accordingly, these baselines compare complete optimization systems under a common execution budget. All methods are evaluated under the same five-intervention execution budget. Because every trial receives the same resource budget, the BOHB-style baseline evaluates its configuration-proposal component in this setting. Multi-fidelity resource scheduling is outside the scope of the protocol.

\paragraph{Random seed sensitivity.}
Table~\ref{tab:appendix-multiseed-results} reports the variability of all locally executable methods. Each entry is the arithmetic mean and sample standard deviation of the corresponding seed level metric over seeds $\{0,1,42\}$.

\begin{table*}[!t]
\centering
\scriptsize
\setlength{\tabcolsep}{2.0pt}
\resizebox{0.99\textwidth}{!}{%
\begin{tabular}{lcccccccccc}
\toprule
\multirow{2}{*}{\textbf{Agent / Method}}
& \multicolumn{8}{c}{\textbf{Mean bounded normalized score}}
& \multicolumn{2}{c}{\textbf{Overall metrics}} \\
\cmidrule(lr){2-9}
\cmidrule(lr){10-11}
& NLP (3) & CV (5) & TS (7) & Graph (2) & RL (3) & LLM (4) & SL (6)
& Overall & BWR (\%) & MAA (\%) \\
\midrule
\multicolumn{11}{l}{\textit{Conventional HPO baselines}} \\
\midrule
Random search
& \(0.034 \pm 0.039\) & \(-0.135 \pm 0.075\) & \(0.299 \pm 0.023\)
& \(-0.463 \pm 0.556\) & \(0.124 \pm 0.192\) & \(-0.292 \pm 0.102\)
& \(-0.136 \pm 0.097\) & \(-0.034 \pm 0.051\) & \(48.9 \pm 10.2\) & \(62.6 \pm 0.3\) \\
TPE
& \(0.020 \pm 0.006\) & \(-0.110 \pm 0.105\) & \(0.005 \pm 0.101\)
& \(-0.406 \pm 0.526\) & \(0.192 \pm 0.184\) & \(-0.219 \pm 0.029\)
& \(-0.302 \pm 0.161\) & \(-0.113 \pm 0.068\) & \(40.0 \pm 3.3\) & \(62.4 \pm 1.8\) \\
BOHB variant
& \(-0.062 \pm 0.011\) & \(-0.006 \pm 0.194\) & \(0.235 \pm 0.111\)
& \(-0.824 \pm 0.305\) & \(0.260 \pm 0.127\) & \(-0.236 \pm 0.025\)
& \(0.153 \pm 0.348\) & \(0.018 \pm 0.056\) & \(45.6 \pm 7.7\) & \(65.3 \pm 2.1\) \\
\midrule
\multicolumn{11}{l}{\textit{Open-weight agents}} \\
\midrule
Gemma2-2B
& \(0.014 \pm 0.011\) & \(-0.103 \pm 0.130\) & \(0.233 \pm 0.088\)
& \(-0.459 \pm 0.505\) & \(0.005 \pm 0.293\) & \(-0.171 \pm 0.128\)
& \(0.066 \pm 0.059\) & \(-0.001 \pm 0.015\) & \(55.6 \pm 5.1\) & \(64.2 \pm 2.5\) \\
DeepSeek-R1-Qwen-14B
& \(0.008 \pm 0.014\) & \(-0.223 \pm 0.061\) & \(0.316 \pm 0.101\)
& \(-0.333 \pm 0.577\) & \(0.074 \pm 0.045\) & \(-0.271 \pm 0.082\)
& \(0.157 \pm 0.174\) & \(0.018 \pm 0.035\) & \(54.4 \pm 5.1\) & \(64.1 \pm 0.8\) \\
Qwen3-8B
& \(-0.001 \pm 0.050\) & \(-0.140 \pm 0.048\) & \(0.233 \pm 0.076\)
& \(-0.271 \pm 0.638\) & \(0.158 \pm 0.042\) & \(-0.219 \pm 0.109\)
& \(0.124 \pm 0.171\) & \(0.024 \pm 0.069\) & \(53.3 \pm 3.3\) & \(65.1 \pm 1.0\) \\
Llama-3.1-8B
& \(0.019 \pm 0.143\) & \(-0.142 \pm 0.147\) & \(0.253 \pm 0.051\)
& \(-0.525 \pm 0.463\) & \(0.125 \pm 0.238\) & \(-0.144 \pm 0.218\)
& \(0.172 \pm 0.190\) & \(0.030 \pm 0.033\) & \(44.4 \pm 10.2\) & \(66.3 \pm 2.0\) \\
Phi-4-14B
& \(0.047 \pm 0.032\) & \(-0.077 \pm 0.138\) & \(0.278 \pm 0.013\)
& \(-0.448 \pm 0.508\) & \(0.306 \pm 0.233\) & \(-0.098 \pm 0.107\)
& \(0.430 \pm 0.075\) & \(0.130 \pm 0.062\) & \(63.3 \pm 3.3\) & \(66.9 \pm 1.9\) \\
Qwen3-32B
& \(0.060 \pm 0.054\) & \(-0.120 \pm 0.104\) & \(0.285 \pm 0.014\)
& \(-0.302 \pm 0.606\) & \(0.349 \pm 0.037\) & \(-0.120 \pm 0.060\)
& \(0.485 \pm 0.175\) & \(0.148 \pm 0.059\) & \(60.0 \pm 3.3\) & \(69.1 \pm 0.6\) \\
\bottomrule
\end{tabular}%
}
\caption{Three-seed results for conventional HPO baselines and open-weight agents under the limited-budget protocol. Entries report mean \(\pm\) sample standard deviation over seeds \(\{0,1,42\}\). Category columns report MBNS. BWR and MAA are computed independently for each seed before aggregation.}
\label{tab:appendix-multiseed-results}
\end{table*}

\subsection{Statistical Robustness}
\label{app:statistical-robustness}

We additionally assess sensitivity to the composition of the benchmark task suite using 20,000 category-stratified paired bootstrap resamples. Each resample preserves the number of tasks in every research category, and the same sampled tasks are used for all methods. For open-weight agents and conventional HPO baselines, the task-level BNS and anchor attainment values are first averaged over the three controlled seeds and then resampled. The resulting intervals therefore quantify uncertainty associated with task composition, whereas the standard deviations in Table~\ref{tab:appendix-multiseed-results} quantify variation across controlled executions. For API agents, which are evaluated once, these intervals reflect sensitivity to benchmark task composition only and should not be interpreted as uncertainty across repeated API executions.

\paragraph{Interval definitions.}
A task-bootstrap 95\% confidence interval is obtained by repeatedly resampling tasks with replacement within each research category, recomputing the aggregate metric for each resample, and taking the 2.5th and 97.5th percentiles of the resulting distribution. It measures the sensitivity of an aggregate result to the composition of the benchmark task suite. A paired MBNS difference is computed as \(\mathrm{MBNS}(A)-\mathrm{MBNS}(B)\) using the same resampled tasks for both methods. A positive difference favors Method A, while a negative difference favors Method B. If its 95\% confidence interval includes zero, the observed ordering is not stable under variation in task composition.

\begin{table*}[!t]
\centering
\small
\setlength{\tabcolsep}{6.0pt}
\begin{tabular}{lcccc}
\toprule
\textbf{Agent / Method}
& \textbf{MBNS [95\% CI]}
& \textbf{MAA (\%) [95\% CI]}
& \textbf{Median BNS}
& \textbf{BWR (\%)} \\
\midrule
\multicolumn{5}{l}{\textit{Conventional HPO baselines}} \\
\midrule
Random search
& \(-0.034~[-0.153,\,0.082]\) & \(62.6~[52.5,\,72.2]\) & \(0.009\) & \(48.9\) \\
TPE
& \(-0.113~[-0.265,\,0.045]\) & \(62.4~[51.9,\,72.3]\) & \(-0.030\) & \(40.0\) \\
BOHB variant
& \(0.018~[-0.122,\,0.161]\) & \(65.3~[55.5,\,74.5]\) & \(0.012\) & \(45.6\) \\
\midrule
\multicolumn{5}{l}{\textit{Open-weight agents}} \\
\midrule
Gemma2-2B
& \(-0.001~[-0.154,\,0.143]\) & \(64.2~[55.0,\,73.2]\) & \(0.003\) & \(55.6\) \\
DeepSeek-R1-Qwen-14B
& \(0.018~[-0.141,\,0.169]\) & \(64.1~[54.1,\,73.9]\) & \(0.007\) & \(54.4\) \\
Llama-3.1-8B
& \(0.030~[-0.118,\,0.183]\) & \(66.3~[57.3,\,75.0]\) & \(-0.041\) & \(44.4\) \\
Qwen3-8B
& \(0.024~[-0.127,\,0.172]\) & \(65.1~[55.8,\,74.2]\) & \(-0.010\) & \(53.3\) \\
Phi-4-14B
& \(0.130~[0.001,\,0.256]\) & \(66.9~[57.8,\,75.9]\) & \(0.066\) & \(63.3\) \\
Qwen3-32B
& \(0.148~[0.016,\,0.283]\) & \(69.1~[60.3,\,77.7]\) & \(0.058\) & \(60.0\) \\
\midrule
\multicolumn{5}{l}{\textit{API agents}} \\
\midrule
GLM-5.1
& \(0.080~[-0.084,\,0.240]\) & \(67.0~[57.4,\,76.4]\) & \(0.036\) & \(56.7\) \\
Kimi-2.6
& \(0.110~[-0.054,\,0.260]\) & \(70.4~[61.0,\,79.8]\) & \(0.012\) & \(56.7\) \\
GLM-4.7
& \(0.112~[-0.049,\,0.268]\) & \(67.8~[58.2,\,77.2]\) & \(0.076\) & \(60.0\) \\
DeepSeek-V4-Pro
& \(0.175~[-0.017,\,0.352]\) & \(70.1~[61.6,\,78.5]\) & \(0.109\) & \(63.3\) \\
GPT-5.5
& \(0.305~[0.174,\,0.433]\) & \(76.7~[66.7,\,86.7]\) & \(0.103\) & \(66.7\) \\
Claude Sonnet 4.6
& \(\mathbf{0.407~[0.264,\,0.541]}\)
& \(\mathbf{79.5~[66.6,\,93.6]}\)
& \(\mathbf{0.382}\) & \(\mathbf{76.7}\) \\
\bottomrule
\end{tabular}
\caption{Task-composition robustness of the evaluated methods under the limited-budget protocol. Intervals are percentile 95\% confidence intervals from 20,000 category-stratified paired bootstrap resamples of the 30 tasks. Median BNS and BWR are point estimates.}
\label{tab:appendix-bootstrap-robustness}
\end{table*}

The bootstrap intervals show that benchmark composition contributes non-negligible uncertainty, particularly for methods whose gains are concentrated in a small number of categories or tasks. Claude Sonnet 4.6 retains the strongest point estimates across the four reported summaries. Qwen3-32B and Phi-4-14B have similar task-composition uncertainty among open-weight agents. The three conventional HPO intervals are also broad under the final-step criterion. These intervals are intended as a robustness diagnostic rather than a multiple-comparison significance test. Small differences between methods should therefore be interpreted together with the paired comparisons reported below.

\subsection{Best-So-Far Performance}
\label{app:best-so-far}

The main results use the configuration produced at the fifth intervention, which evaluates whether a method preserves and refines improvements throughout the full trajectory. As a complementary diagnostic, Table~\ref{tab:appendix-best-so-far} reports the best result observed among interventions one through five. For each task, the best intervention according to the task-specific metric direction is scored against the same common reference baseline used in the main results. The reference baseline itself is not included among the candidate interventions. For methods evaluated with three seeds, scoring is performed independently for each task and seed before averaging across seeds and tasks, matching the aggregation used in the main results.

This analysis distinguishes the quality of the best configuration discovered within the intervention budget from the ability to retain it at the final step. Claude Sonnet 4.6 remains the strongest method overall, reaching an MBNS of 0.469, a BWR of 90.0\%, and an MAA of 80.5\%. It also leads on CV, Graph, LLM, and SL, while GPT-5.5 leads on RL, TPE on TS, and random search on NLP. Compared with their final-step results, the overall MBNS of the three conventional HPO baselines improves by 0.273--0.411 under best-so-far selection, indicating that they often discover useful configurations but do not consistently finish with them.

Under this incumbent-style view, random search, TPE, and the BOHB variant obtain overall MBNS values of 0.325, 0.298, and 0.291, respectively, exceeding all evaluated open-weight agents. Phi-4-14B and Qwen3-32B reach 0.197 and 0.195. This ranking reversal clarifies that the final-step metric combines search quality with the ability to preserve or refine an earlier gain, whereas standard HPO commonly returns the incumbent. The strongest API agents remain competitive under best-so-far selection, with Claude Sonnet 4.6 and GPT-5.5 reaching 0.469 and 0.347. The corresponding seed variation and task-bootstrap intervals are reported below.

\begin{table*}[!t]
\centering
\small
\setlength{\tabcolsep}{2.8pt}
\resizebox{0.98\textwidth}{!}{%
\begin{tabular}{lcccccccccc}
\toprule
\multirow{2}{*}{\textbf{Agent / Method}}
& \multicolumn{8}{c}{\textbf{Mean bounded normalized score}}
& \multicolumn{2}{c}{\textbf{Overall metrics}} \\
\cmidrule(lr){2-9}
\cmidrule(lr){10-11}
& NLP (3) & CV (5) & TS (7) & Graph (2) & RL (3) & LLM (4) & SL (6)
& Overall & BWR (\%) & MAA (\%) \\
\midrule
\multicolumn{11}{l}{\textit{Conventional HPO baselines}} \\
\midrule
Random search
& \textbf{0.089} & 0.179 & 0.629 & 0.011 & 0.481 & -0.097 & 0.515
& 0.325 & 71.1 & 72.7 \\
TPE
& 0.088 & 0.138 & \textbf{0.682} & -0.192 & 0.426 & -0.090 & 0.447
& 0.298 & 74.4 & 71.4 \\
BOHB variant
& 0.015 & 0.225 & 0.570 & -0.167 & 0.305 & -0.061 & 0.538
& 0.291 & 67.8 & 70.4 \\
\midrule
\multicolumn{11}{l}{\textit{Open-weight agents}} \\
\midrule
Gemma2-2B
& 0.020 & 0.170 & 0.259 & -0.459 & 0.125 & -0.091 & 0.142
& 0.089 & 62.2 & 66.3 \\
DeepSeek-R1-Qwen-14B
& 0.018 & 0.051 & 0.320 & -0.333 & 0.404 & -0.206 & 0.181
& 0.112 & 63.3 & 67.1 \\
Qwen3-8B
& 0.016 & 0.123 & 0.303 & -0.271 & 0.194 & -0.038 & 0.180
& 0.125 & 63.3 & 66.7 \\
Llama-3.1-8B
& 0.019 & 0.045 & 0.291 & -0.177 & 0.299 & -0.035 & 0.215
& 0.134 & 56.7 & 68.7 \\
Qwen3-32B
& 0.074 & 0.070 & 0.289 & -0.302 & 0.375 & -0.079 & 0.508
& 0.195 & 66.7 & 70.4 \\
Phi-4-14B
& 0.050 & 0.123 & 0.282 & -0.281 & 0.413 & -0.064 & 0.459
& 0.197 & 70.0 & 69.0 \\
\midrule
\multicolumn{11}{l}{\textit{API agents}} \\
\midrule
GLM-5.1
& -0.088 & 0.171 & 0.286 & 0.031 & 0.032 & 0.009 & 0.713
& 0.235 & 70.0 & 69.1 \\
Kimi-2.6
& -0.050 & 0.122 & 0.277 & 0.031 & 0.486 & -0.052 & 0.684
& 0.261 & 70.0 & 71.4 \\
GLM-4.7
& -0.088 & 0.197 & 0.283 & 0.000 & 0.061 & -0.044 & 0.689
& 0.228 & 66.7 & 69.6 \\
DeepSeek-V4-Pro
& -0.027 & 0.096 & 0.510 & 0.094 & 0.621 & 0.052 & 0.544
& 0.316 & 73.3 & 72.8 \\
GPT-5.5
& -0.028 & 0.164 & 0.375 & 0.281 & \textbf{0.822} & 0.119 & 0.589
& 0.347 & 73.3 & 77.5 \\
Claude Sonnet 4.6
& -0.013 & \textbf{0.258} & 0.519 & \textbf{0.877} & 0.710
& \textbf{0.206} & \textbf{0.746} & \textbf{0.469} & \textbf{90.0} & \textbf{80.5} \\
\bottomrule
\end{tabular}%
}
\caption{Best-so-far AgentHPOBench results under the limited-budget protocol. For each task, the best observed result among interventions one through five is selected and scored. Category columns report MBNS, while BWR and MAA are computed over all 30 tasks. This diagnostic complements the final-step results in the main paper.}
\label{tab:appendix-best-so-far}
\end{table*}

\begin{table*}[!t]
\centering
\small
\setlength{\tabcolsep}{7pt}
\begin{tabular}{lccc}
\toprule
\textbf{Agent / Method}
& \textbf{Best-so-far MBNS}
& \textbf{Seed SD}
& \textbf{Task-bootstrap 95\% CI} \\
\midrule
\multicolumn{4}{l}{\textit{Conventional HPO baselines}} \\
\midrule
Random search & \(0.325\) & \(0.082\) & \([0.199,\,0.447]\) \\
TPE & \(0.298\) & \(0.048\) & \([0.179,\,0.417]\) \\
BOHB variant & \(0.291\) & \(0.040\) & \([0.187,\,0.396]\) \\
\midrule
\multicolumn{4}{l}{\textit{Open-weight agents}} \\
\midrule
Gemma2-2B & \(0.089\) & \(0.008\) & \([-0.071,\,0.241]\) \\
DeepSeek-R1-Qwen-14B & \(0.112\) & \(0.052\) & \([-0.010,\,0.234]\) \\
Qwen3-8B & \(0.125\) & \(0.061\) & \([-0.016,\,0.262]\) \\
Llama-3.1-8B & \(0.134\) & \(0.028\) & \([0.003,\,0.267]\) \\
Phi-4-14B & \(0.197\) & \(0.039\) & \([0.065,\,0.327]\) \\
Qwen3-32B & \(0.195\) & \(0.060\) & \([0.066,\,0.327]\) \\
\midrule
\multicolumn{4}{l}{\textit{API agents}} \\
\midrule
GLM-5.1 & \(0.235\) & -- & \([0.086,\,0.381]\) \\
Kimi-2.6 & \(0.261\) & -- & \([0.139,\,0.380]\) \\
GLM-4.7 & \(0.228\) & -- & \([0.076,\,0.374]\) \\
DeepSeek-V4-Pro & \(0.316\) & -- & \([0.198,\,0.433]\) \\
GPT-5.5 & \(0.347\) & -- & \([0.221,\,0.470]\) \\
Claude Sonnet 4.6 & \(\mathbf{0.469}\) & -- & \(\mathbf{[0.343,\,0.588]}\) \\
\bottomrule
\end{tabular}
\caption{Uncertainty estimates for best-so-far MBNS. Seed SD is the sample standard deviation over seeds \(\{0,1,42\}\) and is reported only for locally controlled methods. Task-bootstrap intervals use the same 20,000 category-stratified paired resamples.}
\label{tab:appendix-best-so-far-robustness}
\end{table*}

\begin{table*}[!t]
\centering
\small
\setlength{\tabcolsep}{8pt}
\begin{tabular}{llc}
\toprule
\textbf{Selection criterion}
& \textbf{Method A -- Method B}
& \textbf{Paired MBNS difference [95\% CI]} \\
\midrule
Final step & Claude Sonnet 4.6 -- GPT-5.5 & \(0.102~[0.007,\,0.208]\) \\
Final step & Phi-4-14B -- Qwen3-32B & \(-0.018~[-0.071,\,0.034]\) \\
Final step & Qwen3-32B -- BOHB variant & \(0.131~[0.024,\,0.238]\) \\
\midrule
Best so far & Claude Sonnet 4.6 -- GPT-5.5 & \(0.122~[0.055,\,0.207]\) \\
Best so far & GPT-5.5 -- Random search & \(0.022~[-0.116,\,0.156]\) \\
Best so far & Random search -- Phi-4-14B & \(0.127~[0.029,\,0.232]\) \\
Best so far & Random search -- TPE & \(0.026~[-0.029,\,0.089]\) \\
Best so far & Random search -- BOHB variant & \(0.033~[-0.018,\,0.091]\) \\
Best so far & Phi-4-14B -- Qwen3-32B & \(0.003~[-0.048,\,0.053]\) \\
\bottomrule
\end{tabular}
\caption{Selected paired method comparisons under the final-step and best-so-far criteria. Each entry is Method A minus Method B. The intervals use 20,000 category-stratified paired task-bootstrap resamples. Controlled methods are first averaged over their three matched seeds at the task level.}
\label{tab:appendix-paired-method-differences}
\end{table*}

Table~\ref{tab:appendix-best-so-far-robustness} separates execution variation from sensitivity to benchmark composition. The three conventional HPO methods have overlapping best-so-far task-bootstrap intervals, as do Phi-4-14B and Qwen3-32B.

Table~\ref{tab:appendix-paired-method-differences} directly bootstraps task-level differences between selected methods. Under the final-step criterion, Claude Sonnet 4.6 has a positive paired difference relative to GPT-5.5, and Qwen3-32B has a positive difference relative to the BOHB variant. The interval for Phi-4-14B versus Qwen3-32B includes zero. Under best-so-far selection, Claude Sonnet 4.6 remains ahead of GPT-5.5, while the interval for GPT-5.5 versus random search includes zero. Random search has a positive paired difference relative to Phi-4-14B, whereas its differences from TPE and the BOHB variant are inconclusive. These results support reporting both selection criteria without treating small point-estimate differences as definitive rankings.

\subsection{Representative Decision Traces}
\label{app:qualitative-traces}

Table~\ref{tab:appendix-qualitative-traces} compares five agents on the same tunedGNN Cora GCN task. For the three open-weight agents, we show the seed-0 trace from the controlled three-seed evaluation so that each row represents one executable configuration trajectory. The API-agent rows show their single evaluation. All displayed traces start from the shared baseline $B=(\mathrm{lr}=0.001,\ \mathrm{hidden}=512,\ \mathrm{layers}=3,\ \mathrm{dropout}=0.7,\ \mathrm{weight\ decay}=0.0005,\ \mathrm{layer\ norm}=\mathrm{False},\ \mathrm{residual}=\mathrm{False})$, which obtains 83.5\% test accuracy at the epoch with the highest validation accuracy. Holding the task, baseline, and intervention space fixed reveals distinct responses to the same type of experimental feedback. Claude Sonnet 4.6 and GPT-5.5 recover from harmful initial changes and preserve subsequent improvements. Qwen3-32B also recovers but stops exploring after a modest gain, whereas Llama-3.1-8B discards its strongest configuration at the final intervention. Gemma2-2B repeatedly executes the baseline without testing an alternative.

\begin{table*}[t]
\centering
\footnotesize
\setlength{\tabcolsep}{3pt}
\renewcommand{\arraystretch}{1.18}
\begin{tabular}{p{1.8cm}p{7.0cm}p{3.2cm}p{4.3cm}}
\toprule
Agent & Configurations selected at interventions 1--5 & \centering Test accuracy (\%) & Qualitative pattern \\
\midrule
Claude Sonnet 4.6
& $I_1$: lr=0.005, hidden=256, dropout=0.5, layer norm=True;
$I_2$: retain $B$ except lr=0.002;
$I_3$: retain $I_2$ and set residual=True;
$I_4$--$I_5$: retain $I_3$
& \centering
$82.2 \rightarrow 83.8 \rightarrow 85.2$\par
$\rightarrow 85.2 \rightarrow 85.2$
& Recovers from a harmful joint change, isolates one architectural change, and retains the improved configuration. \\
\midrule
GPT-5.5
& $I_1$: lr=0.005, dropout=0.5;
$I_2$: retain $B$ except lr=0.002;
$I_3$: retain $I_2$ except layers=2;
$I_4$--$I_5$: retain $I_3$
& \centering
$82.2 \rightarrow 83.8 \rightarrow 84.4$\par
$\rightarrow 84.4 \rightarrow 84.4$
& Uses a conservative correction, tests a single change to model depth, and retains the improved configuration. \\
\midrule
Qwen3-32B
& $I_1$: lr=0.002, hidden=256, layers=4;
$I_2$: retain $B$ except lr=0.002;
$I_3$--$I_5$: retain $I_2$
& \centering
$79.4 \rightarrow 83.8 \rightarrow 83.8$\par
$\rightarrow 83.8 \rightarrow 83.8$
& Recovers from an aggressive first proposal but stops exploring after a modest improvement. \\
\midrule
Llama-3.1-8B
& $I_1$: lr=0.002, hidden=256, dropout=0.5, weight decay=0;
$I_2$: return to $B$;
$I_3$: return to $I_1$;
$I_4$: retain $I_1$ except lr=0.001;
$I_5$: retain $I_1$ except layers=4
& \centering
$85.0 \rightarrow 83.5 \rightarrow 85.0$\par
$\rightarrow 84.5 \rightarrow 81.0$
& Finds a strong configuration early but continues exploring and discards it at the final intervention. \\
\midrule
Gemma2-2B
& $I_1$--$I_5$: retain $B$
& \centering
$83.5 \rightarrow 83.5 \rightarrow 83.5$\par
$\rightarrow 83.5 \rightarrow 83.5$
& Repeats the baseline configuration without testing an alternative. \\
\bottomrule
\end{tabular}
\caption{Decision traces on tunedGNN Cora GCN under the limited budget protocol. $B$ denotes the shared baseline, and $I_i$ denotes intervention $i$. The accuracy column reports the outcomes of $I_1$--$I_5$. For compactness, configurations are described relative to $B$ or a referenced earlier intervention.}
\label{tab:appendix-qualitative-traces}
\end{table*}

% \FloatBarrier

\section{Metric Sensitivity}
\label{app:scope-limitations}

\paragraph{Metric sensitivity.}
Some tasks have a small gap between the limited-budget reference baseline and the reported anchor. In such cases, modest execution noise can produce a relatively large normalized change. Bounding each task score to \([-1,1]\) prevents an arbitrarily large contribution from one task, but does not remove sensitivity near a small denominator and can cause score saturation. We therefore report BWR, MAA, median BNS, task-level results, and task-composition bootstrap intervals alongside MBNS. The current benchmark does not separately estimate a noise-based minimum meaningful difference from repeated no-op executions.

\section{Detailed Experimental Results}
\label{app:detailed-results}

This section reports the complete per-task results for the main limited budget evaluation. The corresponding reference baseline values are provided in Table~\ref{tab:appendix-reference-baselines}. Tables~\ref{tab:appendix-raw-int1}--\ref{tab:appendix-raw-int5} report the raw metric observed after each of the five sequential interventions. For open-weight agents, raw metrics are averaged over seeds $\{0,1,42\}$ at each task and intervention, while bounded normalized scores and anchor attainment values are computed per seed and then averaged. API-agent entries correspond to their single audited evaluation. Values are shown in the same unit and direction as the corresponding task anchor in Table~\ref{tab:appendix-task-suite}. Raw values are rounded to three decimal places, whereas all derived scores are computed from the full precision values in the recorded traces. These tables show whether each agent improves, plateaus, or degrades across interventions instead of only reporting a final aggregate score.

The fifth intervention provides the final result used for aggregate scoring. Tables~\ref{tab:appendix-bounded-int1}--\ref{tab:appendix-bounded-int5} convert the intervention results into bounded normalized scores using the common limited budget baseline for each task. Tables~\ref{tab:appendix-anchor-attainment-int1}--\ref{tab:appendix-anchor-attainment-int5} report the corresponding anchor attainment values. Repeated raw values across adjacent interventions can indicate that an agent retained the same configuration or obtained the same rounded metric after execution. Nonmonotonic trajectories are expected because each intervention is evaluated as a new repository experiment rather than as a best result retained across previous trials.

\begin{table*}[t]
\centering
\tiny
\setlength{\tabcolsep}{1.8pt}
\resizebox{\textwidth}{!}{%
\begin{tabular}{lcccccccccccc}
\toprule
Task & Q8 & Q32 & Gem & DS14 & Phi & Llama & DSV4 & GPT & G4.7 & G5.1 & Kimi & Claude \\
\midrule
FineWeb Pretraining & 6.053 & 5.868 & 5.931 & 5.931 & 6.057 & 6.053 & 5.931 & 5.931 & 5.931 & 5.931 & 5.931 & 5.931 \\
ConvKAN CIFAR-10 & 26.883 & 23.183 & 24.917 & 27.930 & 33.660 & 17.647 & 38.370 & 26.250 & 39.730 & 25.180 & 27.370 & 13.950 \\
Open-R1 MATH-500 & 2.200 & 17.600 & 15.133 & 2.200 & 21.600 & 18.133 & 45.000 & 45.000 & 1.800 & 1.800 & 2.200 & 46.000 \\
Room Selector Tuning & 0.195 & 0.379 & 0.161 & 0.057 & 0.448 & 0.276 & 0.517 & 1.000 & 0.000 & 0.000 & 0.414 & 0.862 \\
Uni2TS ETTh1 Forecasting & 1.141 & 0.398 & 0.398 & 1.151 & 0.398 & 1.179 & 0.375 & 0.375 & 1.150 & 1.248 & 1.131 & 0.375 \\
TimesFM Long Horizon & 0.511 & 0.509 & 0.524 & 0.509 & 0.511 & 0.511 & 0.509 & 0.509 & 0.551 & 0.551 & 0.551 & 0.509 \\
ModernBERT MNLI & 39.803 & 35.446 & 41.481 & 40.051 & 50.382 & 35.446 & 35.446 & 35.446 & 35.446 & 35.446 & 35.446 & 35.446 \\
\bottomrule
\end{tabular}%
}
\caption{Per-task raw metric values after the first intervention in the main limited budget evaluation. }
\label{tab:appendix-raw-int1}
\end{table*}

\begin{table*}[t]
\centering
\tiny
\setlength{\tabcolsep}{1.8pt}
\resizebox{\textwidth}{!}{%
\begin{tabular}{lcccccccccccc}
\toprule
Task & Q8 & Q32 & Gem & DS14 & Phi & Llama & DSV4 & GPT & G4.7 & G5.1 & Kimi & Claude \\
\midrule
FineWeb Pretraining & 5.443 & 5.310 & 5.517 & 5.414 & 5.567 & 5.578 & 5.414 & 5.414 & 5.450 & 5.450 & 5.414 & 5.414 \\
ConvKAN CIFAR-10 & 21.687 & 30.760 & 29.730 & 20.980 & 26.040 & 24.923 & 13.240 & 27.540 & 45.080 & 23.830 & 26.710 & 36.460 \\
Open-R1 MATH-500 & 17.600 & 45.000 & 15.133 & 2.200 & 21.600 & 45.800 & 45.000 & 45.000 & 1.800 & 1.800 & 2.200 & 46.000 \\
Room Selector Tuning & 0.218 & 0.356 & 0.184 & 0.000 & 0.471 & 0.207 & 0.586 & 1.000 & 0.034 & 0.069 & 0.241 & 0.690 \\
Uni2TS ETTh1 Forecasting & 1.151 & 0.383 & 0.398 & 1.151 & 0.398 & 0.918 & 0.375 & 0.375 & 0.375 & 0.375 & 0.375 & 0.375 \\
TimesFM Long Horizon & 0.511 & 0.509 & 0.521 & 0.509 & 0.511 & 0.506 & 0.509 & 0.509 & 0.509 & 0.509 & 0.551 & 0.509 \\
ModernBERT MNLI & 38.553 & 40.499 & 41.464 & 40.353 & 35.446 & 37.694 & 35.446 & 37.830 & 35.446 & 35.446 & 38.462 & 35.446 \\
\bottomrule
\end{tabular}%
}
\caption{Per-task raw metric values after the second intervention in the main limited budget evaluation. }
\label{tab:appendix-raw-int2}
\end{table*}

\begin{table*}[t]
\centering
\tiny
\setlength{\tabcolsep}{1.8pt}
\resizebox{\textwidth}{!}{%
\begin{tabular}{lcccccccccccc}
\toprule
Task & Q8 & Q32 & Gem & DS14 & Phi & Llama & DSV4 & GPT & G4.7 & G5.1 & Kimi & Claude \\
\midrule
FineWeb Pretraining & 4.992 & 4.866 & 5.066 & 4.983 & 5.147 & 5.030 & 4.983 & 4.983 & 5.064 & 5.064 & 4.965 & 4.983 \\
ConvKAN CIFAR-10 & 23.653 & 27.330 & 22.153 & 26.757 & 20.983 & 26.917 & 41.980 & 21.540 & 31.980 & 25.350 & 15.630 & 47.950 \\
Open-R1 MATH-500 & 17.600 & 45.000 & 15.133 & 2.200 & 21.600 & 46.267 & 45.000 & 45.000 & 1.800 & 1.800 & 2.200 & 46.000 \\
Room Selector Tuning & 0.218 & 0.356 & 0.149 & 0.402 & 0.379 & 0.207 & 0.586 & 1.000 & 0.069 & 0.069 & 0.448 & 0.897 \\
Uni2TS ETTh1 Forecasting & 0.899 & 0.383 & 0.398 & 1.147 & 0.398 & 0.390 & 0.375 & 0.375 & 0.375 & 0.375 & 0.375 & 0.375 \\
TimesFM Long Horizon & 0.511 & 0.509 & 0.524 & 0.509 & 0.511 & 0.508 & 0.509 & 0.509 & 0.509 & 0.509 & 0.551 & 0.509 \\
ModernBERT MNLI & 41.729 & 40.316 & 40.285 & 40.598 & 52.304 & 42.272 & 37.565 & 35.446 & 35.446 & 35.446 & 35.446 & 36.943 \\
\bottomrule
\end{tabular}%
}
\caption{Per-task raw metric values after the third intervention in the main limited budget evaluation. }
\label{tab:appendix-raw-int3}
\end{table*}

\begin{table*}[t]
\centering
\tiny
\setlength{\tabcolsep}{1.8pt}
\resizebox{\textwidth}{!}{%
\begin{tabular}{lcccccccccccc}
\toprule
Task & Q8 & Q32 & Gem & DS14 & Phi & Llama & DSV4 & GPT & G4.7 & G5.1 & Kimi & Claude \\
\midrule
FineWeb Pretraining & 4.529 & 4.414 & 4.645 & 4.513 & 4.723 & 4.732 & 4.513 & 4.513 & 4.648 & 4.648 & 4.588 & 4.513 \\
ConvKAN CIFAR-10 & 26.723 & 30.077 & 28.067 & 18.270 & 29.077 & 22.863 & 27.510 & 16.210 & 41.580 & 27.860 & 23.350 & 20.740 \\
Open-R1 MATH-500 & 17.600 & 47.200 & 15.133 & 2.200 & 21.600 & 36.133 & 44.600 & 45.000 & 1.800 & 1.800 & 2.200 & 46.000 \\
Room Selector Tuning & 0.184 & 0.356 & 0.138 & 0.126 & 0.483 & 0.379 & 0.483 & 1.000 & 0.034 & 0.000 & 0.069 & 0.897 \\
Uni2TS ETTh1 Forecasting & 0.635 & 0.383 & 0.398 & 0.635 & 0.398 & 0.375 & 0.375 & 0.375 & 0.375 & 0.375 & 0.375 & 0.375 \\
TimesFM Long Horizon & 0.511 & 0.509 & 0.521 & 0.509 & 0.511 & 0.510 & 0.509 & 0.509 & 0.509 & 0.509 & 0.509 & 0.509 \\
ModernBERT MNLI & 40.628 & 40.156 & 40.645 & 39.168 & 50.053 & 35.446 & 37.310 & 36.577 & 35.446 & 35.446 & 35.446 & 39.918 \\
\bottomrule
\end{tabular}%
}
\caption{Per-task raw metric values after the fourth intervention in the main limited budget evaluation. }
\label{tab:appendix-raw-int4}
\end{table*}

\begin{table*}[t]
\centering
\tiny
\setlength{\tabcolsep}{1.8pt}
\resizebox{\textwidth}{!}{%
\begin{tabular}{lcccccccccccc}
\toprule
Task & Q8 & Q32 & Gem & DS14 & Phi & Llama & DSV4 & GPT & G4.7 & G5.1 & Kimi & Claude \\
\midrule
FineWeb Pretraining & 4.377 & 4.274 & 4.357 & 4.344 & 4.533 & 4.419 & 4.406 & 4.406 & 4.530 & 4.530 & 4.480 & 4.406 \\
ConvKAN CIFAR-10 & 25.413 & 31.240 & 26.400 & 16.160 & 25.390 & 22.883 & 32.210 & 29.590 & 38.780 & 35.970 & 47.030 & 37.620 \\
Open-R1 MATH-500 & 17.600 & 47.200 & 15.133 & 2.200 & 21.600 & 45.600 & 45.000 & 45.000 & 1.800 & 1.800 & 2.200 & 46.000 \\
Room Selector Tuning & 0.195 & 0.379 & 0.161 & 0.057 & 0.448 & 0.276 & 0.483 & 1.000 & 0.000 & 0.000 & 0.517 & 0.862 \\
Uni2TS ETTh1 Forecasting & 0.636 & 0.383 & 0.398 & 0.626 & 0.398 & 0.383 & 0.375 & 0.375 & 0.375 & 0.375 & 0.375 & 0.375 \\
TimesFM Long Horizon & 0.511 & 0.509 & 0.524 & 0.509 & 0.511 & 0.506 & 0.509 & 0.509 & 0.509 & 0.509 & 0.509 & 0.509 \\
ModernBERT MNLI & 41.729 & 45.138 & 41.484 & 39.925 & 56.529 & 45.617 & 38.095 & 38.125 & 35.446 & 35.446 & 38.034 & 38.074 \\
\bottomrule
\end{tabular}%
}
\caption{Per-task raw metric values after the fifth intervention in the main limited budget evaluation. }
\label{tab:appendix-raw-int5}
\end{table*}

\begin{table*}[t]
\centering
\tiny
\setlength{\tabcolsep}{1.8pt}
\resizebox{\textwidth}{!}{%
\begin{tabular}{lcccccccccccc}
\toprule
Task & Q8 & Q32 & Gem & DS14 & Phi & Llama & DSV4 & GPT & G4.7 & G5.1 & Kimi & Claude \\
\midrule
FineWeb Pretraining & -1.000 & -1.000 & -1.000 & -1.000 & -1.000 & -1.000 & -1.000 & -1.000 & -1.000 & -1.000 & -1.000 & -1.000 \\
ConvKAN CIFAR-10 & -0.038 & -0.105 & -0.073 & -0.019 & 0.085 & -0.205 & 0.170 & -0.049 & 0.195 & -0.069 & -0.029 & -0.272 \\
Open-R1 MATH-500 & 0.008 & 0.197 & 0.166 & 0.008 & 0.246 & 0.203 & 0.533 & 0.533 & 0.003 & 0.003 & 0.008 & 0.545 \\
Room Selector Tuning & -0.711 & -0.112 & -0.782 & -1.000 & 0.112 & -0.449 & 0.337 & 1.000 & -1.000 & -1.000 & 0.000 & 1.000 \\
Uni2TS ETTh1 Forecasting & 0.018 & 0.971 & 0.971 & 0.005 & 0.971 & -0.031 & 1.000 & 1.000 & 0.006 & -0.119 & 0.031 & 1.000 \\
TimesFM Long Horizon & 0.953 & 0.991 & 0.635 & 0.991 & 0.953 & 0.953 & 0.991 & 0.990 & 0.000 & 0.000 & 0.000 & 0.991 \\
ModernBERT MNLI & 0.016 & -0.068 & 0.049 & 0.021 & 0.222 & -0.068 & -0.068 & -0.068 & -0.068 & -0.068 & -0.068 & -0.068 \\
\bottomrule
\end{tabular}%
}
\caption{Per-task bounded normalized scores after the first intervention in the main limited budget evaluation. Scores are bounded to $[-1,1]$.}
\label{tab:appendix-bounded-int1}
\end{table*}

\begin{table*}[t]
\centering
\tiny
\setlength{\tabcolsep}{1.8pt}
\resizebox{\textwidth}{!}{%
\begin{tabular}{lcccccccccccc}
\toprule
Task & Q8 & Q32 & Gem & DS14 & Phi & Llama & DSV4 & GPT & G4.7 & G5.1 & Kimi & Claude \\
\midrule
FineWeb Pretraining & -1.000 & -1.000 & -1.000 & -1.000 & -1.000 & -1.000 & -1.000 & -1.000 & -1.000 & -1.000 & -1.000 & -1.000 \\
ConvKAN CIFAR-10 & -0.132 & 0.032 & 0.014 & -0.145 & -0.053 & -0.073 & -0.285 & -0.026 & 0.292 & -0.093 & -0.041 & 0.136 \\
Open-R1 MATH-500 & 0.197 & 0.533 & 0.166 & 0.008 & 0.246 & 0.543 & 0.533 & 0.533 & 0.003 & 0.003 & 0.008 & 0.545 \\
Room Selector Tuning & -0.636 & -0.187 & -0.708 & -1.000 & 0.187 & -0.674 & 0.561 & 1.000 & -1.000 & -1.000 & -0.561 & 0.898 \\
Uni2TS ETTh1 Forecasting & 0.006 & 0.990 & 0.971 & 0.005 & 0.971 & 0.304 & 1.000 & 1.000 & 1.000 & 1.000 & 1.000 & 1.000 \\
TimesFM Long Horizon & 0.953 & 0.991 & 0.667 & 0.991 & 0.953 & 1.000 & 0.991 & 0.990 & 0.991 & 0.991 & 0.000 & 0.991 \\
ModernBERT MNLI & -0.008 & 0.030 & 0.049 & 0.027 & -0.068 & -0.025 & -0.068 & -0.022 & -0.068 & -0.068 & -0.010 & -0.068 \\
\bottomrule
\end{tabular}%
}
\caption{Per-task bounded normalized scores after the second intervention in the main limited budget evaluation. Scores are bounded to $[-1,1]$.}
\label{tab:appendix-bounded-int2}
\end{table*}

\begin{table*}[t]
\centering
\tiny
\setlength{\tabcolsep}{1.8pt}
\resizebox{\textwidth}{!}{%
\begin{tabular}{lcccccccccccc}
\toprule
Task & Q8 & Q32 & Gem & DS14 & Phi & Llama & DSV4 & GPT & G4.7 & G5.1 & Kimi & Claude \\
\midrule
FineWeb Pretraining & -0.705 & -0.568 & -0.784 & -0.694 & -0.873 & -0.746 & -0.694 & -0.694 & -0.782 & -0.782 & -0.675 & -0.694 \\
ConvKAN CIFAR-10 & -0.096 & -0.030 & -0.123 & -0.040 & -0.145 & -0.037 & 0.236 & -0.135 & 0.055 & -0.066 & -0.242 & 0.344 \\
Open-R1 MATH-500 & 0.197 & 0.533 & 0.166 & 0.008 & 0.246 & 0.548 & 0.533 & 0.533 & 0.003 & 0.003 & 0.008 & 0.545 \\
Room Selector Tuning & -0.636 & -0.187 & -0.782 & -0.037 & -0.112 & -0.674 & 0.561 & 1.000 & -1.000 & -1.000 & 0.112 & 1.000 \\
Uni2TS ETTh1 Forecasting & 0.328 & 0.990 & 0.971 & 0.010 & 0.971 & 0.980 & 1.000 & 1.000 & 1.000 & 1.000 & 1.000 & 1.000 \\
TimesFM Long Horizon & 0.953 & 0.991 & 0.635 & 0.991 & 0.953 & 0.994 & 0.991 & 0.990 & 0.991 & 0.991 & 0.000 & 0.991 \\
ModernBERT MNLI & 0.054 & 0.026 & 0.026 & 0.032 & 0.259 & 0.064 & -0.027 & -0.068 & -0.068 & -0.068 & -0.068 & -0.039 \\
\bottomrule
\end{tabular}%
}
\caption{Per-task bounded normalized scores after the third intervention in the main limited budget evaluation. Scores are bounded to $[-1,1]$.}
\label{tab:appendix-bounded-int3}
\end{table*}

\begin{table*}[t]
\centering
\tiny
\setlength{\tabcolsep}{1.8pt}
\resizebox{\textwidth}{!}{%
\begin{tabular}{lcccccccccccc}
\toprule
Task & Q8 & Q32 & Gem & DS14 & Phi & Llama & DSV4 & GPT & G4.7 & G5.1 & Kimi & Claude \\
\midrule
FineWeb Pretraining & -0.201 & -0.076 & -0.327 & -0.183 & -0.411 & -0.421 & -0.183 & -0.183 & -0.330 & -0.330 & -0.265 & -0.183 \\
ConvKAN CIFAR-10 & -0.041 & 0.020 & -0.016 & -0.194 & 0.002 & -0.111 & -0.026 & -0.231 & 0.228 & -0.020 & -0.102 & -0.149 \\
Open-R1 MATH-500 & 0.197 & 0.560 & 0.166 & 0.008 & 0.246 & 0.424 & 0.528 & 0.533 & 0.003 & 0.003 & 0.008 & 0.545 \\
Room Selector Tuning & -0.748 & -0.187 & -0.857 & -0.932 & 0.224 & -0.112 & 0.224 & 1.000 & -1.000 & -1.000 & -1.000 & 1.000 \\
Uni2TS ETTh1 Forecasting & 0.667 & 0.990 & 0.971 & 0.667 & 0.971 & 1.000 & 1.000 & 1.000 & 1.000 & 1.000 & 1.000 & 1.000 \\
TimesFM Long Horizon & 0.953 & 0.991 & 0.667 & 0.991 & 0.953 & 0.978 & 0.991 & 0.990 & 0.991 & 0.991 & 0.991 & 0.991 \\
ModernBERT MNLI & 0.032 & 0.023 & 0.033 & 0.004 & 0.216 & -0.068 & -0.032 & -0.046 & -0.068 & -0.068 & -0.068 & 0.019 \\
\bottomrule
\end{tabular}%
}
\caption{Per-task bounded normalized scores after the fourth intervention in the main limited budget evaluation. Scores are bounded to $[-1,1]$.}
\label{tab:appendix-bounded-int4}
\end{table*}

\begin{table*}[t]
\centering
\tiny
\setlength{\tabcolsep}{1.8pt}
\resizebox{\textwidth}{!}{%
\begin{tabular}{lcccccccccccc}
\toprule
Task & Q8 & Q32 & Gem & DS14 & Phi & Llama & DSV4 & GPT & G4.7 & G5.1 & Kimi & Claude \\
\midrule
FineWeb Pretraining & -0.036 & 0.077 & -0.014 & 0.000 & -0.205 & -0.081 & -0.067 & -0.067 & -0.202 & -0.202 & -0.147 & -0.067 \\
ConvKAN CIFAR-10 & -0.064 & 0.041 & -0.047 & -0.232 & -0.065 & -0.110 & 0.059 & 0.011 & 0.178 & 0.127 & 0.327 & 0.157 \\
Open-R1 MATH-500 & 0.197 & 0.560 & 0.166 & 0.008 & 0.246 & 0.540 & 0.533 & 0.533 & 0.003 & 0.003 & 0.008 & 0.545 \\
Room Selector Tuning & -0.711 & -0.112 & -0.782 & -1.000 & 0.112 & -0.449 & 0.224 & 1.000 & -1.000 & -1.000 & 0.337 & 1.000 \\
Uni2TS ETTh1 Forecasting & 0.665 & 0.990 & 0.971 & 0.678 & 0.971 & 0.990 & 1.000 & 1.000 & 1.000 & 1.000 & 1.000 & 1.000 \\
TimesFM Long Horizon & 0.953 & 0.990 & 0.635 & 0.991 & 0.953 & 1.000 & 0.991 & 0.990 & 0.991 & 0.991 & 0.991 & 0.991 \\
ModernBERT MNLI & 0.054 & 0.120 & 0.049 & 0.019 & 0.342 & 0.129 & -0.017 & -0.016 & -0.068 & -0.068 & -0.018 & -0.017 \\
\bottomrule
\end{tabular}%
}
\caption{Per-task bounded normalized scores after the fifth intervention in the main limited budget evaluation. Scores are bounded to $[-1,1]$.}
\label{tab:appendix-bounded-int5}
\end{table*}

\begin{table*}[t]
\centering
\tiny
\setlength{\tabcolsep}{1.8pt}
\resizebox{\textwidth}{!}{%
\begin{tabular}{lcccccccccccc}
\toprule
Task & Q8 & Q32 & Gem & DS14 & Phi & Llama & DSV4 & GPT & G4.7 & G5.1 & Kimi & Claude \\
\midrule
FineWeb Pretraining & 56.6 & 58.4 & 57.7 & 57.7 & 56.5 & 56.6 & 57.7 & 57.7 & 57.7 & 57.7 & 57.7 & 57.7 \\
ConvKAN CIFAR-10 & 31.9 & 27.5 & 29.6 & 33.2 & 40.0 & 21.0 & 45.6 & 31.2 & 47.2 & 29.9 & 32.5 & 16.6 \\
Open-R1 MATH-500 & 2.6 & 21.2 & 18.2 & 2.6 & 26.0 & 21.8 & 54.2 & 54.2 & 2.2 & 2.2 & 2.6 & 55.4 \\
Room Selector Tuning & 27.1 & 52.6 & 22.3 & 8.0 & 62.2 & 38.3 & 71.7 & 138.7 & 0.0 & 0.0 & 57.4 & 119.6 \\
Uni2TS ETTh1 Forecasting & 32.9 & 94.2 & 94.2 & 32.6 & 94.2 & 31.8 & 100.0 & 100.0 & 32.6 & 30.0 & 33.2 & 100.0 \\
TimesFM Long Horizon & 99.6 & 99.9 & 97.2 & 99.9 & 99.6 & 99.6 & 99.9 & 99.9 & 92.3 & 92.3 & 92.3 & 99.9 \\
ModernBERT MNLI & 44.0 & 39.2 & 45.9 & 44.3 & 55.7 & 39.2 & 39.2 & 39.2 & 39.2 & 39.2 & 39.2 & 39.2 \\
\bottomrule
\end{tabular}%
}
\caption{Per-task anchor attainment after the first intervention in the main limited budget evaluation, reported as percentages.}
\label{tab:appendix-anchor-attainment-int1}
\end{table*}

\begin{table*}[t]
\centering
\tiny
\setlength{\tabcolsep}{1.8pt}
\resizebox{\textwidth}{!}{%
\begin{tabular}{lcccccccccccc}
\toprule
Task & Q8 & Q32 & Gem & DS14 & Phi & Llama & DSV4 & GPT & G4.7 & G5.1 & Kimi & Claude \\
\midrule
FineWeb Pretraining & 62.9 & 64.5 & 62.1 & 63.3 & 61.5 & 61.4 & 63.3 & 63.3 & 62.8 & 62.8 & 63.3 & 63.3 \\
ConvKAN CIFAR-10 & 25.8 & 36.5 & 35.3 & 24.9 & 30.9 & 29.6 & 15.7 & 32.7 & 53.6 & 28.3 & 31.7 & 43.3 \\
Open-R1 MATH-500 & 21.2 & 54.2 & 18.2 & 2.6 & 26.0 & 55.1 & 54.2 & 54.2 & 2.2 & 2.2 & 2.6 & 55.4 \\
Room Selector Tuning & 30.3 & 49.4 & 25.5 & 0.0 & 65.4 & 28.7 & 81.3 & 138.7 & 4.8 & 9.6 & 33.5 & 95.7 \\
Uni2TS ETTh1 Forecasting & 32.6 & 98.1 & 94.2 & 32.6 & 94.2 & 52.7 & 100.0 & 100.0 & 100.0 & 100.0 & 100.0 & 100.0 \\
TimesFM Long Horizon & 99.6 & 99.9 & 97.9 & 99.9 & 99.6 & 100.7 & 99.9 & 99.9 & 99.9 & 99.9 & 92.3 & 99.9 \\
ModernBERT MNLI & 42.6 & 44.8 & 45.9 & 44.6 & 39.2 & 41.7 & 39.2 & 41.8 & 39.2 & 39.2 & 42.5 & 39.2 \\
\bottomrule
\end{tabular}%
}
\caption{Per-task anchor attainment after the second intervention in the main limited budget evaluation, reported as percentages.}
\label{tab:appendix-anchor-attainment-int2}
\end{table*}

\begin{table*}[t]
\centering
\tiny
\setlength{\tabcolsep}{1.8pt}
\resizebox{\textwidth}{!}{%
\begin{tabular}{lcccccccccccc}
\toprule
Task & Q8 & Q32 & Gem & DS14 & Phi & Llama & DSV4 & GPT & G4.7 & G5.1 & Kimi & Claude \\
\midrule
FineWeb Pretraining & 68.6 & 70.4 & 67.7 & 68.7 & 66.5 & 68.1 & 68.7 & 68.7 & 67.6 & 67.6 & 69.0 & 68.7 \\
ConvKAN CIFAR-10 & 28.1 & 32.5 & 26.3 & 31.8 & 24.9 & 32.0 & 49.9 & 25.6 & 38.0 & 30.1 & 18.6 & 57.0 \\
Open-R1 MATH-500 & 21.2 & 54.2 & 18.2 & 2.6 & 26.0 & 55.7 & 54.2 & 54.2 & 2.2 & 2.2 & 2.6 & 55.4 \\
Room Selector Tuning & 30.3 & 49.4 & 20.7 & 55.8 & 52.6 & 28.7 & 81.3 & 138.7 & 9.6 & 9.6 & 62.2 & 124.3 \\
Uni2TS ETTh1 Forecasting & 53.1 & 98.0 & 94.2 & 32.7 & 94.2 & 96.1 & 100.0 & 100.0 & 100.0 & 100.0 & 100.0 & 100.0 \\
TimesFM Long Horizon & 99.6 & 99.9 & 97.2 & 99.9 & 99.6 & 100.2 & 99.9 & 99.9 & 99.9 & 99.9 & 92.3 & 99.9 \\
ModernBERT MNLI & 46.2 & 44.6 & 44.6 & 44.9 & 57.9 & 46.8 & 41.6 & 39.2 & 39.2 & 39.2 & 39.2 & 40.9 \\
\bottomrule
\end{tabular}%
}
\caption{Per-task anchor attainment after the third intervention in the main limited budget evaluation, reported as percentages.}
\label{tab:appendix-anchor-attainment-int3}
\end{table*}

\begin{table*}[t]
\centering
\tiny
\setlength{\tabcolsep}{1.8pt}
\resizebox{\textwidth}{!}{%
\begin{tabular}{lcccccccccccc}
\toprule
Task & Q8 & Q32 & Gem & DS14 & Phi & Llama & DSV4 & GPT & G4.7 & G5.1 & Kimi & Claude \\
\midrule
FineWeb Pretraining & 75.6 & 77.6 & 73.8 & 75.9 & 72.5 & 72.4 & 75.9 & 75.9 & 73.7 & 73.7 & 74.6 & 75.9 \\
ConvKAN CIFAR-10 & 31.7 & 35.7 & 33.3 & 21.7 & 34.5 & 27.2 & 32.7 & 19.3 & 49.4 & 33.1 & 27.7 & 24.6 \\
Open-R1 MATH-500 & 21.2 & 56.8 & 18.2 & 2.6 & 26.0 & 43.5 & 53.7 & 54.2 & 2.2 & 2.2 & 2.6 & 55.4 \\
Room Selector Tuning & 25.5 & 49.4 & 19.1 & 17.5 & 67.0 & 52.6 & 67.0 & 138.7 & 4.8 & 0.0 & 9.6 & 124.3 \\
Uni2TS ETTh1 Forecasting & 75.8 & 98.1 & 94.2 & 77.5 & 94.2 & 100.0 & 100.0 & 100.0 & 100.0 & 100.0 & 100.0 & 100.0 \\
TimesFM Long Horizon & 99.6 & 99.9 & 97.9 & 99.9 & 99.6 & 99.8 & 99.9 & 99.9 & 99.9 & 99.9 & 99.9 & 99.9 \\
ModernBERT MNLI & 44.9 & 44.4 & 45.0 & 43.3 & 55.4 & 39.2 & 41.3 & 40.5 & 39.2 & 39.2 & 39.2 & 44.2 \\
\bottomrule
\end{tabular}%
}
\caption{Per-task anchor attainment after the fourth intervention in the main limited budget evaluation, reported as percentages.}
\label{tab:appendix-anchor-attainment-int4}
\end{table*}

\begin{table*}[t]
\centering
\tiny
\setlength{\tabcolsep}{1.8pt}
\resizebox{\textwidth}{!}{%
\begin{tabular}{lcccccccccccc}
\toprule
Task & Q8 & Q32 & Gem & DS14 & Phi & Llama & DSV4 & GPT & G4.7 & G5.1 & Kimi & Claude \\
\midrule
FineWeb Pretraining & 78.3 & 80.1 & 78.6 & 78.8 & 75.6 & 77.5 & 77.7 & 77.7 & 75.6 & 75.6 & 76.5 & 77.7 \\
ConvKAN CIFAR-10 & 30.2 & 37.1 & 31.4 & 19.2 & 30.2 & 27.2 & 38.3 & 35.2 & 46.1 & 42.7 & 55.9 & 44.7 \\
Open-R1 MATH-500 & 21.2 & 56.8 & 18.2 & 2.6 & 26.0 & 54.9 & 54.2 & 54.2 & 2.2 & 2.2 & 2.6 & 55.4 \\
Room Selector Tuning & 27.1 & 52.6 & 22.3 & 8.0 & 62.2 & 38.3 & 67.0 & 138.7 & 0.0 & 0.0 & 71.7 & 119.6 \\
Uni2TS ETTh1 Forecasting & 75.8 & 98.0 & 94.2 & 77.7 & 94.2 & 98.1 & 100.0 & 100.0 & 100.0 & 100.0 & 100.0 & 100.0 \\
TimesFM Long Horizon & 99.6 & 99.9 & 97.2 & 99.9 & 99.6 & 100.7 & 99.9 & 99.9 & 99.9 & 99.9 & 99.9 & 99.9 \\
ModernBERT MNLI & 46.2 & 49.9 & 45.9 & 44.2 & 62.5 & 50.5 & 42.1 & 42.2 & 39.2 & 39.2 & 42.1 & 42.1 \\
\bottomrule
\end{tabular}%
}
\caption{Per-task anchor attainment after the fifth intervention in the main limited budget evaluation, reported as percentages.}
\label{tab:appendix-anchor-attainment-int5}
\end{table*}

\section{Full Budget and Harness Ablation Details}
\label{app:full-budget-harness}

The full budget evaluation uses the same task definitions, agent interface, result schema, scoring rules, and five intervention protocol as the limited budget evaluation. Each baseline and intervention is instead executed with the full training or evaluation budget of the original repository. The complete intervention level raw results are shown in Tables~\ref{tab:appendix-raw-full-int1}--\ref{tab:appendix-raw-full-int5}. The bounded normalized scores are shown in Tables~\ref{tab:appendix-bounded-full-int1}--\ref{tab:appendix-bounded-full-int5}, and the anchor attainment values are shown in Tables~\ref{tab:appendix-anchor-attainment-full-int1}--\ref{tab:appendix-anchor-attainment-full-int5}. Raw metrics are displayed to three decimal places, while all derived scores use the corresponding full precision values.

The harness ablation keeps the AgentHPOBench task suite, intervention budget, result schema, and scoring pipeline fixed while changing the external execution harness used by the backbone agent. For Claude Sonnet 4.6, the comparison is between the native AgentHPOBench harness and Claude Code CLI. For GPT-5.5, the comparison is between the native AgentHPOBench harness and Codex CLI. These experiments isolate the effect of the execution interface from the benchmark task definitions and scoring code. The complete intervention level raw results are shown in Tables~\ref{tab:appendix-raw-harness-int1}--\ref{tab:appendix-raw-harness-int5}. The bounded normalized scores are shown in Tables~\ref{tab:appendix-bounded-harness-int1}--\ref{tab:appendix-bounded-harness-int5}, and the anchor attainment values are shown in Tables~\ref{tab:appendix-anchor-attainment-harness-int1}--\ref{tab:appendix-anchor-attainment-harness-int5}.

\begin{table*}[t]
\centering
\tiny
\setlength{\tabcolsep}{3pt}
\resizebox{0.5\textwidth}{!}{%
\begin{tabular}{lccc}
\toprule
Task & Q32 & DS-R1 & Claude \\
\midrule
FineWeb Pretraining & 3.548 & 3.548 & 3.529 \\
ConvKAN CIFAR-10 & 41.390 & 36.860 & 35.500 \\
Open-R1 MATH-500 & 22.000 & 2.200 & 46.000 \\
Room Selector Tuning & 0.379 & 0.069 & 1.000 \\
Uni2TS ETTh1 Forecasting & 0.375 & 0.375 & 0.375 \\
TimesFM Long Horizon & 0.509 & 0.509 & 0.509 \\
ModernBERT MNLI & 35.450 & 35.450 & 35.450 \\
AirBench CIFAR-10 & 93.420 & 93.560 & 93.460 \\
NoisyGL Cora GCN & 70.000 & 69.800 & 70.400 \\
TabMini Promoters & 0.928 & 0.968 & 0.982 \\
Adult Tabular Diffusion & 85.520 & 85.530 & 87.480 \\
xLSTM Parity & 0.067 & 0.067 & 0.999 \\
VBLL Yacht Regression & 0.334 & 0.334 & 0.294 \\
TabM California Housing & 0.504 & 0.504 & 0.504 \\
TimeXer PJM Forecasting & 0.121 & 0.121 & 0.121 \\
ABLkit HWF Reasoning & 96.700 & 96.800 & 96.000 \\
tunedGNN Cora GCN & 82.900 & 82.900 & 82.300 \\
verl GRPO GSM8K & 84.460 & 84.000 & 83.240 \\
ForestDiffusion Iris & 0.979 & 0.979 & 0.979 \\
VAR ImageNet 256 & 70.710 & 70.710 & 72.680 \\
RAGEN Bandit Alignment & 0.996 & 0.997 & 0.997 \\
Chronos Weather Forecasting & 0.372 & 0.372 & 0.385 \\
HyperbolicCV CIFAR-100 & 54.370 & 56.200 & 78.170 \\
iTransformer ETTm2 Forecasting & 0.185 & 0.185 & 0.182 \\
TimeMixer ETTm2 Forecasting & 0.175 & 0.175 & 0.175 \\
ART 2048 & 0.280 & 0.280 & 0.000 \\
Multimodal Open-R1 MathVista & 40.200 & 40.200 & 40.200 \\
RankUp UTKFace Regression & 7.690 & 8.910 & 5.790 \\
SparseTSF ETTm1 Forecasting & 0.340 & 0.340 & 0.340 \\
SimpleRL MATH-500 & 5.000 & 4.800 & 5.800 \\
\bottomrule
\end{tabular}%
}
\caption{Per-task raw metric values after the first intervention in the full budget evaluation. }
\label{tab:appendix-raw-full-int1}
\end{table*}

\begin{table*}[t]
\centering
\tiny
\setlength{\tabcolsep}{3pt}
\resizebox{0.5\textwidth}{!}{%
\begin{tabular}{lccc}
\toprule
Task & Q32 & DS-R1 & Claude \\
\midrule
FineWeb Pretraining & 3.508 & 3.508 & 3.466 \\
ConvKAN CIFAR-10 & 35.580 & 60.550 & 48.220 \\
Open-R1 MATH-500 & 22.000 & 2.200 & 46.000 \\
Room Selector Tuning & 0.379 & 0.000 & 0.690 \\
Uni2TS ETTh1 Forecasting & 0.375 & 0.375 & 0.375 \\
TimesFM Long Horizon & 0.509 & 0.509 & 0.509 \\
ModernBERT MNLI & 35.450 & 35.450 & 35.450 \\
AirBench CIFAR-10 & 93.610 & 93.560 & 93.730 \\
NoisyGL Cora GCN & 69.800 & 69.800 & 69.000 \\
TabMini Promoters & 0.968 & 0.968 & 0.970 \\
Adult Tabular Diffusion & 85.530 & 85.530 & 87.640 \\
xLSTM Parity & 0.298 & 1.000 & 1.000 \\
VBLL Yacht Regression & 0.334 & 0.334 & 0.341 \\
TabM California Housing & 0.504 & 0.504 & 0.494 \\
TimeXer PJM Forecasting & 0.121 & 0.121 & 0.122 \\
ABLkit HWF Reasoning & 95.350 & 95.350 & 97.200 \\
tunedGNN Cora GCN & 83.100 & 83.200 & 83.800 \\
verl GRPO GSM8K & 84.530 & 84.690 & 85.370 \\
ForestDiffusion Iris & 0.979 & 0.979 & 0.981 \\
VAR ImageNet 256 & 70.710 & 70.710 & 64.000 \\
RAGEN Bandit Alignment & 0.996 & 0.998 & 0.997 \\
Chronos Weather Forecasting & 0.368 & 0.368 & 0.368 \\
HyperbolicCV CIFAR-100 & 57.140 & 53.200 & 76.580 \\
iTransformer ETTm2 Forecasting & 0.185 & 0.185 & 0.185 \\
TimeMixer ETTm2 Forecasting & 0.174 & 0.175 & 0.180 \\
ART 2048 & 0.280 & 0.280 & 0.000 \\
Multimodal Open-R1 MathVista & 48.600 & 48.600 & 48.700 \\
RankUp UTKFace Regression & 6.060 & 6.070 & 5.610 \\
SparseTSF ETTm1 Forecasting & 0.340 & 0.340 & 0.341 \\
SimpleRL MATH-500 & 4.400 & 4.800 & 7.500 \\
\bottomrule
\end{tabular}%
}
\caption{Per-task raw metric values after the second intervention in the full budget evaluation. }
\label{tab:appendix-raw-full-int2}
\end{table*}

\begin{table*}[t]
\centering

\tiny
\setlength{\tabcolsep}{3pt}
\resizebox{0.5\textwidth}{!}{%
\begin{tabular}{lccc}
\toprule
Task & Q32 & DS-R1 & Claude \\
\midrule
FineWeb Pretraining & 3.488 & 3.488 & 3.435 \\
ConvKAN CIFAR-10 & 29.200 & 50.820 & 41.630 \\
Open-R1 MATH-500 & 22.000 & 2.200 & 45.800 \\
Room Selector Tuning & 0.379 & 0.414 & 1.000 \\
Uni2TS ETTh1 Forecasting & 0.375 & 0.375 & 0.375 \\
TimesFM Long Horizon & 0.509 & 0.509 & 0.509 \\
ModernBERT MNLI & 35.450 & 59.790 & 67.010 \\
AirBench CIFAR-10 & 93.680 & 93.560 & 93.680 \\
NoisyGL Cora GCN & 69.800 & 69.800 & 70.400 \\
TabMini Promoters & 0.968 & 0.968 & 0.982 \\
Adult Tabular Diffusion & 85.530 & 85.530 & 87.420 \\
xLSTM Parity & 0.298 & 0.298 & 1.000 \\
VBLL Yacht Regression & 0.334 & 0.334 & 0.302 \\
TabM California Housing & 0.504 & 0.504 & 0.493 \\
TimeXer PJM Forecasting & 0.121 & 0.121 & 0.114 \\
ABLkit HWF Reasoning & 95.500 & 95.500 & 97.450 \\
tunedGNN Cora GCN & 82.600 & 82.700 & 83.900 \\
verl GRPO GSM8K & 84.990 & 86.050 & 85.290 \\
ForestDiffusion Iris & 0.979 & 0.979 & 0.979 \\
VAR ImageNet 256 & 70.710 & 70.710 & 78.770 \\
RAGEN Bandit Alignment & 0.994 & 0.996 & 0.997 \\
Chronos Weather Forecasting & 0.373 & 0.373 & 0.372 \\
HyperbolicCV CIFAR-100 & 54.680 & 54.640 & 75.090 \\
iTransformer ETTm2 Forecasting & 0.185 & 0.185 & 0.184 \\
TimeMixer ETTm2 Forecasting & 0.175 & 0.175 & 0.175 \\
ART 2048 & 0.280 & 0.280 & 0.080 \\
Multimodal Open-R1 MathVista & 46.300 & 46.300 & 48.600 \\
RankUp UTKFace Regression & 5.450 & 6.450 & 6.220 \\
SparseTSF ETTm1 Forecasting & 0.340 & 0.340 & 0.325 \\
SimpleRL MATH-500 & 2.800 & 4.400 & 5.400 \\
\bottomrule
\end{tabular}%
}
\caption{Per-task raw metric values after the third intervention in the full budget evaluation. }
\label{tab:appendix-raw-full-int3}
\end{table*}

\begin{table*}[t]
\centering
\tiny
\setlength{\tabcolsep}{3pt}
\resizebox{0.5\textwidth}{!}{%
\begin{tabular}{lccc}
\toprule
Task & Q32 & DS-R1 & Claude \\
\midrule
FineWeb Pretraining & 3.476 & 3.476 & 3.415 \\
ConvKAN CIFAR-10 & 32.950 & 49.040 & 39.630 \\
Open-R1 MATH-500 & 22.000 & 2.200 & 45.800 \\
Room Selector Tuning & 0.379 & 0.138 & 1.000 \\
Uni2TS ETTh1 Forecasting & 0.375 & 0.375 & 0.375 \\
TimesFM Long Horizon & 0.509 & 0.509 & 0.509 \\
ModernBERT MNLI & 35.450 & 35.450 & 67.540 \\
AirBench CIFAR-10 & 93.680 & 93.560 & 93.740 \\
NoisyGL Cora GCN & 69.800 & 69.800 & 70.900 \\
TabMini Promoters & 0.968 & 0.968 & 0.965 \\
Adult Tabular Diffusion & 85.530 & 85.530 & 87.240 \\
xLSTM Parity & 0.298 & 0.298 & 1.000 \\
VBLL Yacht Regression & 0.334 & 0.334 & 0.404 \\
TabM California Housing & 0.504 & 0.504 & 0.493 \\
TimeXer PJM Forecasting & 0.121 & 0.121 & 0.119 \\
ABLkit HWF Reasoning & 97.750 & 96.850 & 96.100 \\
tunedGNN Cora GCN & 82.400 & 82.400 & 83.900 \\
verl GRPO GSM8K & 84.760 & 85.670 & 86.660 \\
ForestDiffusion Iris & 0.979 & 0.979 & 0.979 \\
VAR ImageNet 256 & 70.710 & 70.710 & 80.490 \\
RAGEN Bandit Alignment & 0.995 & 0.997 & 0.998 \\
Chronos Weather Forecasting & 0.593 & 0.593 & 0.372 \\
HyperbolicCV CIFAR-100 & 50.450 & 52.380 & 74.820 \\
iTransformer ETTm2 Forecasting & 0.185 & 0.185 & 0.186 \\
TimeMixer ETTm2 Forecasting & 0.174 & 0.176 & 0.175 \\
ART 2048 & 0.280 & 0.280 & 0.160 \\
Multimodal Open-R1 MathVista & 48.400 & 48.400 & 46.300 \\
RankUp UTKFace Regression & 17.310 & 22.490 & 5.570 \\
SparseTSF ETTm1 Forecasting & 0.340 & 0.340 & 0.335 \\
SimpleRL MATH-500 & 2.800 & 5.400 & 5.400 \\
\bottomrule
\end{tabular}%
}
\caption{Per-task raw metric values after the fourth intervention in the full budget evaluation. }
\label{tab:appendix-raw-full-int4}
\end{table*}

\begin{table*}[t]
\centering
\tiny
\setlength{\tabcolsep}{3pt}
\resizebox{0.5\textwidth}{!}{%
\begin{tabular}{lccc}
\toprule
Task & Q32 & DS-R1 & Claude \\
\midrule
FineWeb Pretraining & 3.468 & 3.468 & 3.403 \\
ConvKAN CIFAR-10 & 59.570 & 33.280 & 28.470 \\
Open-R1 MATH-500 & 22.000 & 2.200 & 46.000 \\
Room Selector Tuning & 0.379 & 0.069 & 1.000 \\
Uni2TS ETTh1 Forecasting & 0.375 & 0.375 & 0.375 \\
TimesFM Long Horizon & 0.509 & 0.509 & 0.509 \\
ModernBERT MNLI & 35.450 & 35.450 & 67.970 \\
AirBench CIFAR-10 & 93.680 & 93.490 & 93.730 \\
NoisyGL Cora GCN & 69.800 & 69.800 & 70.800 \\
TabMini Promoters & 0.968 & 0.968 & 0.982 \\
Adult Tabular Diffusion & 85.530 & 85.530 & 87.420 \\
xLSTM Parity & 0.298 & 0.298 & 1.000 \\
VBLL Yacht Regression & 0.334 & 0.334 & 0.356 \\
TabM California Housing & 0.504 & 0.504 & 0.505 \\
TimeXer PJM Forecasting & 0.121 & 0.121 & 0.114 \\
ABLkit HWF Reasoning & 97.500 & 97.550 & 97.450 \\
tunedGNN Cora GCN & 81.600 & 83.500 & 83.900 \\
verl GRPO GSM8K & 85.600 & 84.990 & 84.690 \\
ForestDiffusion Iris & 0.979 & 0.979 & 0.979 \\
VAR ImageNet 256 & 70.710 & 70.710 & 82.390 \\
RAGEN Bandit Alignment & 0.997 & 0.998 & 0.996 \\
Chronos Weather Forecasting & 0.389 & 0.389 & 0.370 \\
HyperbolicCV CIFAR-100 & 57.770 & 54.460 & 74.380 \\
iTransformer ETTm2 Forecasting & 0.185 & 0.183 & 0.185 \\
TimeMixer ETTm2 Forecasting & 0.175 & 0.174 & 0.174 \\
ART 2048 & 0.280 & 0.280 & 0.160 \\
Multimodal Open-R1 MathVista & 39.000 & 39.000 & 48.800 \\
RankUp UTKFace Regression & 6.150 & 6.270 & 5.380 \\
SparseTSF ETTm1 Forecasting & 0.340 & 0.340 & 0.323 \\
SimpleRL MATH-500 & 4.200 & 5.600 & 3.800 \\
\bottomrule
\end{tabular}%
}
\caption{Per-task raw metric values after the fifth intervention in the full budget evaluation. }
\label{tab:appendix-raw-full-int5}
\end{table*}

\begin{table*}[t]
\centering
\tiny
\setlength{\tabcolsep}{3pt}
\resizebox{0.5\textwidth}{!}{%
\begin{tabular}{lccc}
\toprule
Task & Q32 & DS-R1 & Claude \\
\midrule
FineWeb Pretraining & 0.550 & 0.550 & 0.619 \\
ConvKAN CIFAR-10 & 0.121 & 0.028 & 0.000 \\
Open-R1 MATH-500 & 0.245 & 0.000 & 0.541 \\
Room Selector Tuning & 0.000 & -0.908 & 1.000 \\
Uni2TS ETTh1 Forecasting & 1.000 & 1.000 & 1.000 \\
TimesFM Long Horizon & 0.991 & 0.991 & 0.991 \\
ModernBERT MNLI & -1.000 & -1.000 & -1.000 \\
AirBench CIFAR-10 & -0.788 & -0.364 & -0.667 \\
NoisyGL Cora GCN & 0.000 & -0.189 & 0.377 \\
TabMini Promoters & 0.000 & 1.000 & 1.000 \\
Adult Tabular Diffusion & 0.000 & 0.069 & 1.000 \\
xLSTM Parity & 0.000 & 0.000 & 0.999 \\
VBLL Yacht Regression & 0.339 & 0.339 & 0.440 \\
TabM California Housing & 0.308 & 0.308 & 0.308 \\
TimeXer PJM Forecasting & -0.331 & -0.331 & -0.331 \\
ABLkit HWF Reasoning & -0.562 & -0.500 & -1.000 \\
tunedGNN Cora GCN & -0.375 & -0.375 & -0.750 \\
verl GRPO GSM8K & 0.122 & -0.122 & -0.527 \\
ForestDiffusion Iris & 1.000 & 1.000 & 1.000 \\
VAR ImageNet 256 & 0.000 & 0.000 & -0.029 \\
RAGEN Bandit Alignment & 0.995 & 0.996 & 0.996 \\
Chronos Weather Forecasting & 0.000 & 0.000 & -0.057 \\
HyperbolicCV CIFAR-100 & -0.105 & -0.020 & 1.000 \\
iTransformer ETTm2 Forecasting & -0.445 & -0.445 & 0.350 \\
TimeMixer ETTm2 Forecasting & 0.869 & 0.847 & 0.897 \\
ART 2048 & 0.467 & 0.467 & 0.000 \\
Multimodal Open-R1 MathVista & -1.000 & -1.000 & -1.000 \\
RankUp UTKFace Regression & -0.803 & -1.000 & 0.404 \\
SparseTSF ETTm1 Forecasting & 0.000 & 0.000 & -0.006 \\
SimpleRL MATH-500 & -0.007 & -0.014 & 0.021 \\
\bottomrule
\end{tabular}%
}
\caption{Per-task bounded normalized scores after the first intervention in the full budget evaluation. Scores are bounded to $[-1,1]$.}
\label{tab:appendix-bounded-full-int1}
\end{table*}

\begin{table*}[t]
\centering
\tiny
\setlength{\tabcolsep}{3pt}
\resizebox{0.5\textwidth}{!}{%
\begin{tabular}{lccc}
\toprule
Task & Q32 & DS-R1 & Claude \\
\midrule
FineWeb Pretraining & 0.698 & 0.698 & 0.850 \\
ConvKAN CIFAR-10 & 0.002 & 0.515 & 0.261 \\
Open-R1 MATH-500 & 0.245 & 0.000 & 0.541 \\
Room Selector Tuning & 0.000 & -1.000 & 0.908 \\
Uni2TS ETTh1 Forecasting & 1.000 & 1.000 & 1.000 \\
TimesFM Long Horizon & 0.991 & 0.991 & 0.991 \\
ModernBERT MNLI & -1.000 & -1.000 & -1.000 \\
AirBench CIFAR-10 & -0.212 & -0.364 & 0.152 \\
NoisyGL Cora GCN & -0.189 & -0.189 & -0.943 \\
TabMini Promoters & 1.000 & 1.000 & 1.000 \\
Adult Tabular Diffusion & 0.069 & 0.069 & 1.000 \\
xLSTM Parity & 0.247 & 1.000 & 1.000 \\
VBLL Yacht Regression & 0.339 & 0.339 & 0.321 \\
TabM California Housing & 0.308 & 0.308 & 0.425 \\
TimeXer PJM Forecasting & -0.331 & -0.331 & -0.381 \\
ABLkit HWF Reasoning & -1.000 & -1.000 & -0.250 \\
tunedGNN Cora GCN & -0.250 & -0.188 & 0.188 \\
verl GRPO GSM8K & 0.162 & 0.243 & 0.608 \\
ForestDiffusion Iris & 1.000 & 1.000 & 1.000 \\
VAR ImageNet 256 & 0.000 & 0.000 & 0.100 \\
RAGEN Bandit Alignment & 0.996 & 0.997 & 0.997 \\
Chronos Weather Forecasting & 0.020 & 0.020 & 0.020 \\
HyperbolicCV CIFAR-100 & 0.024 & -0.160 & 0.931 \\
iTransformer ETTm2 Forecasting & -0.445 & -0.445 & -0.528 \\
TimeMixer ETTm2 Forecasting & 1.000 & 1.000 & -0.821 \\
ART 2048 & 0.467 & 0.467 & 0.000 \\
Multimodal Open-R1 MathVista & -0.071 & -0.071 & -0.036 \\
RankUp UTKFace Regression & 0.236 & 0.229 & 0.517 \\
SparseTSF ETTm1 Forecasting & 0.000 & 0.000 & -0.027 \\
SimpleRL MATH-500 & -0.027 & -0.014 & 0.079 \\
\bottomrule
\end{tabular}%
}
\caption{Per-task bounded normalized scores after the second intervention in the full budget evaluation. Scores are bounded to $[-1,1]$.}
\label{tab:appendix-bounded-full-int2}
\end{table*}

\begin{table*}[t]
\centering
\tiny
\setlength{\tabcolsep}{3pt}
\resizebox{0.5\textwidth}{!}{%
\begin{tabular}{lccc}
\toprule
Task & Q32 & DS-R1 & Claude \\
\midrule
FineWeb Pretraining & 0.769 & 0.769 & 0.965 \\
ConvKAN CIFAR-10 & -0.129 & 0.315 & 0.126 \\
Open-R1 MATH-500 & 0.245 & 0.000 & 0.539 \\
Room Selector Tuning & 0.000 & 0.101 & 1.000 \\
Uni2TS ETTh1 Forecasting & 1.000 & 1.000 & 1.000 \\
TimesFM Long Horizon & 0.991 & 0.991 & 0.991 \\
ModernBERT MNLI & -1.000 & -0.138 & 0.131 \\
AirBench CIFAR-10 & 0.000 & -0.364 & 0.000 \\
NoisyGL Cora GCN & -0.189 & -0.189 & 0.377 \\
TabMini Promoters & 1.000 & 1.000 & 1.000 \\
Adult Tabular Diffusion & 0.069 & 0.069 & 1.000 \\
xLSTM Parity & 0.247 & 0.247 & 1.000 \\
VBLL Yacht Regression & 0.339 & 0.339 & 0.422 \\
TabM California Housing & 0.308 & 0.308 & 0.429 \\
TimeXer PJM Forecasting & -0.331 & -0.331 & 0.000 \\
ABLkit HWF Reasoning & -1.000 & -1.000 & -0.094 \\
tunedGNN Cora GCN & -0.563 & -0.500 & 0.250 \\
verl GRPO GSM8K & 0.406 & 0.973 & 0.568 \\
ForestDiffusion Iris & 1.000 & 1.000 & 1.000 \\
VAR ImageNet 256 & 0.000 & 0.000 & -0.120 \\
RAGEN Bandit Alignment & 0.993 & 0.995 & 0.997 \\
Chronos Weather Forecasting & -0.005 & -0.005 & 0.000 \\
HyperbolicCV CIFAR-100 & -0.091 & -0.093 & 0.861 \\
iTransformer ETTm2 Forecasting & -0.445 & -0.445 & -0.216 \\
TimeMixer ETTm2 Forecasting & 1.000 & 1.000 & 1.000 \\
ART 2048 & 0.467 & 0.467 & 0.133 \\
Multimodal Open-R1 MathVista & -0.893 & -0.893 & -0.071 \\
RankUp UTKFace Regression & 0.620 & -0.012 & 0.129 \\
SparseTSF ETTm1 Forecasting & 0.000 & 0.000 & 0.572 \\
SimpleRL MATH-500 & -0.082 & -0.027 & 0.007 \\
\bottomrule
\end{tabular}%
}
\caption{Per-task bounded normalized scores after the third intervention in the full budget evaluation. Scores are bounded to $[-1,1]$.}
\label{tab:appendix-bounded-full-int3}
\end{table*}

\begin{table*}[t]
\centering
\tiny
\setlength{\tabcolsep}{3pt}
\resizebox{0.5\textwidth}{!}{%
\begin{tabular}{lccc}
\toprule
Task & Q32 & DS-R1 & Claude \\
\midrule
FineWeb Pretraining & 0.813 & 0.813 & 1.000 \\
ConvKAN CIFAR-10 & -0.052 & 0.278 & 0.085 \\
Open-R1 MATH-500 & 0.245 & 0.000 & 0.539 \\
Room Selector Tuning & 0.000 & -0.706 & 1.000 \\
Uni2TS ETTh1 Forecasting & 1.000 & 1.000 & 1.000 \\
TimesFM Long Horizon & 0.991 & 0.991 & 0.991 \\
ModernBERT MNLI & -1.000 & -1.000 & 0.150 \\
AirBench CIFAR-10 & 0.000 & -0.364 & 0.182 \\
NoisyGL Cora GCN & -0.189 & -0.189 & 0.849 \\
TabMini Promoters & 1.000 & 1.000 & 1.000 \\
Adult Tabular Diffusion & 0.069 & 0.069 & 1.000 \\
xLSTM Parity & 0.247 & 0.247 & 1.000 \\
VBLL Yacht Regression & 0.339 & 0.339 & 0.161 \\
TabM California Housing & 0.308 & 0.308 & 0.436 \\
TimeXer PJM Forecasting & -0.331 & -0.331 & -0.252 \\
ABLkit HWF Reasoning & 0.094 & -0.469 & -0.937 \\
tunedGNN Cora GCN & -0.688 & -0.688 & 0.250 \\
verl GRPO GSM8K & 0.284 & 0.771 & 1.000 \\
ForestDiffusion Iris & 1.000 & 1.000 & 1.000 \\
VAR ImageNet 256 & 0.000 & 0.000 & -0.146 \\
RAGEN Bandit Alignment & 0.994 & 0.997 & 0.998 \\
Chronos Weather Forecasting & -0.984 & -0.984 & 0.000 \\
HyperbolicCV CIFAR-100 & -0.288 & -0.198 & 0.848 \\
iTransformer ETTm2 Forecasting & -0.445 & -0.445 & -0.889 \\
TimeMixer ETTm2 Forecasting & 1.000 & 0.791 & 1.000 \\
ART 2048 & 0.467 & 0.467 & 0.267 \\
Multimodal Open-R1 MathVista & -0.143 & -0.143 & -0.893 \\
RankUp UTKFace Regression & -1.000 & -1.000 & 0.545 \\
SparseTSF ETTm1 Forecasting & 0.000 & 0.000 & 0.189 \\
SimpleRL MATH-500 & -0.082 & 0.007 & 0.007 \\
\bottomrule
\end{tabular}%
}
\caption{Per-task bounded normalized scores after the fourth intervention in the full budget evaluation. Scores are bounded to $[-1,1]$.}
\label{tab:appendix-bounded-full-int4}
\end{table*}

\begin{table*}[t]
\centering
\tiny
\setlength{\tabcolsep}{3pt}
\resizebox{0.5\textwidth}{!}{%
\begin{tabular}{lccc}
\toprule
Task & Q32 & DS-R1 & Claude \\
\midrule
FineWeb Pretraining & 0.841 & 0.841 & 1.000 \\
ConvKAN CIFAR-10 & 0.495 & -0.046 & -0.144 \\
Open-R1 MATH-500 & 0.245 & 0.000 & 0.541 \\
Room Selector Tuning & 0.000 & -0.908 & 1.000 \\
Uni2TS ETTh1 Forecasting & 1.000 & 1.000 & 1.000 \\
TimesFM Long Horizon & 0.991 & 0.991 & 0.991 \\
ModernBERT MNLI & -1.000 & -1.000 & 0.166 \\
AirBench CIFAR-10 & 0.000 & -0.576 & 0.152 \\
NoisyGL Cora GCN & -0.189 & -0.189 & 0.755 \\
TabMini Promoters & 1.000 & 1.000 & 1.000 \\
Adult Tabular Diffusion & 0.069 & 0.069 & 1.000 \\
xLSTM Parity & 0.247 & 0.247 & 1.000 \\
VBLL Yacht Regression & 0.339 & 0.339 & 0.283 \\
TabM California Housing & 0.308 & 0.308 & 0.301 \\
TimeXer PJM Forecasting & -0.331 & -0.331 & 0.000 \\
ABLkit HWF Reasoning & -0.062 & -0.031 & -0.094 \\
tunedGNN Cora GCN & -1.000 & 0.000 & 0.250 \\
verl GRPO GSM8K & 0.730 & 0.406 & 0.243 \\
ForestDiffusion Iris & 1.000 & 1.000 & 1.000 \\
VAR ImageNet 256 & 0.000 & 0.000 & -0.174 \\
RAGEN Bandit Alignment & 0.996 & 0.997 & 0.995 \\
Chronos Weather Forecasting & -0.077 & -0.077 & 0.007 \\
HyperbolicCV CIFAR-100 & 0.053 & -0.101 & 0.828 \\
iTransformer ETTm2 Forecasting & -0.445 & 0.000 & -0.461 \\
TimeMixer ETTm2 Forecasting & 0.916 & 1.000 & 1.000 \\
ART 2048 & 0.467 & 0.467 & 0.267 \\
Multimodal Open-R1 MathVista & -1.000 & -1.000 & 0.000 \\
RankUp UTKFace Regression & 0.178 & 0.098 & 0.667 \\
SparseTSF ETTm1 Forecasting & 0.000 & 0.000 & 0.646 \\
SimpleRL MATH-500 & -0.034 & 0.014 & -0.048 \\
\bottomrule
\end{tabular}%
}
\caption{Per-task bounded normalized scores after the fifth intervention in the full budget evaluation. Scores are bounded to $[-1,1]$.}
\label{tab:appendix-bounded-full-int5}
\end{table*}

\begin{table*}[t]
\centering
\tiny
\setlength{\tabcolsep}{3pt}
\resizebox{0.5\textwidth}{!}{%
\begin{tabular}{lccc}
\toprule
Task & Q32 & DS-R1 & Claude \\
\midrule
FineWeb Pretraining & 96.5 & 96.5 & 97.0 \\
ConvKAN CIFAR-10 & 49.2 & 43.8 & 42.2 \\
Open-R1 MATH-500 & 26.5 & 2.6 & 55.4 \\
Room Selector Tuning & 52.6 & 9.6 & 138.7 \\
Uni2TS ETTh1 Forecasting & 100.0 & 100.0 & 100.0 \\
TimesFM Long Horizon & 99.9 & 99.9 & 99.9 \\
ModernBERT MNLI & 39.2 & 39.2 & 39.2 \\
AirBench CIFAR-10 & 99.4 & 99.5 & 99.4 \\
NoisyGL Cora GCN & 98.5 & 98.2 & 99.1 \\
TabMini Promoters & 99.8 & 104.1 & 105.6 \\
Adult Tabular Diffusion & 99.7 & 99.8 & 102.0 \\
xLSTM Parity & 6.7 & 6.7 & 99.9 \\
VBLL Yacht Regression & 257.5 & 257.5 & 292.0 \\
TabM California Housing & 87.5 & 87.5 & 87.5 \\
TimeXer PJM Forecasting & 77.0 & 77.0 & 77.0 \\
ABLkit HWF Reasoning & 97.5 & 97.6 & 96.8 \\
tunedGNN Cora GCN & 97.4 & 97.4 & 96.7 \\
verl GRPO GSM8K & 98.1 & 97.6 & 96.7 \\
ForestDiffusion Iris & 100.9 & 100.9 & 100.9 \\
VAR ImageNet 256 & 5.0 & 5.0 & 4.9 \\
RAGEN Bandit Alignment & 99.6 & 99.6 & 99.7 \\
Chronos Weather Forecasting & 39.7 & 39.7 & 38.4 \\
HyperbolicCV CIFAR-100 & 69.6 & 72.0 & 100.1 \\
iTransformer ETTm2 Forecasting & 97.5 & 97.5 & 98.9 \\
TimeMixer ETTm2 Forecasting & 99.8 & 99.8 & 99.8 \\
ART 2048 & 46.7 & 46.7 & 0.0 \\
Multimodal Open-R1 MathVista & 77.9 & 77.9 & 77.9 \\
RankUp UTKFace Regression & 63.1 & 54.5 & 83.8 \\
SparseTSF ETTm1 Forecasting & 92.4 & 92.4 & 92.3 \\
SimpleRL MATH-500 & 14.5 & 14.0 & 16.9 \\
\bottomrule
\end{tabular}%
}
\caption{Per-task anchor attainment after the first intervention in the full budget evaluation, reported as percentages.}
\label{tab:appendix-anchor-attainment-full-int1}
\end{table*}

\begin{table*}[t]
\centering
\tiny
\setlength{\tabcolsep}{3pt}
\resizebox{0.5\textwidth}{!}{%
\begin{tabular}{lccc}
\toprule
Task & Q32 & DS-R1 & Claude \\
\midrule
FineWeb Pretraining & 97.6 & 97.6 & 98.8 \\
ConvKAN CIFAR-10 & 42.3 & 71.9 & 57.3 \\
Open-R1 MATH-500 & 26.5 & 2.6 & 55.4 \\
Room Selector Tuning & 52.6 & 0.0 & 95.7 \\
Uni2TS ETTh1 Forecasting & 100.0 & 100.0 & 100.0 \\
TimesFM Long Horizon & 99.9 & 99.9 & 99.9 \\
ModernBERT MNLI & 39.2 & 39.2 & 39.2 \\
AirBench CIFAR-10 & 99.6 & 99.5 & 99.7 \\
NoisyGL Cora GCN & 98.2 & 98.2 & 97.1 \\
TabMini Promoters & 104.1 & 104.1 & 104.3 \\
Adult Tabular Diffusion & 99.8 & 99.8 & 102.2 \\
xLSTM Parity & 29.8 & 100.0 & 100.0 \\
VBLL Yacht Regression & 257.5 & 257.5 & 251.9 \\
TabM California Housing & 87.5 & 87.5 & 89.4 \\
TimeXer PJM Forecasting & 77.0 & 77.0 & 76.3 \\
ABLkit HWF Reasoning & 96.1 & 96.1 & 98.0 \\
tunedGNN Cora GCN & 97.6 & 97.8 & 98.5 \\
verl GRPO GSM8K & 98.2 & 98.4 & 99.1 \\
ForestDiffusion Iris & 100.9 & 100.9 & 101.1 \\
VAR ImageNet 256 & 5.0 & 5.0 & 5.5 \\
RAGEN Bandit Alignment & 99.6 & 99.8 & 99.7 \\
Chronos Weather Forecasting & 40.2 & 40.2 & 40.2 \\
HyperbolicCV CIFAR-100 & 73.2 & 68.1 & 98.1 \\
iTransformer ETTm2 Forecasting & 97.5 & 97.5 & 97.3 \\
TimeMixer ETTm2 Forecasting & 100.4 & 100.1 & 97.2 \\
ART 2048 & 46.7 & 46.7 & 0.0 \\
Multimodal Open-R1 MathVista & 94.2 & 94.2 & 94.4 \\
RankUp UTKFace Regression & 80.1 & 80.0 & 86.4 \\
SparseTSF ETTm1 Forecasting & 92.4 & 92.4 & 92.2 \\
SimpleRL MATH-500 & 12.8 & 14.0 & 21.8 \\
\bottomrule
\end{tabular}%
}
\caption{Per-task anchor attainment after the second intervention in the full budget evaluation, reported as percentages.}
\label{tab:appendix-anchor-attainment-full-int2}
\end{table*}

\begin{table*}[t]
\centering
\tiny
\setlength{\tabcolsep}{3pt}
\resizebox{0.5\textwidth}{!}{%
\begin{tabular}{lccc}
\toprule
Task & Q32 & DS-R1 & Claude \\
\midrule
FineWeb Pretraining & 98.2 & 98.2 & 99.7 \\
ConvKAN CIFAR-10 & 34.7 & 60.4 & 49.5 \\
Open-R1 MATH-500 & 26.5 & 2.6 & 55.1 \\
Room Selector Tuning & 52.6 & 57.4 & 138.7 \\
Uni2TS ETTh1 Forecasting & 100.0 & 100.0 & 100.0 \\
TimesFM Long Horizon & 99.9 & 99.9 & 99.9 \\
ModernBERT MNLI & 39.2 & 66.1 & 74.1 \\
AirBench CIFAR-10 & 99.6 & 99.5 & 99.6 \\
NoisyGL Cora GCN & 98.2 & 98.2 & 99.1 \\
TabMini Promoters & 104.1 & 104.1 & 105.6 \\
Adult Tabular Diffusion & 99.8 & 99.8 & 102.0 \\
xLSTM Parity & 29.8 & 29.8 & 100.0 \\
VBLL Yacht Regression & 257.5 & 257.5 & 285.0 \\
TabM California Housing & 87.5 & 87.5 & 89.4 \\
TimeXer PJM Forecasting & 77.0 & 77.0 & 81.7 \\
ABLkit HWF Reasoning & 96.3 & 96.3 & 98.2 \\
tunedGNN Cora GCN & 97.1 & 97.2 & 98.6 \\
verl GRPO GSM8K & 98.7 & 99.9 & 99.1 \\
ForestDiffusion Iris & 100.9 & 100.9 & 100.9 \\
VAR ImageNet 256 & 5.0 & 5.0 & 4.5 \\
RAGEN Bandit Alignment & 99.4 & 99.6 & 99.7 \\
Chronos Weather Forecasting & 39.6 & 39.6 & 39.7 \\
HyperbolicCV CIFAR-100 & 70.0 & 70.0 & 96.2 \\
iTransformer ETTm2 Forecasting & 97.5 & 97.5 & 97.9 \\
TimeMixer ETTm2 Forecasting & 100.2 & 100.1 & 100.1 \\
ART 2048 & 46.7 & 46.7 & 13.3 \\
Multimodal Open-R1 MathVista & 89.7 & 89.7 & 94.2 \\
RankUp UTKFace Regression & 89.0 & 75.2 & 77.9 \\
SparseTSF ETTm1 Forecasting & 92.4 & 92.4 & 96.6 \\
SimpleRL MATH-500 & 8.1 & 12.8 & 15.7 \\
\bottomrule
\end{tabular}%
}
\caption{Per-task anchor attainment after the third intervention in the full budget evaluation, reported as percentages.}
\label{tab:appendix-anchor-attainment-full-int3}
\end{table*}

\begin{table*}[t]
\centering
\tiny
\setlength{\tabcolsep}{3pt}
\resizebox{0.5\textwidth}{!}{%
\begin{tabular}{lccc}
\toprule
Task & Q32 & DS-R1 & Claude \\
\midrule
FineWeb Pretraining & 98.5 & 98.5 & 100.3 \\
ConvKAN CIFAR-10 & 39.1 & 58.3 & 47.1 \\
Open-R1 MATH-500 & 26.5 & 2.6 & 55.1 \\
Room Selector Tuning & 52.6 & 19.1 & 138.7 \\
Uni2TS ETTh1 Forecasting & 100.0 & 100.0 & 100.0 \\
TimesFM Long Horizon & 99.9 & 99.9 & 99.9 \\
ModernBERT MNLI & 39.2 & 39.2 & 74.7 \\
AirBench CIFAR-10 & 99.6 & 99.5 & 99.7 \\
NoisyGL Cora GCN & 98.2 & 98.2 & 99.8 \\
TabMini Promoters & 104.1 & 104.1 & 103.8 \\
Adult Tabular Diffusion & 99.8 & 99.8 & 101.7 \\
xLSTM Parity & 29.8 & 29.8 & 100.0 \\
VBLL Yacht Regression & 257.5 & 257.5 & 212.7 \\
TabM California Housing & 87.5 & 87.5 & 89.5 \\
TimeXer PJM Forecasting & 77.0 & 77.0 & 78.0 \\
ABLkit HWF Reasoning & 98.5 & 97.6 & 96.9 \\
tunedGNN Cora GCN & 96.8 & 96.8 & 98.6 \\
verl GRPO GSM8K & 98.4 & 99.5 & 100.6 \\
ForestDiffusion Iris & 100.9 & 100.9 & 100.9 \\
VAR ImageNet 256 & 5.0 & 5.0 & 4.4 \\
RAGEN Bandit Alignment & 99.5 & 99.7 & 99.8 \\
Chronos Weather Forecasting & 25.0 & 25.0 & 39.7 \\
HyperbolicCV CIFAR-100 & 64.6 & 67.1 & 95.8 \\
iTransformer ETTm2 Forecasting & 97.5 & 97.5 & 96.7 \\
TimeMixer ETTm2 Forecasting & 100.4 & 99.7 & 100.2 \\
ART 2048 & 46.7 & 46.7 & 26.7 \\
Multimodal Open-R1 MathVista & 93.8 & 93.8 & 89.7 \\
RankUp UTKFace Regression & 28.0 & 21.6 & 87.1 \\
SparseTSF ETTm1 Forecasting & 92.4 & 92.4 & 93.7 \\
SimpleRL MATH-500 & 8.1 & 15.7 & 15.7 \\
\bottomrule
\end{tabular}%
}
\caption{Per-task anchor attainment after the fourth intervention in the full budget evaluation, reported as percentages.}
\label{tab:appendix-anchor-attainment-full-int4}
\end{table*}

\begin{table*}[t]
\centering
\tiny
\setlength{\tabcolsep}{3pt}
\resizebox{0.5\textwidth}{!}{%
\begin{tabular}{lccc}
\toprule
Task & Q32 & DS-R1 & Claude \\
\midrule
FineWeb Pretraining & 98.7 & 98.7 & 100.6 \\
ConvKAN CIFAR-10 & 70.8 & 39.5 & 33.8 \\
Open-R1 MATH-500 & 26.5 & 2.6 & 55.4 \\
Room Selector Tuning & 52.6 & 9.6 & 138.7 \\
Uni2TS ETTh1 Forecasting & 100.0 & 100.0 & 100.0 \\
TimesFM Long Horizon & 99.9 & 99.9 & 99.9 \\
ModernBERT MNLI & 39.2 & 39.2 & 75.2 \\
AirBench CIFAR-10 & 99.6 & 99.4 & 99.7 \\
NoisyGL Cora GCN & 98.2 & 98.2 & 99.6 \\
TabMini Promoters & 104.1 & 104.1 & 105.6 \\
Adult Tabular Diffusion & 99.8 & 99.8 & 102.0 \\
xLSTM Parity & 29.8 & 29.8 & 100.0 \\
VBLL Yacht Regression & 257.5 & 257.5 & 241.6 \\
TabM California Housing & 87.5 & 87.5 & 87.4 \\
TimeXer PJM Forecasting & 77.0 & 77.0 & 81.7 \\
ABLkit HWF Reasoning & 98.3 & 98.3 & 98.2 \\
tunedGNN Cora GCN & 95.9 & 98.1 & 98.6 \\
verl GRPO GSM8K & 99.4 & 98.7 & 98.4 \\
ForestDiffusion Iris & 100.9 & 100.9 & 100.9 \\
VAR ImageNet 256 & 5.0 & 5.0 & 4.3 \\
RAGEN Bandit Alignment & 99.7 & 99.8 & 99.6 \\
Chronos Weather Forecasting & 38.0 & 38.0 & 39.9 \\
HyperbolicCV CIFAR-100 & 74.0 & 69.8 & 95.3 \\
iTransformer ETTm2 Forecasting & 97.5 & 98.2 & 97.5 \\
TimeMixer ETTm2 Forecasting & 99.9 & 100.3 & 100.4 \\
ART 2048 & 46.7 & 46.7 & 26.7 \\
Multimodal Open-R1 MathVista & 75.6 & 75.6 & 94.6 \\
RankUp UTKFace Regression & 78.9 & 77.3 & 90.2 \\
SparseTSF ETTm1 Forecasting & 92.4 & 92.4 & 97.2 \\
SimpleRL MATH-500 & 12.2 & 16.3 & 11.0 \\
\bottomrule
\end{tabular}%
}
\caption{Per-task anchor attainment after the fifth intervention in the full budget evaluation, reported as percentages.}
\label{tab:appendix-anchor-attainment-full-int5}
\end{table*}

\begin{table*}[t]
\centering
\tiny
\setlength{\tabcolsep}{3pt}
\resizebox{0.6\textwidth}{!}{%
\begin{tabular}{lcccc}
\toprule
Task & AH-Claude & CC-CLI & AH-GPT & Codex-CLI \\
\midrule
FineWeb Pretraining & 5.931 & 5.771 & 5.931 & 5.859 \\
ConvKAN CIFAR-10 & 13.950 & 32.050 & 26.250 & 19.970 \\
Open-R1 MATH-500 & 46.000 & 45.000 & 45.000 & 45.000 \\
Room Selector Tuning & 0.862 & 1.000 & 1.000 & 1.000 \\
Uni2TS ETTh1 Forecasting & 0.375 & 0.375 & 0.375 & 0.375 \\
TimesFM Long Horizon & 0.509 & 0.509 & 0.509 & 0.509 \\
ModernBERT MNLI & 35.450 & 35.450 & 35.450 & 35.450 \\
AirBench CIFAR-10 & 88.570 & 88.930 & 89.010 & 89.010 \\
NoisyGL Cora GCN & 68.600 & 70.100 & 69.200 & 69.200 \\
TabMini Promoters & 0.956 & 0.982 & 0.928 & 0.928 \\
Adult Tabular Diffusion & 87.450 & 87.440 & 86.550 & 87.240 \\
xLSTM Parity & 0.021 & -0.021 & -0.021 & -0.021 \\
VBLL Yacht Regression & 0.465 & 0.465 & 0.530 & 0.697 \\
TabM California Housing & 0.530 & 0.530 & 0.530 & 0.530 \\
TimeXer PJM Forecasting & 0.127 & 0.127 & 0.127 & 0.127 \\
ABLkit HWF Reasoning & 97.150 & 97.900 & 97.050 & 97.400 \\
tunedGNN Cora GCN & 82.200 & 82.800 & 82.200 & 83.800 \\
verl GRPO GSM8K & 84.760 & 83.620 & 85.220 & 83.700 \\
ForestDiffusion Iris & 0.979 & 0.979 & 0.979 & 0.979 \\
VAR ImageNet 256 & 75.140 & 72.860 & 72.860 & 72.860 \\
RAGEN Bandit Alignment & 0.690 & 0.728 & 0.686 & 0.746 \\
Chronos Weather Forecasting & 0.385 & 0.385 & 0.380 & 0.368 \\
HyperbolicCV CIFAR-100 & 12.550 & 12.530 & 27.070 & 13.240 \\
iTransformer ETTm2 Forecasting & 0.182 & 0.184 & 0.184 & 0.184 \\
TimeMixer ETTm2 Forecasting & 0.177 & 0.176 & 0.177 & 0.177 \\
ART 2048 & 0.000 & 0.000 & 0.000 & 0.000 \\
Multimodal Open-R1 MathVista & 40.200 & 40.200 & 48.700 & 40.200 \\
RankUp UTKFace Regression & 259.580 & 394.830 & 733.660 & 1376.850 \\
SparseTSF ETTm1 Forecasting & 0.337 & 0.335 & 0.339 & 0.339 \\
SimpleRL MATH-500 & 5.800 & 5.800 & 5.800 & 5.800 \\
\bottomrule
\end{tabular}%
}
\caption{Per-task raw metric values after the first intervention in the harness ablation evaluation. }
\label{tab:appendix-raw-harness-int1}
\end{table*}

\begin{table*}[t]
\centering
\tiny
\setlength{\tabcolsep}{3pt}
\resizebox{0.6\textwidth}{!}{%
\begin{tabular}{lcccc}
\toprule
Task & AH-Claude & CC-CLI & AH-GPT & Codex-CLI \\
\midrule
FineWeb Pretraining & 5.414 & 5.284 & 5.414 & 5.224 \\
ConvKAN CIFAR-10 & 36.460 & 13.800 & 27.540 & 18.820 \\
Open-R1 MATH-500 & 46.000 & 45.000 & 45.000 & 45.000 \\
Room Selector Tuning & 0.690 & 0.690 & 1.000 & 1.000 \\
Uni2TS ETTh1 Forecasting & 0.375 & 0.375 & 0.375 & 0.375 \\
TimesFM Long Horizon & 0.509 & 0.509 & 0.509 & 0.509 \\
ModernBERT MNLI & 35.450 & 42.590 & 37.830 & 37.620 \\
AirBench CIFAR-10 & 89.270 & 89.250 & 89.220 & 89.090 \\
NoisyGL Cora GCN & 70.000 & 70.400 & 70.000 & 70.000 \\
TabMini Promoters & 0.982 & 0.970 & 0.928 & 0.928 \\
Adult Tabular Diffusion & 87.480 & 87.300 & 86.470 & 87.240 \\
xLSTM Parity & 0.021 & 0.001 & 0.011 & 0.011 \\
VBLL Yacht Regression & 0.465 & 0.465 & 0.530 & 0.321 \\
TabM California Housing & 0.513 & 0.513 & 0.499 & 0.511 \\
TimeXer PJM Forecasting & 0.144 & 0.144 & 0.127 & 0.127 \\
ABLkit HWF Reasoning & 97.550 & 94.350 & 97.650 & 97.050 \\
tunedGNN Cora GCN & 83.800 & 83.100 & 83.800 & 84.100 \\
verl GRPO GSM8K & 85.520 & 84.230 & 84.760 & 83.780 \\
ForestDiffusion Iris & 0.981 & 0.979 & 0.979 & 0.981 \\
VAR ImageNet 256 & 66.250 & 72.680 & 70.710 & 75.920 \\
RAGEN Bandit Alignment & 0.997 & 0.995 & 0.999 & 0.998 \\
Chronos Weather Forecasting & 0.372 & 0.368 & 0.368 & 0.372 \\
HyperbolicCV CIFAR-100 & 22.010 & 20.730 & 21.790 & 28.590 \\
iTransformer ETTm2 Forecasting & 0.184 & 0.184 & 0.183 & 0.183 \\
TimeMixer ETTm2 Forecasting & 0.176 & 0.177 & 0.177 & 0.177 \\
ART 2048 & 0.000 & 0.000 & 0.000 & 0.360 \\
Multimodal Open-R1 MathVista & 48.600 & 48.700 & 41.300 & 48.600 \\
RankUp UTKFace Regression & 21.820 & 15.810 & 70.940 & 27.780 \\
SparseTSF ETTm1 Forecasting & 0.337 & 0.335 & 0.339 & 0.339 \\
SimpleRL MATH-500 & 5.200 & 5.400 & 5.600 & 5.200 \\
\bottomrule
\end{tabular}%
}
\caption{Per-task raw metric values after the second intervention in the harness ablation evaluation. }
\label{tab:appendix-raw-harness-int2}
\end{table*}

\begin{table*}[t]
\centering
\tiny
\setlength{\tabcolsep}{3pt}
\resizebox{0.6\textwidth}{!}{%
\begin{tabular}{lcccc}
\toprule
Task & AH-Claude & CC-CLI & AH-GPT & Codex-CLI \\
\midrule
FineWeb Pretraining & 4.983 & 4.761 & 4.983 & 4.761 \\
ConvKAN CIFAR-10 & 47.950 & 42.400 & 21.540 & 22.880 \\
Open-R1 MATH-500 & 46.000 & 47.200 & 45.000 & 45.000 \\
Room Selector Tuning & 0.897 & 0.966 & 1.000 & 1.000 \\
Uni2TS ETTh1 Forecasting & 0.375 & 0.375 & 0.375 & 0.375 \\
TimesFM Long Horizon & 0.509 & 0.509 & 0.509 & 0.509 \\
ModernBERT MNLI & 36.940 & 44.330 & 35.450 & 35.450 \\
AirBench CIFAR-10 & 89.090 & 88.790 & 89.220 & 89.220 \\
NoisyGL Cora GCN & 70.600 & 69.000 & 69.300 & 70.800 \\
TabMini Promoters & 0.970 & 0.982 & 0.928 & 0.928 \\
Adult Tabular Diffusion & 87.640 & 87.440 & 86.550 & 87.480 \\
xLSTM Parity & 0.021 & 0.021 & 0.011 & 0.009 \\
VBLL Yacht Regression & 0.333 & 0.348 & 0.530 & 0.479 \\
TabM California Housing & 0.496 & 0.497 & 0.496 & 0.507 \\
TimeXer PJM Forecasting & 0.127 & 0.127 & 0.127 & 0.127 \\
ABLkit HWF Reasoning & 97.300 & 97.250 & 97.200 & 97.600 \\
tunedGNN Cora GCN & 85.200 & 83.800 & 84.400 & 82.200 \\
verl GRPO GSM8K & 85.820 & 84.000 & 84.910 & 82.870 \\
ForestDiffusion Iris & 0.964 & 0.979 & 0.979 & 0.979 \\
VAR ImageNet 256 & 70.710 & 78.770 & 71.190 & 66.250 \\
RAGEN Bandit Alignment & 0.994 & 0.997 & 0.999 & 0.995 \\
Chronos Weather Forecasting & 0.593 & 0.368 & 0.365 & 0.365 \\
HyperbolicCV CIFAR-100 & 22.450 & 22.500 & 29.180 & 29.780 \\
iTransformer ETTm2 Forecasting & 0.183 & 0.186 & 0.183 & 0.183 \\
TimeMixer ETTm2 Forecasting & 0.177 & 0.176 & 0.177 & 0.177 \\
ART 2048 & 0.000 & 0.360 & 0.000 & 0.120 \\
Multimodal Open-R1 MathVista & 48.800 & 48.600 & 48.800 & 48.700 \\
RankUp UTKFace Regression & 19.940 & 15.440 & 67.190 & 16.500 \\
SparseTSF ETTm1 Forecasting & 0.335 & 0.337 & 0.388 & 0.339 \\
SimpleRL MATH-500 & 6.000 & 5.200 & 4.600 & 5.600 \\
\bottomrule
\end{tabular}%
}
\caption{Per-task raw metric values after the third intervention in the harness ablation evaluation. }
\label{tab:appendix-raw-harness-int3}
\end{table*}

\begin{table*}[t]
\centering
\tiny
\setlength{\tabcolsep}{3pt}
\resizebox{0.6\textwidth}{!}{%
\begin{tabular}{lcccc}
\toprule
Task & AH-Claude & CC-CLI & AH-GPT & Codex-CLI \\
\midrule
FineWeb Pretraining & 4.513 & 4.347 & 4.513 & 4.340 \\
ConvKAN CIFAR-10 & 20.740 & 24.420 & 16.210 & 27.060 \\
Open-R1 MATH-500 & 46.000 & 47.200 & 45.000 & 47.200 \\
Room Selector Tuning & 0.897 & 1.000 & 1.000 & 1.000 \\
Uni2TS ETTh1 Forecasting & 0.375 & 0.375 & 0.375 & 0.375 \\
TimesFM Long Horizon & 0.509 & 0.509 & 0.509 & 0.509 \\
ModernBERT MNLI & 39.920 & 46.500 & 36.580 & 37.750 \\
AirBench CIFAR-10 & 89.250 & 89.220 & 89.270 & 89.220 \\
NoisyGL Cora GCN & 70.800 & 70.400 & 70.000 & 70.800 \\
TabMini Promoters & 0.979 & 0.970 & 0.928 & 0.925 \\
Adult Tabular Diffusion & 87.240 & 87.350 & 86.380 & 87.480 \\
xLSTM Parity & -0.035 & 0.021 & 0.011 & 0.011 \\
VBLL Yacht Regression & 0.333 & 0.333 & 0.530 & 0.321 \\
TabM California Housing & 0.497 & 0.495 & 0.497 & 0.496 \\
TimeXer PJM Forecasting & 0.127 & 0.127 & 0.127 & 0.127 \\
ABLkit HWF Reasoning & 97.450 & 97.850 & 97.600 & 97.350 \\
tunedGNN Cora GCN & 85.200 & 82.900 & 84.400 & 84.100 \\
verl GRPO GSM8K & 84.840 & 83.850 & 84.840 & 83.320 \\
ForestDiffusion Iris & 0.979 & 0.971 & 0.979 & 0.979 \\
VAR ImageNet 256 & 73.650 & 80.490 & 70.710 & 64.000 \\
RAGEN Bandit Alignment & 0.999 & 1.000 & 0.998 & 0.999 \\
Chronos Weather Forecasting & 0.382 & 0.504 & 0.604 & 0.364 \\
HyperbolicCV CIFAR-100 & 22.340 & 30.270 & 29.230 & 27.710 \\
iTransformer ETTm2 Forecasting & 0.186 & 0.183 & 0.183 & 0.183 \\
TimeMixer ETTm2 Forecasting & 0.176 & 0.177 & 0.177 & 0.177 \\
ART 2048 & 0.080 & 0.160 & 0.000 & 0.560 \\
Multimodal Open-R1 MathVista & 48.700 & 46.300 & 48.800 & 48.800 \\
RankUp UTKFace Regression & 16.280 & 45.040 & 17.760 & 74.350 \\
SparseTSF ETTm1 Forecasting & 0.340 & 0.326 & 0.339 & 0.339 \\
SimpleRL MATH-500 & 5.600 & 7.300 & 5.600 & 6.700 \\
\bottomrule
\end{tabular}%
}
\caption{Per-task raw metric values after the fourth intervention in the harness ablation evaluation. }
\label{tab:appendix-raw-harness-int4}
\end{table*}

\begin{table*}[t]
\centering
\tiny
\setlength{\tabcolsep}{3pt}
\resizebox{0.6\textwidth}{!}{%
\begin{tabular}{lcccc}
\toprule
Task & AH-Claude & CC-CLI & AH-GPT & Codex-CLI \\
\midrule
FineWeb Pretraining & 4.406 & 4.199 & 4.406 & 4.194 \\
ConvKAN CIFAR-10 & 37.620 & 21.930 & 29.590 & 23.950 \\
Open-R1 MATH-500 & 46.000 & 47.200 & 45.000 & 47.200 \\
Room Selector Tuning & 0.862 & 1.000 & 1.000 & 1.000 \\
Uni2TS ETTh1 Forecasting & 0.375 & 0.375 & 0.375 & 0.375 \\
TimesFM Long Horizon & 0.509 & 0.509 & 0.509 & 0.509 \\
ModernBERT MNLI & 38.070 & 44.180 & 38.130 & 38.270 \\
AirBench CIFAR-10 & 89.250 & 89.220 & 89.270 & 89.090 \\
NoisyGL Cora GCN & 70.800 & 70.900 & 70.000 & 70.800 \\
TabMini Promoters & 0.970 & 0.982 & 0.928 & 0.925 \\
Adult Tabular Diffusion & 87.420 & 87.470 & 86.550 & 87.530 \\
xLSTM Parity & 0.021 & 0.021 & 0.011 & 0.011 \\
VBLL Yacht Regression & 0.333 & 0.297 & 0.530 & 0.321 \\
TabM California Housing & 0.517 & 0.500 & 0.496 & 0.497 \\
TimeXer PJM Forecasting & 0.127 & 0.127 & 0.127 & 0.127 \\
ABLkit HWF Reasoning & 97.350 & 97.250 & 97.400 & 97.750 \\
tunedGNN Cora GCN & 85.200 & 83.800 & 84.400 & 84.100 \\
verl GRPO GSM8K & 85.440 & 83.620 & 85.370 & 83.930 \\
ForestDiffusion Iris & 0.979 & 0.979 & 0.979 & 0.964 \\
VAR ImageNet 256 & 75.140 & 82.060 & 70.710 & 67.260 \\
RAGEN Bandit Alignment & 0.997 & 0.999 & 0.998 & 0.997 \\
Chronos Weather Forecasting & 0.370 & 0.365 & 0.365 & 0.454 \\
HyperbolicCV CIFAR-100 & 30.450 & 33.060 & 29.140 & 25.760 \\
iTransformer ETTm2 Forecasting & 0.185 & 0.183 & 0.183 & 0.183 \\
TimeMixer ETTm2 Forecasting & 0.176 & 0.182 & 0.177 & 0.177 \\
ART 2048 & 0.080 & 0.240 & 0.280 & 0.000 \\
Multimodal Open-R1 MathVista & 48.800 & 48.800 & 48.800 & 48.800 \\
RankUp UTKFace Regression & 16.540 & 16.690 & 47.040 & 21.900 \\
SparseTSF ETTm1 Forecasting & 0.326 & 0.319 & 0.339 & 0.339 \\
SimpleRL MATH-500 & 4.600 & 6.700 & 3.800 & 4.600 \\
\bottomrule
\end{tabular}%
}
\caption{Per-task raw metric values after the fifth intervention in the harness ablation evaluation. }
\label{tab:appendix-raw-harness-int5}
\end{table*}

\begin{table*}[t]
\centering
\tiny
\setlength{\tabcolsep}{3pt}
\resizebox{0.6\textwidth}{!}{%
\begin{tabular}{lcccc}
\toprule
Task & AH-Claude & CC-CLI & AH-GPT & Codex-CLI \\
\midrule
FineWeb Pretraining & -1.000 & -1.000 & -1.000 & -1.000 \\
ConvKAN CIFAR-10 & -0.272 & 0.056 & -0.049 & -0.163 \\
Open-R1 MATH-500 & 0.545 & 0.533 & 0.533 & 0.533 \\
Room Selector Tuning & 1.000 & 1.000 & 1.000 & 1.000 \\
Uni2TS ETTh1 Forecasting & 1.000 & 1.000 & 1.000 & 1.000 \\
TimesFM Long Horizon & 0.991 & 0.991 & 0.990 & 0.991 \\
ModernBERT MNLI & -0.068 & -0.068 & -0.068 & -0.068 \\
AirBench CIFAR-10 & -0.106 & -0.033 & -0.016 & -0.016 \\
NoisyGL Cora GCN & -1.000 & 0.094 & -0.755 & -0.755 \\
TabMini Promoters & 1.000 & 1.000 & 0.000 & 0.000 \\
Adult Tabular Diffusion & 1.000 & 1.000 & 1.000 & 1.000 \\
xLSTM Parity & 0.011 & -0.033 & -0.033 & -0.033 \\
VBLL Yacht Regression & 1.000 & 1.000 & 1.000 & 1.000 \\
TabM California Housing & 0.066 & 0.066 & 0.066 & 0.066 \\
TimeXer PJM Forecasting & 0.151 & 0.151 & 0.151 & 0.151 \\
ABLkit HWF Reasoning & -0.171 & 0.257 & -0.229 & -0.029 \\
tunedGNN Cora GCN & -0.813 & -0.437 & -0.813 & 0.188 \\
verl GRPO GSM8K & 0.054 & -0.750 & 0.375 & -0.697 \\
ForestDiffusion Iris & 1.000 & 1.000 & 1.000 & 1.000 \\
VAR ImageNet 256 & -0.066 & -0.032 & -0.032 & -0.032 \\
RAGEN Bandit Alignment & 0.627 & 0.673 & 0.623 & 0.694 \\
Chronos Weather Forecasting & -0.057 & -0.057 & -0.035 & 0.020 \\
HyperbolicCV CIFAR-100 & -0.062 & -0.063 & 0.173 & -0.051 \\
iTransformer ETTm2 Forecasting & 0.350 & -0.216 & -0.278 & -0.278 \\
TimeMixer ETTm2 Forecasting & 0.248 & 0.530 & 0.370 & 0.369 \\
ART 2048 & 0.000 & 0.000 & 0.000 & 0.000 \\
Multimodal Open-R1 MathVista & -1.000 & -1.000 & -0.036 & -1.000 \\
RankUp UTKFace Regression & -1.000 & -1.000 & -1.000 & -1.000 \\
SparseTSF ETTm1 Forecasting & 0.151 & 0.215 & 0.056 & 0.056 \\
SimpleRL MATH-500 & -0.538 & -0.538 & -0.538 & -0.538 \\
\bottomrule
\end{tabular}%
}
\caption{Per-task bounded normalized scores after the first intervention in the harness ablation evaluation. Scores are bounded to $[-1,1]$.}
\label{tab:appendix-bounded-harness-int1}
\end{table*}

\begin{table*}[t]
\centering
\tiny
\setlength{\tabcolsep}{3pt}
\resizebox{0.6\textwidth}{!}{%
%
%
}
\caption{Per-task bounded normalized scores after the second intervention in the harness ablation evaluation. Scores are bounded to $[-1,1]$.}
\label{tab:appendix-bounded-harness-int2}
\end{table*}

\begin{table*}[t]
\centering
\tiny
\setlength{\tabcolsep}{3pt}
\resizebox{0.6\textwidth}{!}{%
%
%
}
\caption{Per-task bounded normalized scores after the third intervention in the harness ablation evaluation. Scores are bounded to $[-1,1]$.}
\label{tab:appendix-bounded-harness-int3}
\end{table*}

\begin{table*}[t]
\centering
\tiny
\setlength{\tabcolsep}{3pt}
\resizebox{0.6\textwidth}{!}{%
%
%
}
\caption{Per-task bounded normalized scores after the fourth intervention in the harness ablation evaluation. Scores are bounded to $[-1,1]$.}
\label{tab:appendix-bounded-harness-int4}
\end{table*}

\begin{table*}[t]
\centering
\tiny
\setlength{\tabcolsep}{3pt}
\resizebox{0.6\textwidth}{!}{%
%
%
}
\caption{Per-task bounded normalized scores after the fifth intervention in the harness ablation evaluation. Scores are bounded to $[-1,1]$.}
\label{tab:appendix-bounded-harness-int5}
\end{table*}

\begin{table*}[t]
\centering
\tiny
\setlength{\tabcolsep}{3pt}
\resizebox{0.6\textwidth}{!}{%
%
%
}
\caption{Per-task anchor attainment after the first intervention in the harness ablation evaluation, reported as percentages.}
\label{tab:appendix-anchor-attainment-harness-int1}
\end{table*}

\begin{table*}[t]
\centering
\tiny
\setlength{\tabcolsep}{3pt}
\resizebox{0.6\textwidth}{!}{%
%
%
}
\caption{Per-task anchor attainment after the second intervention in the harness ablation evaluation, reported as percentages.}
\label{tab:appendix-anchor-attainment-harness-int2}
\end{table*}

\begin{table*}[t]
\centering
\tiny
\setlength{\tabcolsep}{3pt}
\resizebox{0.6\textwidth}{!}{%
%
%
}
\caption{Per-task anchor attainment after the third intervention in the harness ablation evaluation, reported as percentages.}
\label{tab:appendix-anchor-attainment-harness-int3}
\end{table*}

\begin{table*}[t]
\centering
\tiny
\setlength{\tabcolsep}{3pt}
\resizebox{0.6\textwidth}{!}{%
%
%
}
\caption{Per-task anchor attainment after the fourth intervention in the harness ablation evaluation, reported as percentages.}
\label{tab:appendix-anchor-attainment-harness-int4}
\end{table*}

\begin{table*}[t]
\centering
\tiny
\setlength{\tabcolsep}{3pt}
\resizebox{0.6\textwidth}{!}{%
%
%
}
\caption{Per-task anchor attainment after the fifth intervention in the harness ablation evaluation, reported as percentages.}
\label{tab:appendix-anchor-attainment-harness-int5}
\end{table*}